# Finite-Sample Guarantees for Wasserstein Distributionally Robust Optimization: Breaking the Curse of Dimensionality

Rui Gao

Department of Information, Risk and Operations Management, University of Texas at Austin, rui.gao@mccombs.utexas.edu

Wasserstein distributionally robust optimization (DRO) aims to find robust and generalizable solutions by hedging against data perturbations in Wasserstein distance. Despite its recent empirical success in operations research and machine learning, existing performance guarantees for generic loss functions are either overly conservative due to the curse of dimensionality, or plausible only in large sample asymptotics. In this paper, we develop a non-asymptotic framework for analyzing the out-of-sample performance for Wasserstein robust learning and the generalization bound for its related Lipschitz and gradient regularization problems. To the best of our knowledge, this gives the first finite-sample guarantee for generic Wasserstein DRO problems without suffering from the curse of dimensionality. Our results highlight that Wasserstein DRO, with a properly chosen radius, balances between the empirical mean of the loss and the variation of the loss, measured by the Lipschitz norm or the gradient norm of the loss. Our analysis is based on two novel methodological developments that are of independent interest: 1) a new concentration inequality controlling the decay rate of large deviation probabilities by the variation of the loss and, 2) a localized Rademacher complexity theory based on the variation of the loss.

## 1. Introduction

Distributionally robust optimization (DRO) is an emerging paradigm for statistical learning and decision-making under uncertainty. It aims to provide robust and generalizable solutions by hedging against a set of distributions in the minimax sense. Different choices of distributional uncertainty set have been investigated thoroughly [72, 108, 79, 21, 35, 69, 28, 42, 96, 9, 97, 50, 8, 94, 36, 109]. In this paper, we focus on Wasserstein DRO [97, 36, 109, 15, 40, 99, 26]

$$\inf_{\theta \in \Theta} \sup_{\mathbb{P}: \mathcal{W}_p(\mathbb{P}, \mathbb{P}_n) \leq \rho_n} \mathbb{E}_{z \sim \mathbb{P}}[f_\theta(z)],$$

which finds a solution $\theta$ from a space $\Theta$ so as to minimize the Wasserstein robust loss, defined as the worst-case expectation of the loss function $f_\theta$ among a ball of distributions whose $p$-Wasserstein distance $\mathcal{W}_p$ to the empirical distribution $\mathbb{P}_n$ of sample size $n$ is at most $\rho_n > 0$. Due to its ability to hedge against data perturbations in high dimensions [15, 40] and its regularization effect [73, 74, 14, 39, 87, 5], Wasserstein DRO has recently been studied in many areas in machine learning [73, 14, 12, 80, 24, 41, 87, 56, 32, 65, 57, 1, 81, 29]; as well as other fields, such as automatic control [103, 75, 104, 27], finance [10], energy systems [91, 90, 31], statistics [73, 60, 41, 66, 11, 67], transportation [22]. We refer to [53] for a recent survey.

Among Wasserstein distances of different orders, $p = 1, 2$ are of particular interest both practically and theoretically. 1-Wasserstein DRO is useful when the loss function is bounded or has linear growth, and often leads to linear programming reformulation when 1-norm or $\infty$-norm is used [53, 73, 74]. 2-Wasserstein DRO applies to a larger class of loss functions such as quadratic loss [66, 75, 27]. Efficient gradient-descent algorithms have been developed by virtue of the convex quadratic subproblem associated with 2-Wasserstein DRO [80, 17, 58, 25]. Moreover, Lipschitz regularization and data-dependent gradient regularization deep learning problems are closely related to 1-Wasserstein DRO and 2-Wasserstein DRO, respectively.

Like many other (distributionally) robust optimization frameworks or regularization methods, obtaining a Wasserstein robust solution with good performance guarantees requires a proper hyperparameter tuning, namely, the selection of the radius of the Wasserstein ball $\rho_n$. On the one hand, the radius $\rho_n$ cannot be too small since otherwise, the problem behaves like empirical risk minimization or sample average approximation, thus losing the purpose of robustification. On the other hand, the radius $\rho_n$ cannot be so large that the solution might be overly conservative, which is one of the major criticism faced by traditional robust optimization. Practically, radius selection is often achieved via cross validation. From a statistical point of view, it is crucial to understand what is the correct scaling of the hyperparameter $\rho_n$ with respect to the sample size $n$ so as to ensure the robustness and generalization of the solution without sacrificing much out-of-sample performance.

Despite promising applications of Wasserstein DRO, its theoretical performance guarantee is limited. Esfahani and Kuhn [36] provides the first out-of-sample performance guarantee for Wasserstein DRO. Using the concentration of empirical Wasserstein distance [38], they show that if the radius is chosen in the order of $n^{-1/\max(2,d)}$, where $d$ is the dimension of the random data $z$, the underlying data-generating distribution $\mathbb{P}_{\text{true}}$ is contained in the Wasserstein ball with high probability. Thereby the Wasserstein robust loss of every feasible solution (and in particular the optimal solution) would be an upper bound of its true loss. This provides a finite-sample non-asymptotic guarantee for the Wasserstein robust solution, but unfortunately, such a bound suffers from the *curse of dimensionality* since the radius shrinks too slow even for problems in moderate dimensions.

To resolve the curse of dimensionality, a series of work by Blanchet et al. [14, 13, 16] consider an approach inspired from the empirical likelihood [54, 33]. Their principle is finding the smallest radius $\rho_n$ such that with high probability, the Wasserstein ball contains at least one distribution $\mathbb{P}$ – not necessarily equal to the true data-generating distribution $\mathbb{P}_{\text{true}}$ – for which there exists an optimal solution to $\min_\theta \mathbb{E}_{\mathbb{P}}[f_\theta]$ that is also optimal to the underlying true problem $\min_\theta \mathbb{E}_{\mathbb{P}_{\text{true}}}[f_\theta]$. Such choice leads to a confidence region of the optimal solution. Through an asymptotic analysis, they show that the radius $\rho_n$ can be chosen in the square-root order $1/\sqrt{n}$ for fixed dimension of the random variable[1]. This gives the first radius selection rule that does not suffer from the curse of dimensionality. However, one potential issue with this result is that the bound is valid only in the *asymptotic* sense, namely, as the sample size $n$ goes to infinity while fixing the dimension of the random variable. Ideally, we would like to have a performance guarantee for any finite sample size and dimension, especially for high-dimensional problems and for robust optimization where the sample size is comparatively not large.

For certain special classes of stochastic optimization problems, the non-asymptotic $1/\sqrt{n}$-rate has been developed. For 1-Wasserstein DRO with certain linear structure, such as linear regression/classification and their kernelization, Shafieezadeh-Abadeh et al. [74] shows that the radius can be chosen as $\tilde{O}(1/\sqrt{n})$ to achieve a finite-sample performance guarantee uniformly for all feasible solutions (we use $\tilde{O}$ to suppress the logarithmic dependence). Chen and Paschalidis [24] derives generalization bounds for certain class of 1-Wasserstein DRO problems that are equivalent to norm regularization. Xie et al. [100] provides performance guarantees for stochastic bottleneck problems by relating them to sample average approximations.

Yet, it remains largely unknown whether the non-asymptotic $1/\sqrt{n}$-rate holds for general loss functions. In this paper, we provide an affirmative answer to this open question under reasonable assumptions. Informally, our main result states the following performance guarantees for Wasserstein DRO.

THEOREM (INFORMAL). *Let $p \in [1,2]$. Set $\rho_n = \tilde{O}(1/\sqrt{n})$. Under appropriate conditions, with high probability, simultaneously for all $\theta \in \Theta$,*

$$\mathbb{E}_{z \sim \mathbb{P}_{\text{true}}}[f_\theta(z)] \leq \sup_{\mathbb{P}: \mathcal{W}_p(\mathbb{P}, \mathbb{P}_n) \leq \rho_n} \mathbb{E}_{z \sim \mathbb{P}}[f_\theta(z)] + \epsilon_n,$$

---

[1] In Blanchet et al. [14, Section 4.4], a result for high-dimensional LASSO problem is derived.

*where $\epsilon_n$ is a $\tilde{O}(1/n)$ higher-order term.*

The constants hidden in the big-O notation are made explicitly in our formal results and are shown to have a mild dependence on the dimension of the random variable for a variety of applications. This theorem shows that under the canonical root-$n$ radius, up to a high-order residual, the true loss is upper bounded by the Wasserstein robust loss uniformly for all $\theta$ and particularly for the robust optimal solution $\theta_{\rm rob}$:

$$\mathbb{E}_{z\sim \mathbb{P}_{\rm true}}[f_{\theta_{\rm rob}}(z)] \leq \sup_{\mathbb{P}:\mathcal{W}_p(\mathbb{P},\mathbb{P}_n)\leq \rho_n} \mathbb{E}_{z\sim \mathbb{P}}[f_{\theta_{\rm rob}}(z)] + \epsilon_n = \min_{\theta\in\Theta} \sup_{\mathbb{P}:\mathcal{W}_p(\mathbb{P},\mathbb{P}_n)\leq \rho_n} \mathbb{E}_{z\sim \mathbb{P}}[f_\theta(z)] + \epsilon_n.$$

When $f_\theta$ is the loss function of a supervised learning problem, the left-hand side represents the *generalization error* of the Wasserstein robust solution, and the right-hand side of the inequality above indicates that the optimal value of the Wasserstein robust loss minimization provides an generalization bound up to a high-order residual.

Recall the regularization effect of Wasserstein DRO [14, 74, 39]

$$\sup_{\mathbb{P}:\mathcal{W}_p(\mathbb{P},\mathbb{P}_n)\leq \rho_n} \mathbb{E}_{z\sim\mathbb{P}}[f_\theta(z)] = \mathbb{E}_{z\sim\mathbb{P}_n}[f_\theta(z)] + \rho_n \cdot \mathcal{V}(f_\theta) + \tilde{O}_p(1/n), \quad \forall \theta \in \Theta,$$

where $\mathcal{V}(\cdot)$ represents the *variation of the loss*, measured by the Lipschitz norm $\|f_\theta\|_{\rm Lip}$ when $p=1$ or the gradient norm $\mathbb{E}_{z\sim\mathbb{P}_n}[\|\nabla_z f_\theta(z)\|^2]^{1/2}$ when $p=2$ (we use $O_p$ for the big O in probability notation). Together with this result, our theorem highlights a principled *bias-variation trade-off* by properly choosing the radius for Wasserstein DRO, which balances between the empirical loss $\mathbb{E}_{\mathbb{P}_n}[f_{\theta_{\rm rob}}]$ and the variation of the loss $\mathcal{V}(f_{\theta_{\rm rob}})$ which controls the *generalization gap*:

$$\mathbb{E}_{z\sim\mathbb{P}_{\rm true}}[f_{\theta_{\rm rob}}(z)] - \mathbb{E}_{z\sim\mathbb{P}_n}[f_{\theta_{\rm rob}}(z)] \leq \frac{\rho_0}{\sqrt{n}} \cdot \mathcal{V}(f_{\theta_{\rm rob}}) + \tilde{O}_p(1/n).$$

Thus, the robust optimal solution $\theta_{\rm rob}$ achieves nice generalization capability by biasing towards a solution with small variation.

Variation-based regularization has become increasingly popular for many deep learning problems recently. For example, Lipschitz regularization and gradient regularization have shown superior empirical performance for adversarial learning and reinforcement learning [20, 43, 62, 82, 46, 64, 87, 49, 86, 37, 49, 68, 83, 95, 106]. Our results also provide statistical guarantees for Lipschitz regularization and gradient regularization.

Below, we briefly describe two methodological advancements that lead to our results. In our analysis, the main object of study is the *Wasserstein regularizer*:

$$\mathcal{R}_{\mathbb{Q},p}(\rho; f_\theta) = \sup_{\mathbb{P}:\mathcal{W}_p(\mathbb{P},\mathbb{Q})\leq \rho} \mathbb{E}_{z\sim\mathbb{P}}[f_\theta(z)] - \mathbb{E}_{z\sim\mathbb{Q}}[f_\theta(z)],$$

that is, the difference between the Wasserstein robust loss and the nominal loss under some distribution $\mathbb{Q}$ such as $\mathbb{P}_n$ or $\mathbb{P}_{\rm true}$.

First, in Section 3, leveraging tools from transportation-information inequalities (see, e.g., [45]) in modern probability theory, we derive a new large-deviation type concentration inequality for the empirical loss (Theorem 1). It shows that under proper conditions on the underlying data-generating distribution $\mathbb{P}_{\rm true}$, the decay rate of the tail probability is upper bounded by the inverse of Wasserstein regularizer $\mathcal{R}^{-1}_{\mathbb{P}_{\rm true},p}(\cdot; f_\theta)$ as well as the variation of the loss $\mathcal{V}(f_\theta)$. This result shows that the variation of the loss has a direct control on the deviation of the empirical loss from the ground truth. This is an analog of variance-based control often resulting from Chebyshev's or Bernstein's concentration inequalities.

Second, to extend the concentration result above from a single loss function to a family of loss functions, we develop two sets of results in Sections 4.1 and 4.2 respectively, one based on covering number arguments, and the other adapts tools from localized Rademacher complexity theory (see, e.g., [6, 34]). For the latter, we consider subsets of function classes whose variations are controlled, as opposed to usual approaches based on the mean or variance of the loss. These results are demonstrated in Section 5 using various examples, including feature-based newsvendor, linear prediction, portfolio optimization, Lipschitz regularization for kernel classes, and gradient regularization for neural networks.

Overall, we develop a non-asymptotic statistical analysis framework for Wasserstein DRO and its associated variation regularization, and demonstrate the bias-variation trade-off in Wasserstein robust learning when the radius is properly chosen. This can be served as a counterpart of the well-known bias-variance trade-off theory in machine learning.

**Related Work**

The generalization bounds for robust optimization dates back to Xu and Mannor [102], which studies generalization of learning algorithms from the viewpoint of robustness. In the introduction, we have elaborated on the literature that provide performance guarantees for Wasserstein DRO [36, 14, 13, 74, 16, 24, 100] and discuss their scopes and limitations. In addition to these literature, motivated by distribution shift in domain adaptation and adversarial learning, Lee and Raginsky [56], Sinha et al. [80], Najafi et al. [65] develop generalization bounds for Wasserstein DRO where the radius is fixed, not varying with the sample size. For divergence DRO and the related variance regularization, Lam [54] studies the calibration of the radius of divergence ball that recovers the best statistical guarantee. Asymptotics and non-asymptotics of divergence DRO and its bias-variance trade-off are investigated in [33, 34]. Besides DRO, Wasserstein distance and transportation-information inequality are also exploited to improve information-theoretic generalization bounds for learning algorithms [101, 59, 71, 92, 93].

The rest of the paper proceeds as follows. In Section 2, we briefly review some results in Wasserstein DRO and its variation regularization effect. We develop a new variation-based concentration inequality in Section 3. Based on these two sections, we derive generalization bounds for variation regularization and the finite-sample guarantees for Wasserstein DRO in Section 4. Applications of these results are in Section 5. We conclude the paper in Section 6. All proofs are deferred to the Appendices.

## 2. Wasserstein DRO and Variation Regularization

In this section, we introduce notations and provide some background on Wasserstein DRO and its variation regularization effect.

**Notation.** Let $\mathcal{Z}$ be a Banach space equipped with some norm $\|\cdot\|$ and let $\|\cdot\|_*$ be its dual norm. Define the diameter of $\mathcal{Z}$ as $\text{diam}(\mathcal{Z}) := \sup_{\tilde{z}, z \in \mathcal{Z}} \|\tilde{z} - z\|$. Let $p \in [1, \infty)$ and denote by $q$ its Hölder conjugate number, i.e., $\frac{1}{p} + \frac{1}{q} = 1$. We denote by $\mathcal{P}_p(\mathcal{Z})$ the set of Borel probability measures on $\mathcal{Z}$ with finite $p$-th moment, namely, $\mathbb{Q} \in \mathcal{P}_p(\mathcal{Z})$ if and only if its expectation $\mathbb{E}_{z \sim \mathbb{Q}}[\|z\|^p] < \infty$. Whenever it is clear from the context, we will write $\mathbb{E}_{z \sim \mathbb{P}}$ as $\mathbb{E}_{\mathbb{P}}$ and omit the random variable inside the expectation. The support of a distribution is denoted by $\text{supp}\,\mathbb{Q}$. The $\mathcal{L}^p(\mathbb{Q})$-norm of a $\mathbb{Q}$-measurable function $h$ is denoted by $\|h\|_{\mathbb{Q}, p} = \mathbb{E}_{z \sim \mathbb{Q}}[|h(z)|^p]^{1/p}$. The sup-norm of a function $h$ is denoted by $\|h\|_\infty$, and the Lipschitz norm of a Lipschitz continuous function $h$ is denoted by $\|h\|_{\text{Lip}}$. We denote $a \vee b = \max(a, b)$ and $a \wedge b = \min(a, b)$. For the expectation operator $\mathbb{E}_{z \sim \mathbb{Q}}[\cdot]$, we often write it as $\mathbb{E}_{\mathbb{Q}}[\cdot]$ provided that the involved random variable is clear from the context.

The *Wasserstein distance* of order $p$ between distributions $\mathbb{P}, \mathbb{Q} \in \mathcal{P}_p(\mathcal{Z})$ is defined via

$$\mathcal{W}_p(\mathbb{P}, \mathbb{Q})^p := \inf_{\pi \in \mathcal{P}_p(\mathcal{Z}^2)} \left\{ \mathbb{E}_{(\tilde{z}, z) \sim \pi}[\|\tilde{z} - z\|^p] : \pi \text{ has marginal distributions } \mathbb{P}, \mathbb{Q} \right\}.$$

We denote by $\mathcal{F} := \{f_\theta : \theta \in \Theta\}$ the class of loss functions. To ease notations, we often suppress the subscript $\theta$ and use $f$ to represent a generic loss function from $\mathcal{F}$. Given a nominal distribution $\mathbb{Q} \in \mathcal{P}_p(\mathcal{Z})$ and a radius $\rho \geq 0$, Wasserstein DRO problem is given by

$$\inf_{f \in \mathcal{F}} \sup_{\mathbb{P} \in \mathcal{P}_p(\mathcal{Z})} \{\mathbb{E}_{z \sim \mathbb{P}}[f(z)] : \mathcal{W}_p(\mathbb{P}, \mathbb{Q}) \leq \rho\}.$$

Suppose there exists $M, L \geq 0$ such that $f(z) \leq M + L\|z\|^p$ for all $z \in \mathcal{Z}$, then the inner maximization problem above has a dual problem that always has a minimizer [40]:

$$\min_{\lambda \geq 0} \left\{\lambda \rho^p + \mathbb{E}_{z \sim \mathbb{Q}}\left[\sup_{\tilde{z} \in \mathcal{Z}} \{f(\tilde{z}) - \lambda \|\tilde{z} - z\|^p\}\right]\right\}. \tag{1}$$

In a data-driven problem, the nominal distribution is often chosen as the empirical distribution $\mathbb{P}_n = \frac{1}{n} \sum_{i=1}^n \boldsymbol{\delta}_{z_i^n}$ constructed from $n$ i.i.d. samples $\{z_i^n\}_{i=1}^n$ from the underlying true distribution $\mathbb{P}_{\text{true}}$, where $\boldsymbol{\delta}_z$ denotes the Dirac point mass on $z$. We use $\mathbb{P}_\otimes$ or $\mathbb{E}_\otimes$ to indicate that the probability or expectation is evaluated with respect to the sampling distribution, namely the $n$-fold product distribution $\otimes_{i=1}^n \mathbb{P}_{\text{true}}$ over $\mathcal{Z}^n$.

We define the *Wasserstein regularizer* as the difference between the Wasserstein robust loss and the nominal loss:

$$\mathcal{R}_{\mathbb{Q},p}(\rho; f) := \sup_{\mathbb{P} \in \mathcal{P}_p(\mathcal{Z})} \{\mathbb{E}_{\mathbb{P}}[f] : \mathcal{W}_p(\mathbb{P}, \mathbb{Q}) \leq \rho\} - \mathbb{E}_{\mathbb{Q}}[f].$$

The connection between Wasserstein DRO and regularization has been established under various settings [36, 73, 14, 74, 39, 87, 5]. The next two results adapted from Gao et al. [39] (see also Bartl et al. [5]) establish connections between the Wasserstein regularizer $\mathcal{R}_{\mathbb{Q},p}$ and Lipschitz regularization ($p = 1$) and gradient regularization ($p = 2$) respectively. For completeness we provide their proofs in Appendix A.

ASSUMPTION 1. *Assume the following holds:*

(I) *There exists $\gamma_1 > 0$ such that for every $f \in \mathcal{F}$,*

$$f(\tilde{z}) - f(z) \leq \gamma_1 \|\tilde{z} - z\|, \quad \forall z, \tilde{z} \in \mathcal{Z}.$$

(II) *Suppose $\operatorname{diam}(\mathcal{Z}) := \sup_{\tilde{z}, z \in \mathcal{Z}} \|\tilde{z} - z\| = \infty$, and for every $f \in \mathcal{F}$, there exists $z_0 \in \mathcal{Z}$ such that*

$$\limsup_{\|z - z_0\| \to \infty} \frac{f(z) - f(z_0)}{\|z - z_0\|} = \|f\|_{\operatorname{Lip}}.$$

Assumption (I) means that every $f$ is Lipschitz continuous, and (II) means that the Lipschitz norm is attained at infinity.

LEMMA 1 (**Lipschitz regularization**). *Let $\mathbb{Q} \in \mathcal{P}_1(\mathcal{Z})$ and $\rho \geq 0$. Assume Assumption 1(I) holds, then*

$$\mathcal{R}_{\mathbb{Q},1}(\rho; f) \leq \rho \cdot \|f\|_{\operatorname{Lip}}.$$

*Assume, in addition, Assumption 1(II) holds. Then*

$$\mathcal{R}_{\mathbb{Q},1}(\rho; f) = \rho \cdot \|f\|_{\operatorname{Lip}}.$$

ASSUMPTION 2. *Assume every $f \in \mathcal{F}$ is differentiable and there exist $\hbar > 0$ such that*

$$\|\nabla f(\tilde{z}) - \nabla f(z)\|_* \leq \hbar \|\tilde{z} - z\|, \quad \forall \tilde{z}, z \in \mathcal{Z}, \ \forall f \in \mathcal{F}.$$

This is a smoothness condition which requires that every $f$ has Lipschitz gradient.

LEMMA 2 (**Gradient regularization**). *Let $\mathbb{Q} \in \mathcal{P}_2(\mathcal{Z})$ and $\rho \geq 0$. Assume Assumption 2 holds. Then*

$$\left|\mathcal{R}_{\mathbb{Q},2}(\rho; f) - \rho \cdot \|\|\nabla f\|_*\|_{\mathbb{Q},2}\right| \leq \hbar \rho^2.$$

## 3. Variation-based Concentration Inequality

In this section, we derive a large-deviation type concentration inequality for the empirical mean of a single loss function. We derive an equivalent representation of $\mathcal{R}^{-1}_{\mathbb{P}_{\text{true}},p}(\cdot;f)$ in Section 3.1 and provide a brief overview of transportation-information inequalities in Section 3.2. The new concentration inequality is developed in Section 3.3, whose proof is postponed to Section B.1.

### 3.1. Inverse of the Wasserstein Regularizer

Fixing $f \in \mathcal{F}$, we define a function $\mathcal{I}_p : \mathbb{R}_+ \to \mathbb{R}_+ \cup \{+\infty\}$ via

$$\mathcal{I}_p(\varepsilon;f)^p := \sup_{t>0}\left\{\varepsilon t - \mathbb{E}_{z\sim\mathbb{P}_{\text{true}}}\left[\sup_{\tilde{z}\in\mathcal{Z}}\left\{t\big(f(\tilde{z}) - f(z)\big) - \|\tilde{z}-z\|^p\right\}\right]\right\},$$

which will play a similar role as the rate function in the large deviation principle. The next proposition establishes its connection to the Wasserstein regularizer $\mathcal{R}_{\mathbb{P}_{\text{true}},p}$, whose proof is given in Appendix B.

PROPOSITION 1. *Let $p \in [1,\infty)$ and $f \in \mathcal{F}$. Suppose there exists $M, L \geq 0$ such that $f(z) \leq M + L\|z\|^p$ for all $z \in \mathcal{Z}$. Let $\rho > 0$ and suppose the dual minimizer $\lambda_o$ of (1) is positive. Set $\underline{\lambda} := \lim_{\|z\|\to\infty} f(z)/\|z\|^p$. Then*

$$\mathcal{I}_p(\mathcal{R}_{\mathbb{P}_{\text{true}},p}(\rho;f);f) \begin{cases} = \rho, & \text{if } \lambda_o > \underline{\lambda}, \\ \geq \rho, & \text{if } \lambda_o = \underline{\lambda}. \end{cases}$$

Note that the dual optimizer of (1) tends to be large when $\rho$ is close to zero, in which case $\lambda_o > \underline{\lambda}$, as observed in [40]. Hence Proposition 1 shows that at least for small $\rho$, the left inverse of $\mathcal{R}_{\mathbb{P}_{\text{true}},p}(\cdot;f)$ is precisely $\mathcal{I}_p(\cdot;f)$.

### 3.2. Transportation-Information Inequalities

Just like many other results on the concentration of measure, appropriate conditions on the function $f$ and the distribution $\mathbb{P}_{\text{true}}$ are required. Since we are dealing with general loss functions that are possibly unbounded, some assumptions on the underlying data-generating distribution are necessary. It turns out for our purpose, it is convenient to work with the *transportation-information inequality*, a useful condition to establish concentration of measure in modern probability theory.

DEFINITION 1 (TRANSPORTATION-INFORMATION INEQUALITY). *Let $p \in [1,\infty)$. A distribution $\mathbb{P} \in \mathcal{P}_p(\mathcal{Z})$ satisfies a transportation-information inequality $T_p(\tau)$ for some positive constant $\tau$, if*

$$\mathcal{W}_p(\mathbb{Q},\mathbb{P}) \leq \sqrt{\tau H(\mathbb{Q}\|\mathbb{P})}, \quad \forall \mathbb{Q} \in \mathcal{P}_p(\mathcal{Z}),$$

*where $H(\mathbb{Q}\|\mathbb{P})$ denotes the relative entropy $H(\mathbb{Q}\|\mathbb{P}) := \int_{\mathcal{Z}} \log(d\mathbb{Q}/d\mathbb{P})\,d\mathbb{Q}$, where $d\mathbb{Q}/d\mathbb{P}$ denotes the Radon-Nikodym derivative.*

We briefly comment on distributions satisfying transportation-information inequalities, and refer the reader to [45] for a recent survey and [85, Chapter 22] for an in-depth discussion. Among different choices of $p$, $T_1$ and $T_2$ are of particular interest and have been widely studied in the literature. $T_1$ is equivalent to the following condition (Theorem 2.3 in [30] and Theorem 22.10 in [85]; see also Lemma 3 in Appendix B): a distribution $\mathbb{P}$ satisfies $T_1$ if and only if there exists $a > 0$ such that $\mathbb{E}[\exp(a\|z\|^2)] < \infty$. In particular, any distribution on a bounded support $\mathcal{Z}$ with $\text{diam}(\mathcal{Z}) < \infty$ satisfies $T_1(2\text{diam}(\mathcal{Z})^2)$. $T_2$ is also known as Talagrand's inequality, which is implied by the log-Sobolev inequality. Examples of distributions satisfying the log-Sobolev inequality include distributions with a strongly log-concave density, and mixture of distributions satisfying log-Sobolev inequality whose pairwise chi-squared divergences are uniformly bounded [23]. Note that for $p_1 \leq p_2$, $T_{p_1}$ is weaker than $T_{p_2}$ since $\mathcal{W}_{p_1} \leq \mathcal{W}_{p_2}$. In the sequel, we will focus on the case $p \in [1,2]$.

### 3.3. Concentration for a Single Loss Function

Now we are ready to state our main result in this section.

THEOREM 1 **(Variation-based concentration)**. *Let $p \in [1,2]$ and $f \in \mathcal{F}$. Assume there exist $M, L > 0$ such that*
$$f(z) \leq M + L\|z\|^p, \quad \forall z \in \mathcal{Z}.$$
*Assume further that $\mathbb{P}_{\text{true}}$ satisfies $T_p(\tau)$ for some $\tau > 0$. Let $\varepsilon > 0$. Then*
$$\mathbb{P}_\otimes \{\mathbb{E}_{\mathbb{P}_n}[f] - \mathbb{E}_{\mathbb{P}_{\text{true}}}[f] < -\varepsilon\} \leq \exp\left(-n\mathcal{I}_p(\varepsilon; -f)^2/\tau\right).$$
*Let $t > 0$. Then with probability at least $1 - e^{-t}$,*
$$\mathbb{E}_{\mathbb{P}_{\text{true}}}[f] \leq \mathbb{E}_{\mathbb{P}_n}[f] + \mathcal{R}_{\mathbb{P}_{\text{true}}, p}\left(\sqrt{\tfrac{\tau t}{n}}; -f\right). \tag{2}$$

Theorem 1 uncovers an interesting connection: the non-asymptotic decay rate of large deviation probabilities is controlled by the inverse of Wasserstein regularizer $\mathcal{I}_p(\varepsilon; -f)$. The negative sign $-f$ appears because here we bound the downside risk, i.e., the probability of empirical loss being smaller than the true loss, whereas $\mathcal{R}_{\mathbb{P}_{\text{true}}, p}$ is defined via upside excess, i.e., the worst-case loss that is greater than the true loss. As a matter of fact, a similar result holds if we swap the empirical loss and true loss in the theorem:
$$\mathbb{P}_\otimes \{\mathbb{E}_{\mathbb{P}_n}[f] - \mathbb{E}_{\mathbb{P}_{\text{true}}}[f] > \varepsilon\} \leq \exp\left(-n\mathcal{I}_p(\varepsilon; f)^2/\tau\right),$$
and with probability at least $1 - e^{-t}$,
$$\mathbb{E}_{\mathbb{P}_n}[f] \leq \mathbb{E}_{\mathbb{P}_{\text{true}}}[f] + \mathcal{R}_{\mathbb{P}_{\text{true}}, p}\left(\sqrt{\tfrac{\tau t}{n}}; f\right).$$

When $p = 1$, if $f$ is Lipschitz continuous, then by Lemma 1 we have
$$\mathcal{R}_{\mathbb{P}_{\text{true}}, p}\left(\sqrt{\tfrac{\tau t}{n}}; -f\right) \leq \sqrt{\tfrac{\tau t}{n}} \cdot \|-f\|_{\text{Lip}} = \sqrt{\tfrac{\tau t}{n}} \cdot \|f\|_{\text{Lip}}.$$

When $p = 2$, if $f$ has Lipschitz continuous gradient, then by Lemma 2 we have
$$\mathcal{R}_{\mathbb{P}_{\text{true}}, 2}\left(\sqrt{\tfrac{\tau t}{n}}; -f\right) \leq \sqrt{\tfrac{\tau t}{n}} \cdot \|\|\nabla f\|_*\|_{\mathbb{P}_{\text{true}}, 2} + \frac{\hbar \tau t}{n}.$$

Substituting these inequalities in Theorem 1 yields following corollary.

COROLLARY 1 **(Variation regularization)**. *Let $p \in \{1, 2\}$. When $p = 1$, assume Assumption 1(I) holds; when $p = 2$, assume Assumption 2 holds. Assume further that $\mathbb{P}_{\text{true}}$ satisfies $T_p(\tau)$ for some $\tau > 0$. Let $t > 0$. Then with probability at least $1 - e^{-t}$,*
$$\mathbb{E}_{\mathbb{P}_{\text{true}}}[f] \leq \mathbb{E}_{\mathbb{P}_n}[f] + \begin{cases} \sqrt{\tfrac{\tau t}{n}} \cdot \|f\|_{\text{Lip}}, & p = 1, \\ \sqrt{\tfrac{\tau t}{n}} \cdot \|\|\nabla f\|_*\|_{\mathbb{P}_{\text{true}}, 2} + \tfrac{\hbar \tau t}{n}, & p = 2. \end{cases}$$

Theorem 1 and Corollary 1 show that the Wasserstein regularizer $\mathcal{R}_{\mathbb{P}_{\text{true}}, p}(\sqrt{\tfrac{\tau t}{n}}; -f)$, as well as the variation of the loss, $\|f\|_{\text{Lip}}$ or $\|\|\nabla f\|_*\|_{\mathbb{P}_{\text{true}}, 2}$, are natural quantities controlling the deviation of the empirical loss for distributions satisfying a transportation-information inequality. For $p = 1$, thanks to the first part of Lemma 1, the bound in Theorem 1 is tighter than the Lipschitz norm bound in Corollary 1, which was obtained in [18]. Since $\|\|\nabla f\|_*\|_{\mathbb{P}_{\text{true}}, 2} \leq \|f\|_{\text{Lip}}$, $p = 2$ suggests a tighter upper bound than $p = 1$, at the cost of a stronger assumption on the underlying distribution.

## 4. Finite-Sample Guarantees

In the previous section, we have derived a concentration inequality for a single loss function, and the goal of this section is to extend it to a family of loss functions $\mathcal{F}$. In the spirit of [36, 14, 74], we would like to determine a proper scaling of the Wasserstein radius $\rho_n$ with respect to sample size $n$ so that with high probability, the Wasserstein robust loss is an upper bound of the true loss uniformly for all functions in the class $\mathcal{F} = \{f_\theta : \theta \in \Theta\}$. Whenever this holds, minimizing the Wasserstein robust loss controls the true loss as well.

When $\mathcal{F}$ is a finite set, then a simple application of the union bound to Theorem 1 yields that (2) holds simultaneously for all $f \in \mathcal{F}$ with probability at least $1 - |\mathcal{F}|e^{-t}$, where $|\cdot|$ denotes the cardinality of a set. When $\mathcal{F}$ contains infinitely many functions, some notion of complexity of the function class $\mathcal{F}$ is needed to obtain uniform convergence. In Section 4.1, we prove the result using a standard covering number argument; and in Section 4.2, we adopt techniques from local Rademacher complexity theory [6, 52].

### 4.1. Covering Number Arguments

Recall that for $\epsilon > 0$, the *covering number* $\mathcal{N}(\epsilon; \mathcal{H}, \|\cdot\|_\mathcal{H})$ of a set $\mathcal{H}$ with respect to a norm $\|\cdot\|_\mathcal{H}$ is defined as the smallest cardinality of an $\epsilon$-cover of $\mathcal{H}$, where $\mathcal{H}_\epsilon$ is an $\epsilon$-cover of $\mathcal{H}$ if for each $h \in \mathcal{H}$, there exists $\tilde{h} \in \mathcal{H}_\epsilon$ such that $\|\tilde{h} - h\|_\mathcal{H} \leq \epsilon$. Similar to the classic stochastic programming literature (e.g., Shapiro et al. [78, Section 5.3.2]), we can obtain a union bound using the standard covering number argument, whose proof is given in Appendix C.1. Throughout this subsection, we let $\mathcal{F} = \{f_\theta : \theta \in \Theta\}$ and we impose the following smoothness assumption with respect to the parameter $\theta$.

ASSUMPTION 3. *Assume there exists a measurable function $\kappa : \mathcal{Z} \to \mathbb{R}_+$ and constants $\kappa_M, \kappa_L \geq 0$ satisfying $\kappa(z) \leq \kappa_M + \kappa_L \|z\|^p$ for all $z \in \mathcal{Z}$, such that*

$$|f_{\tilde\theta}(z) - f_\theta(z)| \leq \kappa(z)\|\tilde\theta - \theta\|_\Theta, \quad \forall \theta, \tilde\theta \in \Theta, \ \mathbb{P}_{\text{true}} - a.e.\ z \in \mathcal{Z}.$$

COROLLARY 2. *Let $p \in [1, 2]$. Assume $\mathbb{P}_{\text{true}}$ satisfies $T_p(\tau)$ for some $\tau > 0$ and Assumption 3 holds. Let $t > 0$. Set*

$$\rho_n = \sqrt{\frac{\tau(t + \log \mathcal{N}(1/n; \Theta, \|\cdot\|_\Theta))}{n}},$$

*and $\epsilon_n = (2\mathbb{E}_{\mathbb{P}_{\text{true}}}[\kappa] + \sqrt{\mathbb{V}\text{ar}_{\mathbb{P}_{\text{true}}}[\kappa]} + \kappa_M + 2^{p-1}\kappa_L((\tau t/n)^{p/2} + \mathbb{E}_{\mathbb{P}_{\text{true}}}[\|z\|^p]))/n$. Then we have the following:*

(I) *Assume there exist constants $M, L > 0$ such that*

$$f_\theta(z) \leq M + L\|z\|^p, \quad \forall z \in \mathcal{Z}, \ \forall \theta \in \Theta.$$

*Then with probability at least $1 - 1/n - e^{-t}$,*

$$\mathbb{E}_{\mathbb{P}_{\text{true}}}[f_\theta] \leq \mathbb{E}_{\mathbb{P}_n}[f_\theta] + \mathcal{R}_{\mathbb{P}_{\text{true}}, p}(\rho_n; -f_\theta) + \epsilon_n, \quad \forall \theta \in \Theta.$$

(II) **(Lipschitz regularization and 1-Wasserstein DRO)** *When $p = 1$, assume Assumption 1(I) holds. Then with probability at least $1 - 1/n - e^{-t}$,*

$$\mathbb{E}_{\mathbb{P}_{\text{true}}}[f_\theta] \leq \mathbb{E}_{\mathbb{P}_n}[f_\theta] + \rho_n \cdot \|f_\theta\|_{\text{Lip}} + \epsilon_n, \quad \forall \theta \in \Theta.$$

*Assume, in addition, that Assumption 1(II) holds. Then with probability at least $1 - 1/n - e^{-t}$,*

$$\mathbb{E}_{\mathbb{P}_{\text{true}}}[f_\theta] \leq \mathbb{E}_{\mathbb{P}_n}[f_\theta] + \mathcal{R}_{\mathbb{P}_n, 1}(\rho_n; f_\theta) + \epsilon_n, \quad \forall \theta \in \Theta.$$

(III) **(Gradient regularization)** When $p = 2$, assume Assumption 2 holds. Then with probability at least $1 - 1/n - e^{-t}$,

$$\mathbb{E}_{\mathbb{P}_{\text{true}}}[f_\theta] \leq \mathbb{E}_{\mathbb{P}_n}[f_\theta] + \rho_n \cdot \|\|\nabla f_\theta\|_*\|_{\mathbb{P}_{\text{true}},2} + \hbar \rho_n^2 + \epsilon_n, \quad \forall \theta \in \Theta.$$

Corollary 2 establishes the generalization bounds for Wasserstein DRO as well as the Lipschitz and gradient regularization. By [89, Example 5.8], $\log \mathcal{N}(\epsilon; \Theta, \|\cdot\|_*) \leq d \log(1 + 2B/\epsilon)$, where $B$ is the diameter of $\Theta$. Thereby, by choosing the radius $\rho_n = \tilde{O}(\sqrt{d/n})$, the Wasserstein robust loss serves as an upper bound of the true loss for all $f_\theta \in \mathcal{F}$ up to an $O(1/n)$ remainder for $p = 1$ and an $O(d/n)$ remainder for $p = 2$. Assumption 1(II) may be restrictive for certain applications. In Example 1, we demonstrate an instance for which finite-sample guarantee holds with an-$O(1/\sqrt{n})$ radius even though this assumption does not hold. Essentially, as long as the Wasserstein regularizer can be sandwiched by multiples of Lipschitz regularizer (see Theorem 2 in [39]), similar finite-sample guarantees can be obtained; we refer to a follow-up work [2] for more in-depth discussions.

In the next result, we provide an empirical counterpart of Corollary 2 for $p = 2$.

ASSUMPTION 4. *Assume there exists a measurable function $\kappa_2 : \mathcal{Z} \to \mathbb{R}_+$ and constants $\kappa_{2,M}, \kappa_{2,L} \geq 0$ satisfying $\kappa_2(z) \leq \kappa_{2,M} + \kappa_{2,L} \|z\|^p$ for all $z \in \mathcal{Z}$, such that*

$$\|\nabla f_{\tilde\theta}(z) - \nabla f_\theta(z)\|_* \leq \kappa_2(z) \|\tilde\theta - \theta\|_\Theta, \quad \forall \theta, \tilde\theta \in \Theta, \ \mathbb{P}_{\text{true}} - a.e. \ z \in \mathcal{Z}.$$

COROLLARY 3 (**2-Wasserstein DRO**). *Assume $\mathbb{P}_{\text{true}}$ satisfies $T_2(\tau)$ for some $\tau > 0$ and Assumptions 2, 3, 4 hold. Assume $\sigma = \sup_{\theta \in \Theta} \mathbb{E}_{\mathbb{P}_{\text{true}}}[\|\nabla f_\theta\|_*^4]^{\frac{1}{2}} / \|\|\nabla f_\theta\|_*\|_{\mathbb{P}_{\text{true}},2}^2 < \infty$. Let $t > 0$ and $n > 8\sigma^2 t$. Set*

$$\rho_n = \sqrt{\frac{\tau(t + \log(1 + \mathcal{N}(1/n; \Theta, \|\cdot\|_\Theta)))}{n}} \left(1 + \sigma \sqrt{\frac{2(t + \log(1 + \mathcal{N}(1/n; \Theta, \|\cdot\|_\Theta)))}{n}}\right),$$

*and $\tilde\epsilon_n = (2\mathbb{E}_{\mathbb{P}_{\text{true}}}[\kappa_2] + \sqrt{\mathbb{V}\text{ar}_{\mathbb{P}_{\text{true}}}[\kappa_2]} + \rho_n \sqrt{\mathbb{E}_{\mathbb{P}_{\text{true}}}[\kappa_2^2] + \sqrt{\mathbb{V}\text{ar}_{\mathbb{P}_{\text{true}}}[\kappa_2^2]}})/n$. Then with probability at least $1 - 2/n - 2e^{-t}$,*

$$\mathbb{E}_{\mathbb{P}_{\text{true}}}[f_\theta] \leq \mathbb{E}_{\mathbb{P}_n}[f_\theta] + \rho_n \|\|\nabla f_\theta\|_*\|_{\mathbb{P}_n,2} + \tilde\epsilon_n + \frac{\hbar \tau(t + \log(1 + \mathcal{N}(1/n; \Theta, \|\cdot\|_\Theta)))}{n}, \quad \forall \theta \in \Theta,$$

*and*

$$\mathbb{E}_{\mathbb{P}_{\text{true}}}[f_\theta] \leq \mathbb{E}_{\mathbb{P}_n}[f_\theta] + \mathcal{R}_{\mathbb{P}_n,2}(\rho_n; f_\theta) + \tilde\epsilon_n + \frac{2\hbar \tau(t + \log(1 + \mathcal{N}(1/n; \Theta, \|\cdot\|_\Theta)))}{n}, \quad \forall \theta \in \Theta.$$

In Section 5, we will demonstrate Corollary 2 for feature-based newsvendor problem in Example 1 and Corollary 3 for linear prediction with Lipschitz loss in Example 3.

### 4.2. Local Rademacher Complexity Arguments

The covering number bound developed in the previous subsection may be loose. Indeed, the discussion after Corollary 2 indicates that for smooth parametric family the radius $\tilde{O}(\sqrt{d/n})$ is always dimension-dependent. To obtain a tighter bound in a more general setting, we derive results using local Rademacher complexity theory. As it turns out, this approach sometimes leads to a bound that has a better dependence on $d$ and even dimension-independent bound; see Examples 2, 4, 5.

Let us begin with some technical preparation. Recall the *Rademacher complexity* of a function class $\mathcal{F}$ with respect to a sample $\{z_i^n\}_{i=1}^n$ is defined as

$$\mathfrak{R}_n(\mathcal{F}) := \mathbb{E}_\sigma \left[\sup_{f \in \mathcal{F}} \frac{1}{n} \sum_{i=1}^n \sigma_i f(z_i^n)\right],$$

where $\sigma_i$'s are i.i.d. Rademacher random variables with $\mathbb{P}\{\sigma_i = \pm 1\} = \frac{1}{2}$. The Rademacher complexity of the function class $\mathcal{F}$ with respect to $\mathbb{P}_{\text{true}}$ for sample size $n$ is defined as $\mathbb{E}_\otimes[\mathfrak{R}_n(\mathcal{F})]$. Rademacher complexity plays an important role in bounding the generalization error of statistical learning problems but may be vacuous if $\mathcal{F}$ is too large. The idea of localization is to restrict on a small subset of $\mathcal{F}$ around the optimal solution that often admits low complexity. The *localized Rademacher complexity* [6] at level $r > 0$ is defined as

$$\mathbb{E}_\otimes\Big[\mathfrak{R}_n(\{cf :\ f \in \mathcal{F},\ 0 \le c \le 1,\ T(cf) \le r\})\Big],$$

where $T : \mathcal{F} \to \mathbb{R}_+$. In our analysis, we choose $T(f) = \|f\|_{\text{Lip}}^2$ when $p = 1$ and $T(f) = \|\|\nabla f\|_*\|_{\mathbb{P}_{\text{true}},2}^2$ when $p = 2$. Part of our techniques below are adapted from the framework developed in [6, 34], which primarily considers $T(f) = \mathbb{E}_{\mathbb{P}_{\text{true}}}[f^2]$.

By choosing a proper level $r_n$, the localized Rademacher complexity of the functions of the subset can be much smaller than the entire family, which enables a better bound. Often, the level $r_n$ is chosen to be the fixed point $r_{n\star}$ of some function $\psi_n(r)$, which serves as an upper bound on the localized Rademacher complexity at level $r$. A typical assumption imposed on $\psi_n$ is the so-called sub-root condition. A function $\psi : \mathbb{R}_+ \to \mathbb{R}_+$ is *sub-root* if it is non-constant, non-negative, non-decreasing and the map $r \mapsto \psi(r)/\sqrt{r}$ is non-increasing for all $r > 0$. A sub-root function always has a unique fixed point $r_{n\star}$ [6]. Similar to the literature, we impose the following assumption.

ASSUMPTION 5 (Sub-root local complexity). *Assume there exists a sub-root function $\psi_n : \mathbb{R}_+ \to \mathbb{R}_+$ such that*

$$\psi_n(r) \ge \mathbb{E}_\otimes\Big[\mathfrak{R}_n(\{cf :\ f \in \mathcal{F},\ 0 \le c \le 1,\ T(cf) \le r\})\Big].$$

*Denote by $r_{n\star}$ the fixed point of $\psi_n$.*

We will verify this assumption for various examples considered in Section 5.

We first study the case of $p = 1$. The proof is given in Appendix C.2.2.

THEOREM 2 (**Lipschitz regularization**). *Assume $\mathbb{P}_{\text{true}}$ satisfies $T_1(\tau)$, Assumption 1(I) holds, and Assumption 5 holds with $T(f) = \|f\|_{\text{Lip}}^2$. Let $t > 0$. Then with probability at least $1 - \lceil \log_2(\sqrt{\gamma_1 \tau t n}) \rceil e^{-t}$,*

$$\mathbb{E}_{\mathbb{P}_{\text{true}}}[f] \le \mathbb{E}_{\mathbb{P}_n}[f] + \left(2\sqrt{\frac{\tau t}{n}} + \sqrt{4r_{n\star} + \frac{2}{n}}\right)\|f\|_{\text{Lip}} + 4r_{n\star} + \frac{2}{n}, \quad \forall f \in \mathcal{F}.$$

Together with Lemma 1, we obtain the following result.

COROLLARY 4 (**1-Wasserstein DRO**). *Assume $\mathbb{P}_{\text{true}}$ satisfies $T_1(\tau)$, Assumption 1 holds and Assumption 5 holds with $T(f) = \|f\|_{\text{Lip}}^2$. Let $t > 0$. Set*

$$\rho_n = 2\sqrt{\frac{\tau t}{n}} + \sqrt{4r_{n\star} + \frac{2}{n}}.$$

*Then with probability at least $1 - \lceil \log_2(\sqrt{\gamma_1 \tau t n}) \rceil e^{-t}$,*

$$\mathbb{E}_{\mathbb{P}_{\text{true}}}[f] \le \mathbb{E}_{\mathbb{P}_n}[f] + \mathcal{R}_{\mathbb{P}_n,1}(\rho_n; f) + 4r_{n\star} + \frac{2}{n}, \quad \forall f \in \mathcal{F}.$$

Note that $\rho_n = \tilde{O}(1/\sqrt{n})$ if $r_{n\star} = \tilde{O}(\frac{1}{n})$. A sufficient condition for this to hold is the sub-root function $\psi_n(r) = \tilde{O}(\sqrt{r/n})$, which holds for many important cases as we illustrated in Sections 5. As such, Theorem 2 and Corollary 4 show that by choosing a radius in the order of $1/\sqrt{n}$, with high probability, the Wasserstein robust loss serves as a upper bound for the true loss up to an $\tilde{O}(1/n)$ gap. Here the probability bound has a $O(\log n)$ term, nearly independent of sample size $n$. By mapping $t$ to $t + \log\lceil\log_2(\sqrt{\gamma_1 \tau t n})\rceil$, one can obtain a probability bound that is independent of sample size, while the radius $\rho_n \sim O(\sqrt{\log \log n / n})$. In the rest of the paper, we will not make such a transformation, but just keep in mind that these two results are equivalent.

In the next corollary, we consider the loss functions of a composition form $\ell \circ f$, where $\ell : \mathbb{R} \to \mathbb{R}$ is a given Lipschitz function and $f \in \mathcal{F}$, which occurs often in supervised learning. The following result is useful to establish the generalization bound for problems of this type.

COROLLARY 5 (**Lipschitz composition**). *Assume $\mathbb{P}_{\text{true}}$ satisfies $T_1(\tau)$, Assumption 1(I) holds, and Assumption 5 holds with $T(f) = \|f\|_{\text{Lip}}^2$. Let $\ell$ be an $L_\ell$-Lipschitz function and $t > 0$. Then with probability at least $1 - \lceil \log_2(\sqrt{L_\ell \gamma_1 \tau t n}) \rceil e^{-t}$,*

$$\mathbb{E}_{\mathbb{P}_{\text{true}}}[\ell \circ f] \leq \mathbb{E}_{\mathbb{P}_n}[\ell \circ f] + 2\left(\sqrt{\frac{\tau t}{n}} L_\ell + \sqrt{4 L_\ell^2 r_{n\star} + \frac{2 L_\ell}{n}}\right) \|f\|_{\text{Lip}} + 4 L_\ell^2 r_{n\star} + \frac{2 L_\ell}{n}, \quad \forall f \in \mathcal{F}.$$

In Section 5, we will illustrate Corollary 5 in supervised learning with linear class (Example 2) and with nonlinear kernel class (Example 5).

The analysis for 2-Wasserstein DRO is aligned with the previous case but requires more care to deal with the data-dependent regularization $\|\|\nabla f\|_*\|_{\mathbb{P}_n, 2}$; see details in Appendix C.2.3. Define the family of normalized gradient norm functions

$$\mathcal{G} := \left\{ \frac{\|\nabla f\|_*^2}{\|\|\nabla f\|_*\|_{\mathbb{P}_{\text{true}}, 2}^2} : f \in \mathcal{F} \right\}.$$

THEOREM 3 (**Gradient regularization**). *Assume that $\mathbb{P}_{\text{true}}$ satisfies $T_2(\tau)$, Assumption 2 holds, and Assumption 5 holds with $T(f) = \|\|\nabla f\|_*\|_{\mathbb{P}_{\text{true}}, 2}^2$. Assume there exists $\gamma_2 > 0$ such that $\|\|\nabla f\|_*\|_{\mathbb{P}_{\text{true}}, 2} \leq \gamma_2$ for all $f \in \mathcal{F}$. Let $t > 0$. Set*

$$\rho_n = 2\sqrt{\frac{\tau t}{n}}(1 + \mathbb{E}_\otimes[\mathfrak{R}_n(\mathcal{G})]) + \sqrt{4 r_{n\star} + 2 \epsilon_n},$$

*and*

$$\epsilon_n = \frac{\hbar \tau t + 1 + \mathbb{E}_\otimes[\mathfrak{R}_n(\mathcal{G})]}{n}.$$

*Then with probability at least $1 - \lceil \log_2(\sqrt{\gamma_2 \tau t n}) \rceil e^{-t}$,*

$$\mathbb{E}_{\mathbb{P}_{\text{true}}}[f] \leq \mathbb{E}_{\mathbb{P}_n}[f] + \rho_n \|\|\nabla f\|_*\|_{\mathbb{P}_{\text{true}}, 2} + 4 r_{n\star} + 2 \epsilon_n, \quad \forall f \in \mathcal{F}.$$

Whenever $\mathbb{E}_\otimes[\mathfrak{R}_n(\mathcal{G})] = O(1)$ and $r_{n\star} = O(1/n)$, Theorem 3 shows that by choosing a radius in the order of $1/\sqrt{n}$, with high probability, the gradient regularized loss serves as an upper bound for the true loss up to an $O(1/n)$ gap. Note that $\mathbb{E}_\otimes[\mathfrak{R}_n(\mathcal{G})] = O(1/\sqrt{n})$ as long as $\mathcal{G}$ has finite VC dimension (see, e.g., Lemma 4.14 and Proposition 4.18 in [89]). In Section 5, we will illustrate this result in portfolio optimization (Example 4) and neural networks (Example 6) which show that $\mathbb{E}_\otimes[\mathfrak{R}_n(\mathcal{G})] = O(1/\sqrt{n})$ and $r_{n\star} = O(1/n)$ with explicit constants.

The distribution-dependent gradient regularization $\|\|\nabla f\|_*\|_{\mathbb{P}_{\text{true}}, 2}$ in Theorem 3 can be replaced with its empirical counterpart $\|\|\nabla f\|_*\|_{\mathbb{P}_n, 2}$. In the next result, we provide the generalization bound for data-dependent gradient regularization problems and 2-Wasserstein DRO.

COROLLARY 6 (**2-Wasserstein DRO**). *Under the setting in Theorem 3, assume additionally that there exists $\kappa_g > 0$ such that $\frac{\|\nabla f(z)\|_*}{\|\|\nabla f\|_*\|_{\mathbb{P}_{\text{true}}, 2}} \leq \kappa_g$ for all $f \in \mathcal{F}$ and $z \in \mathcal{Z}$. Set*

$$\tilde{\rho}_n = \rho_n \left(1 + 2\mathbb{E}_\otimes[\mathfrak{R}_n(\mathcal{G})] + \kappa_g^2 \sqrt{\frac{t}{2n}}\right).$$

*Then whenever* $2\mathbb{E}_\otimes[\mathfrak{R}_n(\mathcal{G})] + \kappa_g^2 \sqrt{\frac{t}{2n}} < 1/2$, *with probability at least* $1 - (\lceil \log_2(\sqrt{\gamma_2 \tau t n}) \rceil + 1)e^{-t}$, *for every* $f \in \mathcal{F}$,

$$\mathbb{E}_{\mathbb{P}_{\text{true}}}[f] \leq \mathbb{E}_{\mathbb{P}_n}[f] + \tilde{\rho}_n \| \|\nabla f\|_* \|_{\mathbb{P}_n, 2} + 4r_{n\star} + 2\epsilon_n,$$

*and*

$$\mathbb{E}_{\mathbb{P}_{\text{true}}}[f] \leq \mathbb{E}_{\mathbb{P}_n}[f] + \mathcal{R}_{\mathbb{P}_n, 2}(\tilde{\rho}_n; f) + 4r_{n\star} + 2\epsilon_n + \hbar \tilde{\rho}_n^2.$$

REMARK 1 (PERFORMANCE BOUND ON THE ROBUST OPTIMAL SOLUTION). Denote $\mathcal{L}(\theta) := \mathbb{E}_{\mathbb{P}_{\text{true}}}[f_\theta]$, $\mathcal{L}_n(\theta) := \mathbb{E}_{\mathbb{P}_n}[f_\theta]$ and $\mathcal{L}_n^{\text{rob}}(\theta; \rho) := \sup_{\mathbb{P}: \mathcal{W}_p(\mathbb{P}, \mathbb{P}_n) \leq \rho} \mathbb{E}_\mathbb{P}[f_\theta]$. Then the high-probability bounds we have developed in this section have the form

$$\mathcal{L}(\theta) \leq \mathcal{L}_n^{\text{rob}}(\theta; \rho_n) + \tilde{\epsilon}_n, \quad \forall \theta \in \Theta.$$

Let $\theta_* \in \arg\min_{\theta \in \Theta} \mathcal{L}(\theta)$. Using the decomposition

$$\mathcal{L}(\theta_{\text{rob}}) - \mathcal{L}(\theta_*)$$
$$= \left(\mathcal{L}(\theta_{\text{rob}}) - \mathcal{L}_n^{\text{rob}}(\theta_{\text{rob}}; \rho_n)\right) + \left(\mathcal{L}_n^{\text{rob}}(\theta_{\text{rob}}; \rho_n) - \mathcal{L}_n^{\text{rob}}(\theta_*; \rho_n)\right) + \left(\mathcal{L}_n^{\text{rob}}(\theta_*; \rho_n) - \mathcal{L}_n(\theta_*)\right) + \left(\mathcal{L}_n(\theta_*) - \mathcal{L}(\theta_*)\right)$$
$$\leq \tilde{\epsilon}_n + 0 + \left(\rho_n \mathcal{V}(\theta_*) + O(\rho_n^2)\right) + \left(\mathcal{L}_n(\theta_*) - \mathcal{L}(\theta_*)\right).$$

Observe that $\mathcal{L}_n(\theta_*) - \mathcal{L}(\theta_*) = O_p(1/\sqrt{n})$ under mild conditions. Hence, when $\tilde{\epsilon}_n = \tilde{O}(1/n)$ and $\rho_n = \tilde{O}(1/\sqrt{n})$, the right-hand side provides an $\tilde{O}(1/\sqrt{n})$ bound on the losses between the robust optimal solution and the true optimal solution. If $\mathcal{L}$ satisfies the Polyak-Łojasiewicz condition (or equivalently, the quadratic growth condition) [51]

$$\mathcal{L}(\theta) - \mathcal{L}(\theta_*) \geq \frac{\mu}{2} \|\theta - \theta_*\|^2, \quad \forall \theta \in \Theta,$$

where $\mu > 0$, then $\|\theta_{\text{rob}} - \theta_*\|$ is upper bounded by a multiple of $\mathcal{L}(\theta_{\text{rob}}) - \mathcal{L}(\theta_*)$, thereby $\|\theta_{\text{rob}} - \theta_*\| = \tilde{O}_p(1/\sqrt{n})$.

## 5. Applications

In this section, we demonstrate our theoretical results in the context of various applications in operations research and machine learning.

### 5.1. Performance Guarantees for Wasserstein DRO

**5.1.1. Big-data Newsvendor** We first consider a big-data newsvendor problem in the spirit of [3]. In this problem, the decision maker needs to find the optimal ordering quantity for a product with an unknown random demand $y$, subject to holding cost $h > 0$ and back-order cost $b > 0$. In the world of big data, before deciding the ordering quantity, the decision maker observes a vector of features (such as product information, customer profiles, economic indicators, etc.) and thus can use them make a better ordering decision using these feature information. The vector of features is modeled as a $d$-dimensional random variable $X$ and the decision maker has collected historical observations of the feature-demand vector of the product. The goal is to find a decision rule that maps every realization $x$ from the feature space $\mathcal{X} \subset \mathbb{R}^d$ to an ordering decision. For illustration, we focus on a simple linear decision rule parameterized by $\theta \in \mathbb{R}^d$, thereby the ordering quantity for a product with feature $x \in \mathcal{X}$ is $\theta^\top x$. Thus, for a given $\theta$, the expected cost equals $\mathbb{E}_\mathbb{P}\left[h(\theta^\top x - y)_+ + b(y - \theta^\top x)_+\right]$, where $\mathbb{P}$ is the joint distribution of feature-demand vector $z = (x, y)$.

EXAMPLE 1 (FEATURE-BASED NEWSVENDOR). Let $\mathcal{Z} = (\mathcal{X} \times \mathbb{R}_+, \|\cdot\|)$ be the space of feature-demand vectors. Consider the following distributionally robust feature-based newsvendor problem

$$\min_{\theta \in \Theta} \sup_{\mathbb{P}: \mathcal{W}_1(\mathbb{P}, \mathbb{P}_n) \le \rho_n} \mathbb{E}_{\mathbb{P}}\left[h(\theta^\top x - y)_+ + b(y - \theta^\top x)_+\right].$$

Suppose $\Theta \subset \{\theta \in \mathbb{R}^d : \|\theta\|_* \le B\}$ and $\mathbb{P}_{\text{true}}$ satisfies $\mathcal{T}_1(\tau)$. Let $f_\theta(x,y) = h(\theta^\top x - y)_+ + b(y - \theta^\top x)_+$ and $\mathcal{F} = \{f_\theta : \theta \in \Theta\}$.

We have $|f_{\tilde{\theta}}(z) - f_\theta(z)| \le (h \vee b)\|x\|\|\tilde{\theta} - \theta\|_*$ and thus Assumption 3 holds with $\kappa(z) = (h \vee b)\|x\|$. Hence, by Corollary 2(II), setting $\epsilon_n = \frac{1}{n}(3\mathbb{E}_{\mathbb{P}_{\text{true}}}[\|x\|] + \mathbb{V}\mathrm{ar}_{\mathbb{P}_{\text{true}}}[\|x\|] + \sqrt{\frac{\tau t}{n}}) = O(1/n)$, with probability at least $1 - 1/n - \exp(-t)$, we have

$$\mathbb{E}_{\mathbb{P}_{\text{true}}}[f_\theta] \le \mathbb{E}_{\mathbb{P}_n}[f_\theta] + \sqrt{\frac{\tau t(1 + d\log(1 + 2Bn))}{n}} \|f_\theta\|_{\text{Lip}} + \epsilon_n, \quad \forall \theta \in \Theta.$$

Set $\rho_n = \frac{h \vee b}{h \wedge b} \sqrt{\frac{\tau t(1 + d\log(1 + 2Bn))}{n}}$. By Lemma 15 in Appendix D.1, we have $\rho_n \|f_\theta\|_{\text{Lip}} \le \mathcal{R}_{\mathbb{P}_n, 1}(f_\theta; \rho_n) = \tilde{O}(\sqrt{d/n})$. Thus, with probability at least $1 - 1/n - \exp(-t)$,

$$\mathbb{E}_{\mathbb{P}_{\text{true}}}[f_\theta] \le \mathbb{E}_{\mathbb{P}_n}[f_\theta] + \mathcal{R}_{\mathbb{P}_n, 1}(f_\theta; \rho_n) + \epsilon_n, \quad \forall \theta \in \Theta. \qquad \clubsuit$$

We remark that in this example, the newsvendor loss function satisfies only Assumption 1(I) but not Assumption 1(II). Hence we did not directly use Corollary 2(II) to derive the performance guarantee. Instead, Lemma 15 in Appendix D.1 actually shows that $\mathcal{R}_{\mathbb{P}_n, 1}(f_\theta; \rho_n)$ can achieve a fraction $\frac{h \wedge b}{h \vee b}$ of $\|f_\theta\|_{\text{Lip}}$ (and thus $\mathcal{R}_{\mathbb{P}_{\text{true}}, 1}(f_\theta; \rho_n)$) uniformly for all $\theta$, thereby we can still choose $\rho_n = \tilde{O}(\sqrt{d/n})$ to ensure a good performance guarantee for the Wasserstein robust solution.

**5.1.2. Linear Prediction** We consider supervised learning with linear models. Let $z = (x,y) \in \mathcal{Z} = \mathcal{X} \times \mathcal{Y} \subset \mathbb{R}^d \times \mathbb{R}$. To ease the exposition, we assume $\|z - \tilde{z}\| = \|x - \tilde{x}\|_2 + \infty \mathbf{1}\{y \ne \tilde{y}\}$, thereby we only focus on the $x$-component when calculating the gradient. Set

$$l(u, y) := \begin{cases} \ell(u - y), & \text{regression}, \\ \ell(yu), & \text{classification}, \end{cases} \qquad (3)$$

where $\ell : \mathbb{R} \to \mathbb{R}$ is $L$-Lipschitz continuous, and $\mathcal{Y} \subset \mathbb{R}$ for regression and $\mathcal{Y} = \{\pm 1\}$ for classification, and denote $l \circ f_\theta(z) := l(f_\theta(x), y)$. We denote by $l'$ the derivative function of $l$ with respect to its first argument, which is well-define almost everywhere in $\mathbb{R}$. Denote by $\mathbb{P}^x_{\text{true}}$ the $x$-marginal distribution of $\mathbb{P}_{\text{true}}$. The two examples considered in this subsection are on linear predictions for $p = 1$ and $p = 2$ respectively.

EXAMPLE 2 (LINEAR CLASS WITH LIPSCHITZ LOSS, 1-WASSERSTEIN DRO). Let $\Theta \subset \{\theta \in \mathbb{R}^d : \|\theta\|_2 \le B\}$ for some $B > 0$. Define

$$\mathcal{F} = \{x \mapsto f_\theta(x) = \theta^\top x : \theta \in \Theta\}.$$

Consider loss functions of the composite form (3) and let $f_\theta(x) = \theta^\top x$. Assume $\mathbb{P}^x_{\text{true}}$ is sub-Gaussian, i.e., there exists $a > 0$ such that $C := \log \mathbb{E}_{\mathbb{P}_{\text{true}}}[\exp(a\|x\|_2^2)] < \infty$. Assume $\ell$ satisfies $\limsup_{|t| \to \infty} \frac{\ell(t)}{|t|} = L$. Examples of $\ell(t)$ include convex losses such as hinge loss $(1-t)_+$, softplus (logistic) loss $\log(1 + e^t)$, as well as non-convex losses such as inverse S-shaped curve $\mathrm{sgn}(t)\log(\frac{1}{2}(1 + e^t))$.

Let us verify the assumptions in Corollary 5. By Lemma 3 in Appendix B, $\mathbb{P}^x_{\text{true}}$ satisfies $\mathcal{T}_1(\frac{2}{a}(1+C))$. Assumption 1 holds with $\|f_\theta\|_{\text{Lip}} = \|\theta\|_2 \le B = \gamma_1$. Furthermore, since

$$\{cf_\theta : \theta \in \Theta, 0 \le c \le 1, c^2 \|\theta\|_2^2 \le r\} \subset \{f_\theta : \|\theta\|_2 \le \sqrt{r}\},$$

by Lemma 16 in Appendix D.2, $\sqrt{r\mathbb{E}_{\mathbb{P}_{\text{true}}}[\|x\|_2^2]/n}$ is an upper bound of $\mathbb{E}_\otimes[\mathfrak{R}_n(\{f_\theta : \|\theta\|_2 \le \sqrt{r}\})]$. By Jensen's inequality, we have

$$\mathbb{E}_{\mathbb{P}_{\text{true}}}[\|x\|_2^2] \le \frac{1}{a}\log\mathbb{E}_{\mathbb{P}_{\text{true}}}\big[\exp(a\|x\|_2^2)\big] = \frac{C}{a}.$$

It follows that we can set $\psi_n(r) = \sqrt{\frac{rC}{an}}$ and thus $r_{n\star} = \frac{C}{an}$.

Let $t > 0$. Set

$$\rho_n = 2\Big(L\sqrt{\frac{2(1+C)t}{an}} + \sqrt{\epsilon_n}\Big), \quad \epsilon_n = 4L^2\frac{C}{an} + \frac{2L}{n}.$$

Then $\rho_n = O(1/\sqrt{n})$ and $\epsilon_n = O(1/n)$ are dimension-independent, and by Corollary 5 and Lemma 1, with probability at least $1 - \lceil(\frac{1}{2}\log_2(LB\frac{2}{a}(1+C)tn)\rceil e^{-t}$,

$$\mathbb{E}_{\mathbb{P}_{\text{true}}}[l(\theta^\top x, y)] - \mathbb{E}_{\mathbb{P}_n}[l(\theta^\top x, y)] \le \mathcal{R}_{\mathbb{P}_n,1}(\rho_n; l \circ f_\theta) + \epsilon_n = \rho_n\|\theta\|_2 + \epsilon_n, \quad \forall \theta \in \Theta. \quad \clubsuit$$

The bound obtained in Example 2 is consistent to the existing literature on the generalization bounds for linear predictions. But unlike the typical results (e.g. [77, 24]), we do not impose boundedness assumptions on the loss function $\ell$ or the domain $\mathcal{Z}$. If imposing a positive lower bound on $\|\theta\|_2 \ge c > 0$, the bound given in the example further becomes

$$\mathbb{E}_{\mathbb{P}_{\text{true}}}[l(\theta^\top x, y)] - \mathbb{E}_{\mathbb{P}_n}[l(\theta^\top x, y)] \le (\rho_n + \epsilon_n/c)\|\theta\|_2 = \mathcal{R}_{\mathbb{P}_n,1}(\rho_n + \epsilon_n/c; l \circ f_\theta).$$

Thereby one can bound the true loss using only the Wasserstein robust loss with an inflated radius $\rho_n + \epsilon_n/c$ without having a higher order error term. This bound is of the same form as in Shafieezadeh-Abadeh et al. [74, Theorem 39] which has a linear dependence on the dimension $d$ of $\mathcal{X}$ (albeit under a light-tail assumption that is slightly weaker than $T_1$); while our bound $\rho_n = O(1/\sqrt{n})$ is independent of $d$ (at least for the case of 2-norm).

EXAMPLE 3 (LINEAR CLASS WITH LIPSCHITZ LOSS, 2-WASSERSTEIN DRO). Consider a similar setup as in Example 2 but with slightly different notations in order to be consistent with Corollary 3. Let $\Theta \subset \{\theta \in \mathbb{R}^d : \|\theta\|_* \le B\}$ for some $B > 0$. Define

$$\mathcal{F} = \big\{(x, y) \mapsto l(\theta^\top x, y) : \theta \in \Theta\big\},$$

where $l$ is defined in (3). Let $f_\theta(z) = l(\theta^\top x, y)$. Then $\|\nabla f_\theta(z)\|_* = \|\nabla_x l(\theta^\top x, y)\|_* = \|\theta\|_*|l'(\theta^\top x, y)|$, recalling $l'$ denotes the derivative of $l$ with respect to its first argument. Assume further that $\ell$ in (3) has $\hbar$-Lipschitz gradient; $\mathbb{P}_{\text{true}}^x$ satisfies $T_2(\tau)$ for some $\tau > 0$; and $\inf_{\theta \in \Theta}\mathbb{E}_{\mathbb{P}_{\text{true}}}[l'(\theta^\top x, y)^2] > 0$. Note that the last condition is mild – indeed, it is satisfied if for every $\theta \in \Theta$, $l'(\theta^\top \cdot, y)$ is non-zero on some subset of $\mathcal{X}$ with positive $\mathbb{P}_{\text{true}}^x$-measure (together with the boundedness assumption on $\Theta$).

Let us verify the conditions and compute the constants in Corollary 3. Assumption 2 is satisfied since $f_\theta$ has $\hbar B^2$-Lipschitz gradient, and $\sigma = \sup_{\theta \in \Theta}\mathbb{E}_{\mathbb{P}_{\text{true}}}[l'(\theta^\top x, y)^4]^{\frac{1}{2}} / \mathbb{E}_{\mathbb{P}_{\text{true}}}[l'(\theta^\top x, y)^2] \le L^2 / \inf_{\theta \in \Theta}\mathbb{E}_{\mathbb{P}_{\text{true}}}[l'(\theta^\top x, y)^2] < \infty$. We have $|f_{\tilde\theta}(z) - f_\theta(z)| \le L_\ell\|x\|\|\tilde\theta - \theta\|_*$ and by Lemma 17, $\|\nabla f_\theta(z) - \nabla f_{\tilde\theta}(z)\|_* \le (L_\ell + B\hbar\|x\|)\|\tilde\theta - \theta\|_*$. Hence Assumptions 3 and 4 hold.

Let $t > 0$ and set

$$\rho_n = \sqrt{\frac{\tau t(1+d\log(2+2Bn))}{n}}\Big(1 + \sigma\sqrt{\frac{2t(1+d\log(2+2Bn))}{n}}\Big),$$

$\tilde\epsilon_n = \big(2L_\ell + 2B\hbar\mathbb{E}_{\mathbb{P}_{\text{true}}}[\|x\|] + B^2\hbar^2\mathbb{V}\text{ar}_{\mathbb{P}_{\text{true}}}[\|x\|] + \rho_n\sqrt{\mathbb{E}_{\mathbb{P}_{\text{true}}}[(L_\ell + B\hbar\|x\|)^2] + \mathbb{V}\text{ar}_{\mathbb{P}_{\text{true}}}[(L_\ell + B\hbar\|x\|)^2]}\big)/n$.

By Corollary 3, with probability at least $1 - 2/n - 2\exp(-t)$, for every $\theta \in \Theta$,

$$\mathbb{E}_{\mathbb{P}_{\text{true}}}[l(\theta^\top x, y)] - \mathbb{E}_{\mathbb{P}_n}[l(\theta^\top x, y)] \le \mathcal{R}_{\mathbb{P}_n,2}(\rho_n; f_\theta) + \tilde\epsilon_n + \frac{2\hbar B^2\tau t(1+d\log(2+2Bn))}{n}.$$

Note that we have $\rho_n = \tilde O(\sqrt{d/n})$ and an $\tilde O(d/n)$ remainder.

**5.1.3. Portfolio Optimization** In this subsection, we study the classic Markowitz's mean-variance portfolio optimization problem. Let $x$ be a vector of random losses of $d$ assets with distribution $\mathbb{P}^x_{\text{true}}$ and let $w \in \mathcal{W} \subset \mathbb{R}^d$ be the portfolio weights satisfying $w^\top \mathbf{1} = 1$. Note that the variance of a one-dimensional random variable $Y$ has an equivalent representation $\mathbb{V}\text{ar}[Y] = \min_{u \in \mathbb{R}} \mathbb{E}[(Y-u)^2]$.

EXAMPLE 4 (MARKOWITZ MODEL). Let $\alpha > 0$ and suppose $\mathcal{W} \subset \{w \in \mathbb{R}^d : w^\top \mathbf{1} = 1, \|w\|_2 \leq B\}$, where $B > 0$. Consider the following distributionally robust mean-variance minimization

$$\min_{w \in \mathcal{W}, u \in \mathbb{R}} \sup_{\mathbb{P} : \mathcal{W}_2(\mathbb{P}^x, \mathbb{P}^x_n) \leq \rho_n} \mathbb{E}_{\mathbb{P}^x} \left[ (w^\top x - u)^2 + \alpha w^\top x \right].$$

Assume $\mathbb{P}^x_{\text{true}}$ satisfies $T_2(\tau)$. Then it also satisfies $T_1(\tau)$ and by Corollary 1, for every $w \in \mathbb{R}^d$ with $\|w\|_2 = 1$, $\mathbb{P}_\otimes\{|\mathbb{E}_{\mathbb{P}^x_n}[w^\top x] - \mathbb{E}_{\mathbb{P}^x_{\text{true}}}[w^\top x]| \geq \epsilon\} \leq 2e^{-n\epsilon^2/\tau}$ for all $\epsilon > 0$, thus $\mathbb{P}^x_{\text{true}}$ is $\sqrt{\tau/2}$-subgaussian. Let $\mu_j = \mathbb{E}_{\mathbb{P}^x_{\text{true}}}[\|x\|_2^j]^{\frac{1}{j}} < \infty$, $j = 1, 2, 3, 4$. Assume there exists $\zeta > 0$ such that $\mathbb{C}\text{ov}_{\mathbb{P}_{\text{true}}}[x] \geq \zeta I$.

Set $U_n := B(\mu_1 + \sqrt{\tau t/n} + \rho_n)$ and $U := \sup_n U_n$, which is bounded whenever $\{\rho_n\}_n$ is bounded. By Lemma 18 in Appendix D.3, with probability at least $1 - e^{-t}$, the problem is equivalent to

$$\min_{w \in \mathcal{W}, |u| \leq U_n} \sup_{\mathbb{P} : \mathcal{W}_2(\mathbb{P}^x, \mathbb{P}^x_n) \leq \rho_n} \mathbb{E}_{\mathbb{P}^x} \left[ (w^\top x - u)^2 + \alpha w^\top x \right].$$

Let $\theta = (w, \tilde{u})$ where $\tilde{u} = -u + \alpha/2$, $z = (x, y)$ and $f_\theta(z) = (w^\top x + \tilde{u}^\top y)^2 + \alpha u - \alpha^2/4 = (\theta^\top z)^2 - \alpha \tilde{u} + \alpha^2/4$. Thereby we have $f_\theta(x, 1) = (w^\top x - u)^2 + \alpha w^\top x$. Set $\Theta = \{(w, \tilde{u}) : w \in \mathcal{W}, |-\tilde{u} + \alpha/2| \leq U\}$. For any distribution $\mathbb{P}^x$, we represent $\mathbb{P} = \mathbb{P}^x \otimes \boldsymbol{\delta}_1$. Consider the problem

$$\min_{\theta = (w, \tilde{u}) \in \Theta} \left\{ \sup_{\mathbb{P} : \mathcal{W}_2(\mathbb{P}, \mathbb{P}_n) \leq \rho_n} \mathbb{E}_{\mathbb{P}} \left[ (\theta^\top z)^2 \right] - \alpha \tilde{u} + \alpha^2/4 \right\}.$$

Let us verify the assumptions and compute the constants in Theorem 3 for the inner maximization problem. $\mathbb{P}_{\text{true}}$ satisfies $T_2(\tau)$ since any distribution $\mathbb{Q} \in \mathcal{P}_2(\mathcal{Z})$ with finite $H(\mathbb{Q}, \mathbb{P}_{\text{true}})$ has the form $\mathbb{Q}^x \otimes \boldsymbol{\delta}_1$, where $\mathbb{Q}^x$ is the $x$-marginal of $\mathbb{Q}$. We have $\nabla f_\theta(z) = 2(\theta^\top z)\theta$, thus $\nabla^2 f_\theta(z) = 2\theta\theta^\top$. Note that $\|\theta\|_2^2 \leq B^2 + (U + \alpha/2)^2$, hence Assumption 2 is satisfied with $\hbar = 2(B^2 + (U + \alpha/2)^2)$. To find a sub-root function $\psi_n(r)$ required by Assumption 5, observe that

$$\| \|\nabla f_\theta\|_2 \|_{\mathbb{P}_{\text{true}},2} = 2\|\theta\|_2 \mathbb{E}_{\mathbb{P}_{\text{true}}}[(\theta^\top z)^2]^{\frac{1}{2}} \geq 2\sqrt{\zeta}\|\theta\|_2^2.$$

It follows from Lemma 19 in Appendix D.3 that

$$\mathbb{E}_\otimes \left[ \mathfrak{R}_n(\{c f_\theta : \theta \in \Theta, 0 \leq c \leq 1, c^2 \| \|\nabla f_\theta\|_2 \|^2_{\mathbb{P}_{\text{true}},2} \leq r\}) \right] \leq \mu_4^2 \sqrt{\frac{r}{4\zeta n}} := \psi_n(r).$$

Thus $r_{n\star} = \frac{\mu_4^4}{4\zeta n}$. Moreover, by Lemma 20 in Appendix D.3,

$$\mathbb{E}_\otimes \left[ \mathfrak{R}_n\left(\left\{ \frac{\|\nabla f_\theta\|_2^2}{\| \|\nabla f_\theta\|_2 \|^2_{\mathbb{P}_{\text{true}},2}} : \theta \in \Theta \right\}\right) \right] \leq \frac{1}{1 \wedge \zeta} \sqrt{\frac{\mu_4^4 + 2\mu_2^2 + 1}{n}} =: R_n.$$

In addition, $\sup_{\theta \in \Theta} \| \|\nabla f_\theta\|_2 \|_{\mathbb{P}_{\text{true}},2} = \sup_{\theta \in \Theta} 2\|\theta\|_2 \mathbb{E}_{\mathbb{P}_{\text{true}}}[(\theta^\top z)^2]^{\frac{1}{2}} \leq 2\mu_2(B^2 + (U + \alpha/2)^2) = \gamma_2$.

Let $t > 0$. Then in Theorem 3, set

$$\rho_n = 2\sqrt{\frac{\tau t}{n}}(1 + R_n) + \sqrt{\frac{\mu_4^4}{\zeta n} + 2\epsilon_n}, \quad \epsilon_n = \frac{2(B^2 + (U + \alpha/2)^2)\tau t + 1 + R_n}{n}.$$

We have with probability at least $1 - \lceil \log_2(\sqrt{\gamma_2 \tau t n}) \rceil e^{-t}$,

$$\mathbb{E}_{\mathbb{P}_{\text{true}}}[(\theta^\top z)^2] \leq \mathbb{E}_{\mathbb{P}_n}[(\theta^\top z)^2] + \rho_n \|\|\nabla f_\theta\|_2\|_{\mathbb{P}_{\text{true}},2} + \epsilon_n, \quad \forall \theta \in \Theta.$$

By Lemma 22 in Appendix D.3, a generalized version of Corollary 6 applied to unbounded gradient norm, we set

$$\tilde{\rho}_n = C(1+R_n), \quad \tilde{\epsilon}_n = \epsilon_n + 2(B^2 + (U+\alpha/2)^2)\tilde{\rho}_n^2,$$

where $C$ depends only on the distribution of $\|z\|_2$, $z \sim \mathbb{P}_{\text{true}}$. Then $\tilde{\rho}_n = \tilde{O}(1/\sqrt{n})$, $\tilde{\epsilon}_n = O(1/n)$ are both dimension-independent, and with probability at least $1 - (\lceil \log_2(\sqrt{\gamma_2 \tau t n}) \rceil + 2)e^{-t}$,

$$\mathbb{E}_{\mathbb{P}_{\text{true}}}[(\theta^\top z)^2] \leq \mathbb{E}_{\mathbb{P}_n}[(\theta^\top z)^2] + \mathcal{R}_{\mathbb{P}_n, p}(\tilde{\rho}_n; f_\theta) + \tilde{\epsilon}_n, \quad \forall \theta \in \Theta,$$

and therefore by minimizing over $u$ or $\tilde{u}$,

$$\mathbb{V}\text{ar}_{\mathbb{P}_{\text{true}}}[w^\top x] + \alpha \mathbb{E}_{\mathbb{P}_{\text{true}}}[w^\top x] \leq \min_{u \in \mathbb{R}} \sup_{\mathbb{P}: \mathcal{W}_2(\mathbb{P}^x, \mathbb{P}_n^x) \leq \tilde{\rho}_n} \mathbb{E}_{\mathbb{P}^x}\left[(w^\top x - u)^2 + \alpha w^\top x\right] + \tilde{\epsilon}_n, \quad \forall w \in \mathcal{W}. \quad \clubsuit$$

The last inequality in Example 4 shows that the true mean-variance of the portfolio $w_n$ is upper bounded by its robust mean-variance up to a higher-order term. We remark that in this example, the parameter space is not bounded as $u \in \mathbb{R}$, which makes it impossible to obtain a bounded complexity for the entire class of loss functions. We circumvent such a difficulty by showing that there exists an optimal solution lying in a bounded set $\Theta$ with high probability, thereby it suffices to restrict on $\Theta$.

### 5.2. Performance Guarantees for Variation Regularization

In the next two examples, we illustrate our results for Lipschitz regularization and gradient regularization for nonlinear classes. Similar to Section 5.1.2, we let $z = (x, y) \in \mathcal{Z} = \mathcal{X} \times \mathcal{Y} \subset \mathbb{R}^d \times \mathbb{R}$ and assume $\|z - \tilde{z}\| = \|x - \tilde{x}\|_2 + \infty \mathbf{1}\{y \neq \tilde{y}\}$.

**5.2.1. Kernel Method** We consider Lipschitz regularization of kernel class (see, e.g., [89, Chapter 12]). Let $\boldsymbol{k}: \mathcal{X} \times \mathcal{X} \to \mathbb{R}_+$ be a positive definite kernel on $\mathcal{X} \subset (\mathbb{R}^d, \|\cdot\|_2)$ with $\sigma := (\mathbb{E}_{x \sim \mathbb{P}_{\text{true}}}[\boldsymbol{k}(x, x)])^{\frac{1}{2}} < \infty$. We can associate $\boldsymbol{k}$ with a feature map $\Phi : \mathcal{X} \to \mathcal{H}$, where $\mathcal{H}$ is a Hilbert space with inner product $\langle \cdot, \cdot \rangle$ and $\boldsymbol{k}(x_1, x_2) = \langle \Phi(x_1), \Phi(x_2) \rangle$. Denote by $\|\cdot\|_{\mathcal{H}}$ a norm on $\mathcal{H}$. Let $m \in \mathbb{N}_{\geq 1}$ and $\{x_j\}_{j=1}^\infty \subset \mathcal{X}$. Then we have $\|\sum_{j=1}^m \theta_j \Phi(x_j)\|^2 = \sum_{j,k=1}^m \theta_j \theta_k \boldsymbol{k}(x_j, x_k)$.

In kernel method, one often consider the following parameterized class

$$\mathcal{F} = \left\{ x \mapsto \sum_{j=1}^\infty \theta_j \boldsymbol{k}(x, x_j) : \theta \in \Theta \right\}, \text{ where } \Theta = \left\{ \theta = (\theta_j)_{j=1}^m : \|\theta\|_{\boldsymbol{k}} \leq B, \ m \in \mathbb{N}_{\geq 1} \right\},$$

for some $B > 0$, where for $\theta = (\theta_j)_{j=1}^m$, we adopt the convention $\sum_{j=1}^\infty \theta_j \boldsymbol{k}(x, x_j) = \sum_{j=1}^m \theta_j \boldsymbol{k}(x, x_j)$ and $\|\theta\|_{\boldsymbol{k}}^2 = \sum_{j,k=1}^\infty \theta_j \theta_k \boldsymbol{k}(x_j, x_k) = \sum_{j,k=1}^m \theta_j \theta_k \boldsymbol{k}(x_j, x_k)$.

EXAMPLE 5 (LIPSCHITZ REGULARIZATION FOR KERNEL CLASS). Consider loss functions of the form (3) with $u = f_\theta \in \mathcal{F}$ defined as above. Assume $\boldsymbol{k}$ is differentiable and there exists $\zeta > 0$ such that $\|\|\nabla f_\theta\|_2\|_{\mathbb{P}_{\text{true}},2}^2 = \mathbb{E}_{x \sim \mathbb{P}_{\text{true}}}[\|\sum_{j=1}^\infty \theta_j \nabla_x \boldsymbol{k}(x, x_j)\|_2^2] \geq \zeta \sum_{j,k=1}^\infty \theta_j \theta_k \boldsymbol{k}(x_j, x_k)$ for all $\theta \in \Theta$, which can be satisfied when the matrix $(\mathbb{E}_{x \sim \mathbb{P}_{\text{true}}}[\nabla_x \boldsymbol{k}(x, x_j)^\top \nabla_x \boldsymbol{k}(x, x_k)])_{1 \leq j, k \leq \infty}$ is positive definite. Furthermore, assume $\gamma_1 = \sup_{\theta \in \Theta, x \in \mathcal{X}} \|\sum_{j=1}^\infty \theta_j \nabla_x \boldsymbol{k}(x, x_j)\| < \infty$, thus Assumption 1(I) is satisfied.

Let us compute the generalization bound using Corollary 5. To this end, we need to specify $\psi_n$ in Assumption 5 and compute its fixed point $r_{n\star}$. Observe that

$$\left\{ c f_\theta : \theta \in \Theta, 0 \leq c \leq 1, c^2 \|f_\theta\|_{\text{Lip}}^2 \leq r \right\} \subset \left\{ c f_\theta : \theta \in \Theta, 0 \leq c \leq 1, \zeta c^2 \|\theta\|_{\boldsymbol{k}}^2 \leq r \right\} \subset \left\{ f_\theta : \|\theta\|_{\boldsymbol{k}} \leq \sqrt{r/\zeta} \right\},$$

hence $\psi_n(r)$ can be chosen as $2\sigma\sqrt{\frac{r}{n\zeta}}$, an upper bound of $\mathbb{E}_\otimes[\mathfrak{R}_n(\{f_\theta : \|\theta\|_k \leq \sqrt{r/\zeta}\})]$ according to [7, Lemma 22]. Thus $r_{n\star} = \frac{4\sigma^2}{n\zeta}$. Set

$$\rho_n = 2\Big(L\sqrt{\frac{\tau t}{n}} + \sqrt{\epsilon_n}\Big), \quad \epsilon_n = \frac{16L^2\sigma^2}{n\zeta} + \frac{2L}{n}.$$

Then $\rho_n = O(1/\sqrt{n})$ and $\epsilon_n = O(1/n)$ are dimension-independent, and by Corollary 5, with probability at least $1 - \lceil \log_2(\sqrt{L\gamma_1\tau t n})\rceil e^{-t}$,

$$\mathbb{E}_{\mathbb{P}_{\text{true}}}[l \circ f_\theta] \leq \mathbb{E}_{\mathbb{P}_n}[l \circ f_\theta] + \rho_n\|f_\theta\|_{\text{Lip}} + \epsilon_n, \quad \forall \theta \in \Theta. \qquad \clubsuit$$

This result provides a generalization bound for Lipschitz regularization problems [61, 37, 86] when the loss function class belongs to a kernel class. We remark that the setup in this example is different from Shafieezadeh-Abadeh et al. [74, Section 3.3], in which the distributional uncertainty is imposed on the feature space, while Example 5 considers distributional uncertainty on the original data space.

**5.2.2. Neural Networks** In the last example, we illustrate the generalization bound of gradient regularization for a simple two-layer neural network. Consider

$$\mathcal{F} = \Big\{(x, y) \mapsto l\big(W_2\phi(W_1 x), y\big) : (W_1, W_2) \in \Theta\Big\},$$

where $l$ is defined in (3), $\phi = (\phi_1, \ldots, \phi_{d_2})$ are entry-wise 1-Lipschitz activation functions, and $\Theta$ is the space of weight matrices

$$\Theta = \{\theta = (W_1, W_2) : W_1 \in \mathbb{R}^{d_2 \times d_1}, W_2 \in \mathbb{R}^{1 \times d_2}, W_1 W_1^\top = I, \|W_2\|_2 \leq B\}.$$

Here the constraint $W_1 W_1^\top = I$ enforces the orthonormal regularization on the weight matrix [88, 98, 4, 48], which is a popular way to ensure the training stability and performance for neural nets. Let $f_\theta(z) = l(W_2\phi(W_1 x), y)$.

EXAMPLE 6 (GRADIENT REGULARIZATION FOR NEURAL NETWORKS). Assume $l$ and $\phi_j$ has $\hbar_l$- and $\hbar_\phi$-Lipschitz gradient, respectively, $j = 1, \ldots, d_2$, thereby $f_\theta$ is smooth and $\mathcal{F}$ satisfies Assumption 2 with $\hbar = 2L(L\hbar_\phi + B\hbar_l)\sqrt{d_2}$ by Lemma 25 in Appendix D.4. Assume $\eta := \inf_{\theta\in\Theta, z\in\mathcal{Z}} l'(W_2\phi(W_1 x), y) > 0$, which can be satisfied, for example, when $\mathcal{Z}$ is bounded and $l$ is the logistic loss. Furthermore, assume there exists $\zeta > 0$ such that $\mathbb{E}_{\mathbb{P}_{\text{true}}^x}[\|W_2\phi'(W_1 x)\|_2^2] \geq \zeta\|W_2\|_2^2$ for every $(W_1, W_2) \in \Theta$, where $\phi' = (\phi'_1, \ldots, \phi'_{d_2})$, and $\sigma = \mathbb{E}_{\mathbb{P}_{\text{true}}^x}[\|x\|_2^2]^{\frac{1}{2}} < \infty$.

Let us compute the constants in Corollary 6. We have

$$\|\nabla f(z)\|_* = \|l'(W_2\phi(W_1 x), y)W_2\phi'(W_1 x)W_1\|_2^2 = l'(W_2\phi(W_1 x), y)^2\|W_2\phi'(W_1 x)\|_2^2.$$

Thus, $\|\nabla f(z)\|_* \leq LB$, $\|\|\nabla f\|_*\|_{\mathbb{P}_{\text{true}},2} \geq \eta\sqrt{\zeta}\|W_2\|_2$. It follows that

$$\frac{\|\nabla f(z)\|_*}{\|\|\nabla f\|_*\|_{\mathbb{P}_{\text{true}},2}} \leq \frac{LB}{\eta\sqrt{\zeta}\|W_2\|_2} =: \kappa_g.$$

In addition, $c^2\|\|\nabla f\|_*\|_{\mathbb{P}_{\text{true}},2}^2 \leq r$ implies $\|W_2\|_2 \leq \frac{\sqrt{r}}{c\eta\sqrt{\zeta}}$. As a result, we can choose $\psi_n(r)$ in Assumption 5 as

$$\mathbb{E}_\otimes\Big[\mathfrak{R}_n\Big\{cf_\theta : \theta \in \Theta, 0 \leq c \leq 1, c^2\|\|\nabla f_\theta\|_*\|_{\mathbb{P}_{\text{true}},2}^2 \leq r\Big\}\Big]$$
$$\leq L\mathbb{E}_\otimes\Big[\mathfrak{R}_n\big\{x \mapsto W_2\phi(W_1 x) : W_1 W_1^\top = I, \|W_2\|_2 \leq \frac{\sqrt{r}}{\eta\sqrt{\zeta}}\big\}\Big]$$
$$\leq \frac{L\sigma\sqrt{2rd_2}}{\eta\sqrt{\zeta n}} =: \psi_n(r),$$

where the first inequality follows from Lemma 8 in Appendix D, and the second inequality is due to Lemma 24 in Appendix D.4. It follows that $r_{n\star} = \frac{2L^2\sigma^2 d_2}{\eta^2 \zeta n}$. Moreover, by Lemma 25 in Appendix D.4,

$$\mathbb{E}_\otimes[\mathfrak{R}_n(\mathcal{G})] \leq \frac{2L(L\hbar_\phi + B\hbar_l)\sigma\sqrt{2d_2}}{\eta^2 \zeta \sqrt{n}} =: R_n.$$

Thereby, in Corollary 6, setting

$$\tilde{\rho}_n = \left(1 + 2R_n + \kappa_g \sqrt{\frac{t}{2n}}\right)\left(2\sqrt{\frac{\tau t}{n}}(1 + R_n) + \sqrt{\frac{8L^2\sigma^2 d_2}{\eta^2 \zeta n} + 2\epsilon_n}\right), \quad \epsilon_n = \frac{\hbar \tau t + 1 + R_n}{n},$$

it holds that $\tilde{\rho}_n = O(\sqrt{d_2/n})$, $\epsilon_n = O(\sqrt{d_2}/n)$, and with probability at least $1 - (\lceil \log_2(\sqrt{LB\tau tn})\rceil + 1)e^{-t}$, for every $\theta \in \Theta$,

$$\mathbb{E}_{\mathbb{P}_{\text{true}}}[f_\theta] \leq \mathbb{E}_{\mathbb{P}_n}[f_\theta] + \tilde{\rho}_n \mathbb{E}_{\mathbb{P}_n}[l'(W_2\phi(W_1x), y)^2 \|W_2\phi'(W_1x)\|_2^2]^{\frac{1}{2}} + \epsilon_n. \qquad \clubsuit$$

## 6. Concluding Remarks

In this paper, we have developed finite-sample non-asymptotic performance guarantees for Wasserstein DRO and its associated variation regularization without suffering from the curse of dimensionality. These results help us to understand the empirical success of Wasserstein DRO and/or Lipschitz and gradient regularization. In the meantime, many issues worth investigating are left to future work.

*More general loss function families and distribution families.* We restrict the families of loss functions consistent with Lemma 1 and Lemma 2 that establish the equivalence between Wasserstein DRO and variation regularization. One can extend the results to more general families such as non-smooth losses using the results in [39]; see a follow-up work [2]. In Section 3, we adopt the widely used transportation inequalities $T_p$, $p \in [1,2]$, which covers most subgaussian distributions and works for loss functions of linear and quadratic growth. One may obtain results for more general distributions and loss functions by considering other families of transportation-information inequalities [19, 44, 85, 45]. We remark that the finite-sample performance guarantee in [36], though suffers from the curse of dimensionality, does not restrict the loss function family other than the growth condition and works for any distribution that admits Wasserstein concentration.

*Wasserstein distance of other orders.* We focus primarily on the case $p \in [1,2]$. Indeed, our proof of Theorem 1 relies crucially on this setting where the tensorization lemma (Lemma 4 in Appendix B.1) applies. We leave the study for other orders of Wasserstein distance, including another important case $p = \infty$ that has been widely considered in adversarial robust learning (e.g., [43, 76, 105]), to future work.

*Complexity theory based on variation of the loss.* We developed a local Rademacher complexity theory based on the variation of the loss. Investigation of these techniques in the context of other problems in statistical learning theory seems interesting, and hopefully would yield new results.

In summary, we hope our results can inspire more fruitful findings for problems in operations research and machine learning in which Wasserstein distributional robustness plays an increasingly prominent role.

## Appendix A: Proofs for Section 2

*Proof of Lemma 1.* Using the strong dual problem (1), we have

$$\mathcal{R}_{\mathbb{Q},1}(\rho; f) = \min_{\lambda \geq 0} \left\{ \lambda\rho + \mathbb{E}_{\mathbb{Q}} \left[ \sup_{\tilde{z} \in \mathcal{Z}} \{f(\tilde{z}) - f(z) - \lambda\|\tilde{z} - z\|\} \right] \right\}.$$

Using the Lipschitz Assumption 1(I), for any $\lambda > \|f\|_{\text{Lip}}$, the inner supremum equals to zero attained at $\tilde{z} = z$. Hence, by taking a feasible solution $\lambda = \|f\|_{\text{Lip}}$, we obtain that

$$\mathcal{R}_{\mathbb{Q},1}(\rho; f) \leq \rho\|f\|_{\text{Lip}}.$$

Moreover, if Assumption 1(II) holds, then there exists a sequence $\{z_m\}_{m=1}^\infty \subset \mathcal{Z}$ such that

$$\|z_m - z_0\| \geq \max(\rho, m), \quad f(z_m) - f(z_0) \geq (\|f\|_{\text{Lip}} - 1/m)\|z_m - z_0\|.$$

Let $E \subset \mathcal{Z}$ be such that $\mu := \mathbb{Q}(E) > 0$. Consider a sequence of distributions

$$\mathbb{Q}_m = \mathbb{Q}_{|\mathcal{Z}\setminus E} + (1 - \epsilon_m)\mathbb{Q}_{|E} + \epsilon_m \boldsymbol{\delta}_{z_m},$$

where $\mathbb{Q}_{|\cdot}$ denotes the restriction of $\mathbb{Q}$ on a subset of $\mathcal{Z}$, and $\epsilon_m$ is chosen such that

$$\mathcal{W}_1(\mathbb{Q}, \mathbb{Q}_m) = \epsilon_m \cdot \mathbb{E}_{\mathbb{Q}_{|E}}[\|z_m - z\|]) = \rho.$$

It follows that $\epsilon_m \to 0$ and

$$\mathcal{R}_{\mathbb{Q},1}(\rho; f) \geq \mathbb{E}_{\mathbb{Q}_m}[f] - \mathbb{E}_{\mathbb{Q}}[f] \geq \epsilon_m \cdot \mathbb{E}_{\mathbb{Q}_{|E}}[(\|f\|_{\text{Lip}} - 1/m)\|z_m - z_0\|] + \epsilon_m(f(z_0) - \mathbb{E}_{\mathbb{Q}_{|E}}[f(z)])$$
$$= \rho(\|f\|_{\text{Lip}} - 1/m) + \epsilon_m(f(z_0) - \mathbb{E}_{\mathbb{Q}_{|E}}[f(z)]),$$

which converges to $\rho\|f\|_{\text{Lip}}$ as $m \to \infty$. Therefore we complete the proof. □

*Proof of Lemma 2.* Using the strong dual problem (1), we have

$$\mathcal{R}_{\mathbb{Q},2}(\rho; f) = \min_{\lambda \geq 0} \left\{ \lambda\rho^2 + \mathbb{E}_{\mathbb{Q}} \left[ \sup_{\tilde{z} \in \mathcal{Z}} \{f(\tilde{z}) - f(z) - \lambda\|\tilde{z} - z\|^2\} \right] \right\}.$$

By Assumption 2, we have that

$$\mathcal{R}_{\mathbb{Q},2}(\rho; f) \leq \min_{\lambda \geq 0} \left\{ \lambda\rho^2 + \mathbb{E}_{\mathbb{Q}} \left[ \sup_{\tilde{z} \in \mathbb{R}^d} \{\nabla f(z)^\top (\tilde{z} - z) - (\lambda - \hbar)\|\tilde{z} - z\|^2\} \right] \right\}$$
$$= \hbar\rho^2 + \min_{\lambda \geq 0} \left\{ \lambda\rho^2 + \mathbb{E}_{\mathbb{Q}} \left[ \sup_{\tilde{z} \in \mathbb{R}^d} \{\nabla f(z)^\top (\tilde{z} - z) - \lambda\|\tilde{z} - z\|^2\} \right] \right\}$$
$$= \hbar\rho^2 + \min_{\lambda \geq 0} \left\{ \lambda\rho^2 + \frac{1}{4\lambda} \mathbb{E}_{\mathbb{Q}} \left[ \|\nabla f(z)\|_*^2 \right] \right\}$$
$$= \hbar\rho^2 + \rho\|\|\nabla f\|_*\|_{\mathbb{Q},2},$$

and that

$$\mathcal{R}_{\mathbb{Q},2}(\rho; f) \geq \min_{\lambda \geq 0} \left\{ \lambda\rho^2 + \mathbb{E}_{\mathbb{Q}} \left[ \sup_{\tilde{z} \in \mathcal{Z}} \{\nabla f(z)^\top (\tilde{z} - z) - (\lambda + \hbar)\|\tilde{z} - z\|^2\} \right] \right\}$$
$$= -\hbar\rho^2 + \min_{\lambda \geq 0} \left\{ \lambda\rho^2 + \mathbb{E}_{\mathbb{Q}} \left[ \sup_{\tilde{z} \in \mathcal{Z}} \{\nabla f(z)^\top (\tilde{z} - z) - \lambda\|\tilde{z} - z\|^2\} \right] \right\}$$
$$= -\hbar\rho^2 + \min_{\lambda \geq 0} \left\{ \lambda\rho^2 + \frac{1}{4\lambda} \mathbb{E}_{\mathbb{Q}} \left[ \|\nabla f(z)\|_*^2 \right] \right\}$$
$$= -\hbar\rho^2 + \rho\|\|\nabla f\|_*\|_{\mathbb{Q},2},$$

which completes the proof. □

**Appendix B: Proofs for Section 3**

*Proof of Proposition 1.* Define

$$\Phi(t) := \mathbb{E}_{\mathbb{P}_{\text{true}}}\left[\sup_{\tilde{z} \in \mathcal{Z}} \{t(f(\tilde{z}) - f(z)) - \|\tilde{z} - z\|^p\}\right].$$

Using Theorem 1 in [40] and by our assumptions, a dual optimizer $\lambda_o > 0$ exists and we have strong duality

$$\mathcal{R}_{\mathbb{P}_{\text{true}},p}(\rho) = \lambda_o \rho^p + \lambda_o \Phi(1/\lambda_o). \qquad (4)$$

By [40], $\lambda_o \in [\underline{\lambda}, \infty)$. We first consider the case there exists a dual minimizer $\lambda_o > \underline{\lambda}$, in which case the optimizer is in the interior of the domain of the dual objective. The first-order optimality condition of the convex optimization (1) reads

$$\rho^p + \Phi(1/\lambda_o) \in \frac{1}{\lambda_o} \partial \Phi(1/\lambda_o),$$

where $\partial \Phi$ denotes the subdifferential. Set $\epsilon = \mathcal{R}_{\mathbb{P}_{\text{true}},p}(\rho)$. It follows from the equations above that

$$\epsilon := \lambda_o \rho^p + \lambda_o \Phi(1/\lambda_o) \in \partial \Phi(1/\lambda_o).$$

But by definition $\mathcal{I}_p(\varepsilon; f)^p = \sup_{t>0}\{\varepsilon t - \Phi(t)\}$, which is a concave maximization. This suggests that $t = 1/\lambda_o > 0$ is an optimizer of $\sup_{t>0}\{\epsilon t - \Phi(t)\}$. Hence,

$$\mathcal{I}_p(\mathcal{R}_{\mathbb{P}_{\text{true}},p}(\rho; f); f)^p = \mathcal{I}_p(\epsilon; f)^p = \frac{\epsilon}{\lambda_o} - \Phi(1/\lambda_o) = \frac{1}{\lambda_o}(\lambda_o \rho^p + \lambda_o \Phi(1/\lambda_o)) - \Phi(1/\lambda_o) = \rho^p.$$

Next, consider the other case that the unique dual minimizer $\lambda_o = \underline{\lambda} > 0$. Taking a feasible solution $t = 1/\lambda_o$, using (4) we obtain that

$$\begin{aligned}\mathcal{I}_p(\mathcal{R}_{\mathbb{P}_{\text{true}},p}(\rho; f); f)^p &= \sup_{t>0}\left\{t\mathcal{R}_{\mathbb{P}_{\text{true}},p}(\rho; f) - t\mathbb{E}_{\mathbb{P}_{\text{true}}}\left[\sup_{\tilde{z} \in \mathcal{Z}}\{(f(\tilde{z}) - f(z)) - \frac{1}{t}\|\tilde{z} - z\|^p\}\right]\right\} \\ &\geq \frac{\mathcal{R}_{\mathbb{P}_{\text{true}},p}(\rho; f)}{\lambda_o} - \frac{\mathcal{R}_{\mathbb{P}_{\text{true}},p}(\rho; f) - \lambda_o \rho^p}{\lambda_o} \\ &= \rho^p.\end{aligned}$$

□

The next lemma is mentioned in Section 3.2.

LEMMA 3 **(Corollary 2.4 in Bolley and Villani [19]).** *Assume there exists $a > 0$ such that $C := \log \mathbb{E}_{\mathbb{P}}[\exp(a\|Z\|^2)] < \infty$. Then $\mathbb{P}$ satisfies $T_1(\tau)$, where*

$$\tau = \inf_{\tilde{z} \in \mathcal{Z}, \tilde{a} > 0}\left\{\frac{2}{\tilde{a}}\left(1 + \log \mathbb{E}_{\mathbb{P}}[\exp(\tilde{a}\|Z - \tilde{z}\|^2)]\right)\right\} \leq \frac{2}{a}(1 + C).$$

**B.1. Proof of Theorem 1**

Our proof is based on Marton's argument and Herbst's argument [55, 70]. Let us begin with some definitions and lemmas.

Denote $z^n := (z_1^n, \ldots, z_n^n) \in \mathcal{Z}^n$. We define a product distance $\mathsf{d}_p$ on the space $\mathcal{Z}^n$ as

$$\mathsf{d}_p(z^n, \tilde{z}^n) := \left(\sum_{i=1}^n \|z_i^n - \tilde{z}_i^n\|^p\right)^{1/p},$$

The $p$-Wasserstein distance between probability distributions $\mu$ and $\mathbb{P}_\otimes$ is given by

$$\mathcal{W}_p(\mu, \mathbb{P}_\otimes) = \min_{\pi}\left\{\left(\mathbb{E}_{(z^n, \tilde{z}^n) \sim \pi}[\mathsf{d}_p(z^n, \tilde{z}^n)^p]\right)^{1/p} : \pi \text{ has marginal distributions } \mu, \mathbb{P}_\otimes\right\}.$$

The following tensorization lemma establishes a transportation-information inequality for the product distribution $\mathbb{P}_\otimes$ (see, for example, Proposition 22.5 in [85]).

LEMMA 4. *Let $p \in [1, 2]$. Suppose $\mathbb{P} \in \mathcal{P}_p(\mathcal{Z})$ satisfies $T_p(\tau)$. Then $\mathbb{P}_\otimes$ satisfies $T_p(\tau n^{\frac{2}{p}-1})$.*

Given any function $g : \mathcal{Z} \to \mathbb{R}$ which is exponentially integrable with respect to $\nu$, we define a distribution $\nu^{(g)}$, called the *g-exponential tilting* of $\nu$ as (see, e.g., Section 3.1.2 in [70]):

$$\frac{d\nu^{(g)}}{d\nu} = \frac{\exp(g)}{\mathbb{E}_\nu[\exp(g)]}.$$

It follows that

$$H(\nu^{(g)}||\nu) = \mathbb{E}_{\nu^{(g)}}[g] - \ln \mathbb{E}_\nu[\exp(g)]. \tag{5}$$

We prove below a more general concentration result that applies not only for the empirical mean.

LEMMA 5. *Let $p \in [1, 2]$. Assume $\mathbb{P}_{\text{true}}$ satisfies $T_p(\tau)$. Let $F : \mathcal{Z}^n \to \mathbb{R}$. Assume $\mathbb{E}_\otimes[F] = 0$ and there exist $M, L > 0$ and $z_0^n \in \mathcal{Z}^n$ such that*

$$F(\tilde{z}^n) \leq M + \frac{L}{n} \mathsf{d}_p(\tilde{z}^n, z_0^n)^p, \quad \forall \tilde{z}^n \in \mathcal{Z}^n.$$

*Define $\mathcal{I}(\cdot; F) : \mathbb{R}_+ \to \mathbb{R}_+$ via*

$$\mathcal{I}(\epsilon; F)^p := \sup_{t > 0} \left\{ \epsilon t - \mathbb{E}_\otimes \left[ \sup_{\tilde{z}^n \in \mathcal{Z}^n} \left\{ t(F(\tilde{z}^n) - F(z^n)) - \frac{1}{n} \mathsf{d}_p(\tilde{z}^n, z^n)^p \right\} \right] \right\},$$

*and $\mathcal{R}(\cdot; F) : \mathbb{R}_+ \to \mathbb{R}_+$ as*

$$\mathcal{R}(\rho; F) = \min_{\lambda \geq 0} \left\{ \lambda \rho^p + \mathbb{E}_\otimes \left[ \sup_{\tilde{z}^n \in \mathcal{Z}^n} \left\{ F(\tilde{z}^n) - F(z^n) - \frac{\lambda}{n} \mathsf{d}_p(\tilde{z}^n, z^n)^p \right\} \right] \right\}.$$

*Then for any $\epsilon > 0$,*

$$\mathbb{P}_\otimes \{F(z^n) > \epsilon\} \leq \exp\left(-n\mathcal{I}(\epsilon; F)^2/\tau\right).$$

*Let $t > 0$. Then with probability at least $1 - e^{-t}$,*

$$F(z^n) \leq \mathcal{R}\left(\sqrt{\frac{\tau t}{n}}; F\right).$$

*Proof of Lemma 5.* Define

$$\Phi(t; F) := \mathbb{E}_\otimes \left[ \sup_{\tilde{z}^n \in \mathcal{Z}^n} \left\{ t(F(\tilde{z}^n) - F(z^n)) - \frac{1}{n} \mathsf{d}_p(\tilde{z}^n, z^n)^p \right\} \right],$$

which is in $[0, \infty)$ for all sufficiently small $t$ because of the growth rate condition on $F$, and thus $\mathcal{I}(\epsilon; F)^p = \sup_{t>0}\{\epsilon t - \Phi(t; F)\} > -\infty$. To ensure integrability, let us assume temporarily that $F$ is bounded from above.

We first consider a simpler case $p = 2$. Using Lemma 4, for every $\mu \in \mathcal{P}_2(\mathcal{Z}^n)$, it holds that $\mathcal{W}_2(\mu, \mathbb{P}_\otimes)^2 \leq \tau H(\mu||\mathbb{P}_\otimes)$. Let $t > 0$. Setting $\mu = \mathbb{P}_\otimes^{(\tau t F)}$, by (5) we have

$$\mathcal{W}_2(\mu, \mathbb{P}_\otimes)^2 \leq \mathbb{E}_\mu[\tau t F] - \ln \mathbb{E}_\otimes[\exp(\tau t F)],$$

On the other hand, using Kantorovich's duality (see, e.g., Theorem 5.10 in Villani [85]) and the assumption $\mathbb{E}_\otimes[F] = 0$,

$$\mathcal{W}_2(\mu, \mathbb{P}_\otimes)^2 \geq \mathbb{E}_\mu[\tau t F] + \mathbb{E}_\otimes \left[ \inf_{\tilde{z}^n \in \mathcal{Z}^n} \left\{ \sum_{i=1}^n \|\tilde{z}_i^n - z_i^n\|^2 - \tau t F(\tilde{z}^n) \right\} \right]$$

$$= \mathbb{E}_\mu[\tau t F] + \mathbb{E}_\otimes \left[ \inf_{\tilde{z}^n \in \mathcal{Z}^n} \left\{ \sum_{i=1}^n \|\tilde{z}_i^n - z_i^n\|^2 - \tau t (F(\tilde{z}^n) - F(z^n)) \right\} \right]$$

$$= \mathbb{E}_\mu[\tau t F] - n\Phi(\frac{\tau t}{n}; F).$$

Combining the inequalities above and canceling out the term $\mathbb{E}_\mu[\tau t F]$, we obtain

$$\tau \ln \mathbb{E}_\otimes[\exp(\tau t F)] \le n\Phi(\frac{\tau t}{n}; F).$$

Using Markov's inequality, for all $t > 0$,

$$\begin{aligned}\mathbb{P}_\otimes\{F(z^n) > \epsilon\} &= \mathbb{P}_\otimes\left\{\tau t F(z^n) > \tau t \epsilon\right\} \\ &\le \mathbb{E}_\otimes\left[\exp(\tau t F)\right]/\exp(\tau t \epsilon) \\ &\le \exp\left(\frac{n}{\tau}\Phi(\frac{\tau t}{n}; F) - \tau t \epsilon\right).\end{aligned}$$

Mapping $\tau t/n$ to $t$ and minimizing over $s, t > 0$ yields

$$\mathbb{P}_\otimes\{F(z^n) > \epsilon\} \le \exp\left(\frac{n}{\tau}\inf_{t>0}\left\{\Phi(t;F) - t\epsilon\right\}\right) = \exp\left(-n\mathcal{I}(\epsilon;F)^2/\tau\right).$$

Next, we consider $p \in [1, 2)$. Let $s > 0$. Using Lemma 4, for every $\mu \in \mathcal{P}_p(\mathcal{Z}^n)$, it holds that

$$\mathcal{W}_p(\mu, \mathbb{P}_\otimes)^p \le \left(\tau n^{\frac{2}{p}-1} H(\mu\|\mathbb{P}_\otimes)\right)^{\frac{p}{2}} = \left(\tau^{\frac{p}{2}}(\frac{2}{p})^{\frac{p}{2}} s^{1-\frac{p}{2}} H(\mu\|\mathbb{P}_\otimes)^{\frac{p}{2}}\right)\left((\frac{p}{2})^{\frac{p}{2}} s^{\frac{p}{2}-1} n^{1-\frac{p}{2}}\right).$$

Applying Young's inequality to the right side yields that

$$\begin{aligned}\mathcal{W}_p(\mu, \mathbb{P}_\otimes)^p &\le \frac{p}{2}\left(\tau^{\frac{p}{2}}(\frac{2}{p})^{\frac{p}{2}} s^{1-\frac{p}{2}} H(\mu\|\mathbb{P}_\otimes)^{\frac{p}{2}}\right)^{\frac{2}{p}} + (1-\frac{p}{2})\left((\frac{p}{2})^{\frac{p}{2}} s^{\frac{p}{2}-1} n^{1-\frac{p}{2}}\right)^{\frac{1}{1-\frac{p}{2}}} \\ &= s^{\frac{2}{p}-1}\tau H(\mu\|\mathbb{P}_\otimes) + (1-\frac{p}{2})(\frac{p}{2})^{\frac{p}{2-p}} s^{-1} n,\end{aligned}$$

Let $t > 0$. Setting $\mu = \mathbb{P}_\otimes^{(\tau s^{1-\frac{2}{p}} t F)}$, by (5) we have

$$\mathcal{W}_p(\mu, \mathbb{P}_\otimes)^p \le s^{\frac{2}{p}-1}\left(\mathbb{E}_\mu\left[\tau s^{1-\frac{2}{p}} t F\right] - \ln \mathbb{E}_\otimes\left[\exp\left(\tau s^{1-\frac{2}{p}} t F\right)\right]\right) + (1-\frac{p}{2})(\frac{p}{2})^{\frac{p}{2-p}} s^{-1} n.$$

On the other hand, using Kantorovich's duality (see, e.g., Theorem 5.10 in Villani [85]) and the assumption $\mathbb{E}_\otimes[F] = 0$,

$$\begin{aligned}\mathcal{W}_p(\mu, \mathbb{P}_\otimes)^p &\ge \mathbb{E}_\mu[\tau t F] + \mathbb{E}_\otimes\left[\inf_{\tilde{z}^n \in \mathcal{Z}^n}\left\{\sum_{i=1}^n \|\tilde{z}_i^n - z_i^n\|^p - \tau t F(\tilde{z}^n)\right\}\right] \\ &= \mathbb{E}_\mu[\tau t F] + \mathbb{E}_\otimes\left[\inf_{\tilde{z}^n \in \mathcal{Z}^n}\left\{\sum_{i=1}^n \|\tilde{z}_i^n - z_i^n\|^p - \tau t (F(\tilde{z}^n) - F(z^n))\right\}\right] \\ &= \mathbb{E}_\mu[\tau t F] - n\Phi(\frac{\tau t}{n}; F).\end{aligned}$$

Combining the inequalities above and canceling out the term $\mathbb{E}_\mu[\tau t F]$, we obtain

$$s^{\frac{2}{p}-1}\tau \ln \mathbb{E}_\otimes[\exp(\tau s^{1-\frac{2}{p}} t F)] \le (1-\frac{p}{2})(\frac{p}{2})^{\frac{p}{2-p}} s^{-1} n + n\Phi(\frac{\tau t}{n}; F).$$

Using Markov's inequality, for all $s, t > 0$,

$$\begin{aligned}\mathbb{P}_\otimes\{F(z^n) > \epsilon\} &= \mathbb{P}_\otimes\left\{\tau s^{1-\frac{2}{p}} t F(z^n) > \tau s^{1-\frac{2}{p}} t \epsilon\right\} \\ &\le \mathbb{E}_\otimes\left[\exp(\tau s^{1-\frac{2}{p}} t F)\right]/\exp(\tau s^{1-\frac{2}{p}} t \epsilon) \\ &\le \exp\left\{\left((1-\frac{p}{2})(\frac{p}{2})^{\frac{p}{2-p}} s^{-\frac{2}{p}} n + s^{1-\frac{2}{p}} n\Phi(\frac{\tau t}{n}; F)\right)/\tau - \tau s^{1-\frac{2}{p}} t \epsilon\right\}.\end{aligned}$$

Mapping $\tau t/n$ to $t$ and minimizing over $s, t > 0$ yields

$$\mathbb{P}_\otimes\{F(z^n) > \epsilon\} \leq \exp\left(\frac{n}{\tau}\inf_{s,t>0}\left\{(1-\tfrac{p}{2})(\tfrac{p}{2})^{\frac{p}{2-p}}s^{-\frac{2}{p}} + s^{1-\frac{2}{p}}(\Phi(t;F) - t\epsilon)\right\}\right)$$

$$= \exp\left(\frac{n}{\tau}\inf_{s>0}\left\{(1-\tfrac{p}{2})(\tfrac{p}{2})^{\frac{p}{2-p}}s^{-\frac{2}{p}} - s^{1-\frac{2}{p}}\sup_{t>0}\{t\epsilon/\tau - \Phi(t;F)\}\right\}\right)$$

$$= \exp\left(\frac{n}{\tau}\inf_{s>0}\left\{(1-\tfrac{p}{2})(\tfrac{p}{2})^{\frac{p}{2-p}}s - s^{1-\frac{p}{2}}\mathcal{I}(\epsilon/\tau;F)^p\right\}\right)$$

$$= \exp\left(-n\mathcal{I}(\epsilon;F)^2/\tau\right).$$

Setting $\rho = \sqrt{\frac{\tau t}{n}}$. If the dual minimizer defining $\mathcal{R}(\rho;F)$ is strictly positive, taking $\epsilon = \mathcal{R}(\rho;F)$ and applying Proposition 1 yields the second part of the result. Otherwise, $\mathcal{R}(\rho;F) = \sup_{\tilde{z}^n \in \mathcal{Z}^n} F(\tilde{z}^n)$, thereby $F(z^n) \leq \mathcal{R}(\rho;F)$. To deal with an unbounded $F$, define $F_k = F \wedge k$ for $k \in \mathbb{N}_{\geq 1}$. We have proved that the result holds for $F_k$. Observe that for all $z^n, \tilde{z}^n \in \mathcal{Z}$ with $F(\tilde{z}^n) \geq F(z^n)$, it holds that

$$(F(\tilde{z}^n) \wedge k) - (F(z^n) \wedge k) = \begin{cases} 0, & F(z^n) \geq k, \\ k - F(z^n), & F(z^n) < k < F(\tilde{z}^n), \\ F(\tilde{z}^n) - F(z^n), & F(\tilde{z}^n) \leq k. \end{cases}$$

Hence for all $t > 0$ and $k \geq 1$,

$$\Phi(t;F_k) = \mathbb{E}_\otimes\left[\sup_{\tilde{z}^n \in \mathcal{Z}^n: F(\tilde{z}^n) \geq F(z^n)}\left\{t(F_k(\tilde{z}^n) - F_k(z^n)) - \frac{1}{n}\mathsf{d}_p(\tilde{z}^n, z^n)^p\right\}\right]$$

$$\leq \mathbb{E}_\otimes\left[\sup_{\tilde{z}^n \in \mathcal{Z}^n: F(\tilde{z}^n) \geq F(z^n)}\left\{t(F(\tilde{z}^n) - F(z^n)) - \frac{1}{n}\mathsf{d}_p(\tilde{z}^n, z^n)^p\right\}\right],$$

and thus $\mathcal{I}(\epsilon;F_k) \geq \mathcal{I}(\epsilon;F)$. Therefore, by the monotone convergence,

$$\mathbb{P}_\otimes\{F(z^n) > \epsilon\} = \lim_{k \to \infty}\mathbb{P}_\otimes\{F_k(z^n) > \epsilon\} \leq \lim_{k \to \infty}\exp\left(-n\mathcal{I}(\epsilon;F_k)^2/\tau\right) \leq \exp\left(-n\mathcal{I}(\epsilon;F)^2/\tau\right),$$

which completes the proof. $\square$

*Proof of Theorem 1.* Set $F(z^n) = \mathbb{E}_{\mathbb{P}_{\text{true}}}[f(z)] - \mathbb{E}_{\mathbb{P}_n}[f]$. Then $F$ satisfies the assumptions in Lemma 5 due to Assumptions 1(I) and 2. Applying Lemma 5 yields that

$$\Phi(t;F) = \frac{1}{n}\mathbb{E}_\otimes\left[\sup_{\tilde{z}^n \in \mathcal{Z}^n}\left\{\sum_{i=1}^n\left(t(f(z_i^n) - f(\tilde{z}_i^n)) - \|\tilde{z}_i^n - z_i^n\|^p\right)\right\}\right]$$

$$= \mathbb{E}_{\mathbb{P}_{\text{true}}}\left[\sup_{\tilde{z} \in \mathcal{Z}}\left\{-t(f(\tilde{z}) - f(z)) - \|\tilde{z} - z\|^p\right\}\right],$$

and thus $\mathcal{I}(\cdot;F) = \mathcal{I}_p(\cdot;-f)$ and $\mathcal{R}(\cdot;F) = \mathcal{R}_{\mathbb{P}_{\text{true}},p}(\cdot;-f)$, therefore the result follows. $\square$

**Appendix C: Proofs for Section 4**

**C.1. Proof for Section 4.1**

*Proof of Corollary 2.* (I) By Assumption 3, for any distribution $\mathbb{P}$ it holds that

$$\mathbb{E}_\mathbb{P}[f_{\tilde{\theta}}] - \mathbb{E}_\mathbb{P}[f_\theta] \leq \mathbb{E}_\mathbb{P}[\kappa] \cdot \|\tilde{\theta} - \theta\|_\Theta.$$

Thus

$$\mathcal{R}_{\mathbb{P}_{\text{true}},p}\left(\sqrt{\tfrac{\tau t}{n}};-f_{\tilde{\theta}}\right) - \mathcal{R}_{\mathbb{P}_{\text{true}},p}\left(\sqrt{\tfrac{\tau t}{n}};-f_\theta\right) \leq \sup_{\mathbb{P}:\mathcal{W}_p(\mathbb{P},\mathbb{P}_{\text{true}}) \leq \sqrt{\tfrac{\tau t}{n}}}\left|\mathbb{E}_{\mathbb{P}_{\text{true}}}[f_{\tilde{\theta}}] - \mathbb{E}_{\mathbb{P}_{\text{true}}}[f_\theta]\right|$$

$$\leq \sup_{\mathbb{P}:\mathcal{W}_p(\mathbb{P},\mathbb{P}_{\text{true}}) \leq \sqrt{\tfrac{\tau t}{n}}} \mathbb{E}_\mathbb{P}[\kappa] \cdot \|\tilde{\theta} - \theta\|_\Theta.$$

Note from the duality (1) and Assumption 3 that

$$\sup_{\mathbb{P}:\mathcal{W}_p(\mathbb{P},\mathbb{P}_{\text{true}})\leq\sqrt{\frac{\tau t}{n}}} \mathbb{E}_{\mathbb{P}}[\kappa] = \min_{\lambda\geq 0}\left\{\lambda(\sqrt{\tfrac{\tau t}{n}})^p + \mathbb{E}_{\mathbb{P}_{\text{true}}}\left[\sup_{\tilde{z}\in\mathcal{Z}}\{\kappa(\tilde{z}) - \lambda\|\tilde{z}-z\|^p\}\right]\right\}$$

$$\leq \min_{\lambda\geq 0}\left\{\lambda(\sqrt{\tfrac{\tau t}{n}})^p + \mathbb{E}_{\mathbb{P}_{\text{true}}}\left[\sup_{\tilde{z}\in\mathcal{Z}}\{\kappa_M + \kappa_L\|\tilde{z}\|^p - \lambda\|\tilde{z}-z\|^p\}\right]\right\}$$

$$\leq \min_{\lambda\geq 0}\left\{\lambda(\sqrt{\tfrac{\tau t}{n}})^p + \mathbb{E}_{\mathbb{P}_{\text{true}}}\left[\sup_{\tilde{z}\in\mathcal{Z}}\{\kappa_M + \kappa_L(\|\tilde{z}-z\| + \|z\|)^p - \lambda\|\tilde{z}-z\|^p\}\right]\right\}$$

$$\leq \min_{\lambda\geq 0}\left\{\lambda(\sqrt{\tfrac{\tau t}{n}})^p + \mathbb{E}_{\mathbb{P}_{\text{true}}}\left[\sup_{\tilde{z}\in\mathcal{Z}}\{\kappa_M + \kappa_L 2^{p-1}(\|\tilde{z}-z\|^p + \|z\|^p) - \lambda\|\tilde{z}-z\|^p\}\right]\right\}$$

$$\leq 2^{p-1}\kappa_L(\sqrt{\tfrac{\tau t}{n}})^p + \kappa_M + 2^{p-1}\kappa_L\mathbb{E}_{\mathbb{P}_{\text{true}}}[\|z\|^p]$$

$$=: \bar{\kappa},$$

where the last inequality holds by taking a dual feasible solution $\lambda = 2^{p-1}\kappa_L$. By the assumption on $\kappa$ and Chebyshev's inequality,

$$\mathbb{P}_{\otimes}\left\{\mathbb{E}_{\mathbb{P}_n}[\kappa] - \mathbb{E}_{\mathbb{P}_{\text{true}}}[\kappa] > \sqrt{\mathbb{V}\text{ar}_{\mathbb{P}_{\text{true}}}[\kappa]}\right\} \leq \frac{1}{n}.$$

Let $\epsilon > 0$ and $\Theta_\epsilon$ be an $\epsilon$-cover of $\Theta$. We have that

$$\mathbb{P}_{\otimes}\left\{\exists \theta \in \Theta, s.t.\ \mathbb{E}_{\mathbb{P}_{\text{true}}}[f_\theta] > \mathbb{E}_{\mathbb{P}_n}[f_\theta] + \mathcal{R}_{\mathbb{P}_{\text{true}},p}\left(\sqrt{\tfrac{\tau t}{n}}; -f_\theta\right) + \epsilon \cdot (2\mathbb{E}_{\mathbb{P}_{\text{true}}}[\kappa] + \sqrt{\mathbb{V}\text{ar}_{\mathbb{P}_{\text{true}}}[\kappa]} + \bar{\kappa})\right\}$$

$$\leq \frac{1}{n} + \mathbb{P}_{\otimes}\left\{\exists \theta \in \Theta, s.t.\ \mathbb{E}_{\mathbb{P}_{\text{true}}}[f_\theta] > \mathbb{E}_{\mathbb{P}_n}[f_\theta] + \mathcal{R}_{\mathbb{P}_{\text{true}},p}\left(\sqrt{\tfrac{\tau t}{n}}; -f_\theta\right) + \epsilon \cdot (\mathbb{E}_{\mathbb{P}_{\text{true}}}[\kappa] + \mathbb{E}_{\mathbb{P}_n}[\kappa] + \bar{\kappa})\right\}$$

$$\leq \frac{1}{n} + \mathbb{P}_{\otimes}\left\{\exists \tilde{\theta} \in \Theta_\epsilon, s.t.\ \mathbb{E}_{\mathbb{P}_{\text{true}}}[f_{\tilde{\theta}}] > \mathbb{E}_{\mathbb{P}_n}[f_{\tilde{\theta}}] + \mathcal{R}_{\mathbb{P}_{\text{true}},p}\left(\sqrt{\tfrac{\tau t}{n}}; -f_{\tilde{\theta}}\right)\right\}$$

$$\leq \frac{1}{n} + \sum_{\tilde{\theta}\in\Theta_\epsilon} \mathbb{P}_{\otimes}\left\{\mathbb{E}_{\mathbb{P}_{\text{true}}}[f_{\tilde{\theta}}] > \mathbb{E}_{\mathbb{P}_n}[f_{\tilde{\theta}}] + \mathcal{R}_{\mathbb{P}_{\text{true}},p}\left(\sqrt{\tfrac{\tau t}{n}}; -f_{\tilde{\theta}}\right)\right\}$$

$$\leq \frac{1}{n} + \mathcal{N}(\epsilon; \Theta, \|\cdot\|_\Theta) \cdot e^{-t},$$

where we have used the definition of the covering number and Theorem 1 in the last step. Letting $\epsilon = 1/n$ and replacing $t$ with $t + \log \mathcal{N}(1/n; \Theta, \|\cdot\|_\Theta)$ yields the result.

(II)(III) are simple consequences of (I), together with Lemma 1 and Lemma 2. □

*Proof of Corollary 3.* Fix $f \in \mathcal{F}$. Applying Bennett's inequality (Lemma 6 below) to $X_i = -\frac{\|\nabla f(z_i^n)\|_*^2}{n\|\|\nabla f\|_*\|_{\mathbb{P}_{\text{true}},2}^2}$, $b = 0$ and $v_i = \sigma^2/n^2$, we obtain that

$$\mathbb{P}\left\{\mathbb{E}_{\mathbb{P}_n}\left[\frac{\|\nabla f\|_*^2}{\|\|\nabla f\|_*\|_{\mathbb{P}_{\text{true}},2}^2}\right] - 1 < -\epsilon\right\} \leq \exp\left(-\frac{n\epsilon^2}{2\sigma^2}\right).$$

Hence, with probability at least $1 - e^{-t}$,

$$\frac{\|\|\nabla f\|_*\|_{\mathbb{P}_n,2}^2}{\|\|\nabla f\|_*\|_{\mathbb{P}_{\text{true}},2}^2} \geq 1 - \sigma\sqrt{\frac{2t}{n}}.$$

Thus, for every $n > 8\sigma^2 t$,

$$\|\|\nabla f\|_*\|_{\mathbb{P}_{\text{true}},2} \leq \|\|\nabla f\|_*\|_{\mathbb{P}_n,2}\left(1 - \sigma\sqrt{\tfrac{2t}{n}}\right)^{-\frac{1}{2}} \leq \|\|\nabla f\|_*\|_{\mathbb{P}_n,2}\left(1 + \sigma\sqrt{\tfrac{2t}{n}}\right),$$

where the second inequality follows from the simple fact $1/\sqrt{1-a} \leq 1+a$ for $a \in [0, 1/2]$.

Next we consider a family of losses. By Assumptions 3 and 4, it holds that

$$\mathbb{E}_{\mathbb{P}_{\text{true}}}[f_{\tilde{\theta}}] - \mathbb{E}_{\mathbb{P}_{\text{true}}}[f_\theta] \leq \mathbb{E}_{\mathbb{P}_{\text{true}}}[\kappa] \cdot \|\tilde{\theta} - \theta\|_\Theta,$$
$$\mathbb{E}_{\mathbb{P}_n}[f_{\tilde{\theta}}] - \mathbb{E}_{\mathbb{P}_n}[f_\theta] \leq \mathbb{E}_{\mathbb{P}_n}[\kappa] \cdot \|\tilde{\theta} - \theta\|_\Theta,$$
$$\|\|\nabla f_{\tilde{\theta}}\|_*\|_{\mathbb{P}_n,2} - \|\|\nabla f_\theta\|_*\|_{\mathbb{P}_n,2} \leq \|\kappa_2\|_{\mathbb{P}_n,2} \cdot \|\tilde{\theta} - \theta\|_\Theta.$$

By Chebyshev's inequality,

$$\mathbb{P}_\otimes\left\{\mathbb{E}_{\mathbb{P}_n}[\kappa_2^2] - \mathbb{E}_{\mathbb{P}_{\text{true}}}[\kappa_2^2] > \sqrt{\mathbb{V}\text{ar}_{\mathbb{P}_{\text{true}}}[\kappa_2^2]}\right\} \leq \frac{1}{n}.$$

Let $\epsilon > 0$ and let $\Theta_\epsilon$ be an $\epsilon$-cover of $\Theta$. Set $\rho = \sqrt{\frac{\tau t}{n}}(1 + \sigma\sqrt{\frac{2t}{n}})$. It follows that

$$\mathbb{P}_\otimes\Big\{\exists \theta \in \Theta, s.t.\ \mathbb{E}_{\mathbb{P}_{\text{true}}}[f_\theta] > \mathbb{E}_{\mathbb{P}_n}[f_\theta] + \rho\|\|\nabla f_\theta\|_*\|_{\mathbb{P}_n,2} + \frac{\hbar\tau t}{n}$$
$$+ \epsilon \cdot (2\mathbb{E}_{\mathbb{P}_{\text{true}}}[\kappa] + \sqrt{\mathbb{V}\text{ar}_{\mathbb{P}_{\text{true}}}[\kappa]} + \rho\sqrt{\mathbb{E}_{\mathbb{P}_{\text{true}}}[\kappa_2^2] + \sqrt{\mathbb{V}\text{ar}_{\mathbb{P}_{\text{true}}}[\kappa_2^2]}})\Big\}$$
$$\leq \frac{2}{n} + \mathbb{P}_\otimes\Big\{\exists \theta \in \Theta, s.t.\ \mathbb{E}_{\mathbb{P}_{\text{true}}}[f_\theta] > \mathbb{E}_{\mathbb{P}_n}[f_\theta] + \rho\|\|\nabla f_\theta\|_*\|_{\mathbb{P}_n,2} + \frac{\hbar\tau t}{n} + \epsilon \cdot (\mathbb{E}_{\mathbb{P}_{\text{true}}}[\kappa] + \mathbb{E}_{\mathbb{P}_n}[\kappa] + \rho\|\kappa_2\|_{\mathbb{P}_n,2})\Big\}$$
$$\leq \frac{2}{n} + \mathbb{P}_\otimes\Big\{\exists \tilde{\theta} \in \Theta_\epsilon, s.t.\ \mathbb{E}_{\mathbb{P}_{\text{true}}}[f_{\tilde{\theta}}] > \mathbb{E}_{\mathbb{P}_n}[f_{\tilde{\theta}}] + \rho\|\|\nabla f_{\tilde{\theta}}\|_*\|_{\mathbb{P}_n,2} + \frac{\hbar\tau t}{n}\Big\}$$
$$\leq \frac{2}{n} + e^{-t} + \sum_{\tilde{\theta} \in \Theta_\epsilon} \mathbb{P}_\otimes\Big\{\mathbb{E}_{\mathbb{P}_{\text{true}}}[f_{\tilde{\theta}}] > \mathbb{E}_{\mathbb{P}_n}[f_{\tilde{\theta}}] + \sqrt{\frac{\tau t}{n}}\|\|\nabla f_{\tilde{\theta}}\|_*\|_{\mathbb{P}_{\text{true}}, 2} + \frac{\hbar\tau t}{n}\Big\}$$
$$\leq \frac{2}{n} + (1 + \mathcal{N}(\epsilon, \Theta, \|\cdot\|_\Theta)) \cdot e^{-t},$$

where we have used Corollary 1(III) in the last inequality. Hence the proof is completed by setting $\epsilon = 1/n$, replacing $t$ with $t + \log(1 + \mathcal{N}(1/n; \Theta, \|\cdot\|_\Theta))$, and invoking Lemma 2 for the second part. $\square$

LEMMA 6 (**Bennett's inequality**). *Suppose $X_1, \ldots, X_n$ are independent random variables for which $X_i \leq b$ and $\mathbb{E}[X_i^2] \leq v_i$ for each $i$, for nonnegative constants $b$ and $v_i$. Let $W = \sum_{i=1}^n v_i$. Then for $\epsilon \geq 0$.*

$$\mathbb{P}\left\{\sum_{i=1}^n (X_i - \mathbb{E}[X_i]) \geq \epsilon\right\} \leq \exp\left(-\frac{\epsilon^2}{2W}\psi_{\text{Benn}}\left(\frac{b\epsilon}{W}\right)\right),$$

*where $\psi_{\text{Benn}}$ denotes the function defined on $[-1, \infty)$ by*

$$\psi_{\text{Benn}}(t) := \frac{(1+t)\log(1+t) - t}{t^2/2}, \text{ for } t \neq 0, \text{ and } \psi_{\text{Benn}}(0) = 1.$$

### C.2. Proofs for Section 4.2

#### C.2.1. Auxiliary Results
We prepare some auxiliary results that will be used shortly. The following two lemmas are useful properties of Rademacher complexity (see, e.g., [77, Chapter 26]).

LEMMA 7 (**Symmetrization**). *Let $\mathcal{H}$ be a family of functions. Then*

$$\mathbb{E}_\otimes\left[\sup_{h \in \mathcal{H}}\{\mathbb{E}_{\mathbb{P}_{\text{true}}}[h] - \mathbb{E}_{\mathbb{P}_n}[h]\}\right] \leq 2\mathbb{E}_\otimes[\mathfrak{R}_n(\mathcal{H})].$$

LEMMA 8 (**Contraction**). *Let $\mathcal{H}$ be a family of functions. Let $\ell : \mathbb{R} \to \mathbb{R}$ be a Lipschitz function. Denote $\ell \circ \mathcal{H} = \{\ell \circ h : h \in \mathcal{H}\}$. Then*

$$\mathbb{E}_\otimes[\mathfrak{R}_n(\ell \circ \mathcal{H})] \leq \|\ell\|_{\text{Lip}} \cdot \mathbb{E}_\otimes[\mathfrak{R}_n(\mathcal{H})].$$

Let us define

$$\mathcal{R}_{\otimes,p}(\rho;\mathcal{F}) := \min_{\lambda \geq 0}\left\{\lambda\rho^p + \mathbb{E}_{\otimes}\left[\sup_{f \in \mathcal{F}, \tilde{z}^n \in \mathcal{Z}^n} \frac{1}{n}\sum_{i=1}^n \left(f(\tilde{z}_i^n) - f(z_i^n) - \lambda\|\tilde{z}_i^n - z_i^n\|^p\right)\right]\right\},$$

and

$$-\mathcal{F} := \{-f : f \in \mathcal{F}\},$$

LEMMA 9. *Let $p \in \{1,2\}$. Assume Assumption 1(I) holds when $p = 1$ and Assumptions 2 when $p = 2$. Assume $\mathbb{P}_{\text{true}}$ satisfies $T_p(\tau)$. Let $t > 0$. Then with probability at least $1 - e^{-t}$, for every $f \in \mathcal{F}$,*

$$\mathbb{E}_{\mathbb{P}_{\text{true}}}[f] \leq \mathbb{E}_{\mathbb{P}_n}[f] + \mathcal{R}_{\otimes,p}(\sqrt{\frac{\tau t}{n}}; -\mathcal{F}) + 2\mathbb{E}_{\otimes}[\mathfrak{R}_n(\mathcal{F})].$$

*Proof.* Set

$$F(z^n) = \sup_{f \in \mathcal{F}}\left\{\mathbb{E}_{\mathbb{P}_{\text{true}}}[f] - \mathbb{E}_{\mathbb{P}_n}[f]\right\} - \mathbb{E}_{\otimes}\left[\sup_{f \in \mathcal{F}}\left\{\mathbb{E}_{\mathbb{P}_{\text{true}}}[f] - \mathbb{E}_{\mathbb{P}_n}[f]\right\}\right].$$

Then the assumption on $f$ implies that $F$ satisfies the growth assumptions in Lemma 5. Applying Lemma 5 yields that with probability at least $1 - e^{-t}$, for every $f \in \mathcal{F}$,

$$\mathbb{E}_{\mathbb{P}_{\text{true}}}[f] - \mathbb{E}_{\mathbb{P}_n}[f] - \mathbb{E}_{\otimes}\left[\sup_{f \in \mathcal{F}}\left\{\mathbb{E}_{\mathbb{P}_{\text{true}}}[f] - \mathbb{E}_{\mathbb{P}_n}[f]\right\}\right]$$

$$\leq \inf_{\lambda > 0}\left\{\lambda(\sqrt{\frac{\tau t}{n}})^p + \mathbb{E}_{\otimes}\left[\sup_{\tilde{z}^n \in \mathcal{Z}^n}\left\{F(\tilde{z}^n) - F(z^n) - \frac{\lambda}{n}\mathsf{d}_p(\tilde{z}^n, z^n)^p\right\}\right]\right\}$$

$$\leq \inf_{\lambda > 0}\left\{\lambda(\sqrt{\frac{\tau t}{n}})^p + \mathbb{E}_{\otimes}\left[\sup_{\tilde{z}^n \in \mathcal{Z}^n}\sup_{f \in \mathcal{F}}\left\{\frac{1}{n}\sum_{i=1}^n -f(\tilde{z}_i^n) + f(z_i^n) - \lambda\|\tilde{z}_i^n - z_i^n\|^p\right\}\right]\right\}$$

$$= \mathcal{R}_{\otimes,p}(\sqrt{\frac{\tau t}{n}}; -\mathcal{F}).$$

Thus the result then follows by applying Lemma 7. □

The next lemma is used for bounding the fixed point of $\psi_n$ for local Rademacher complexity.

LEMMA 10 ([6], p. 1512). *Let $A, B > 0$. Let $r_0$ be the largest solution to the equation $B\sqrt{r} + A = r$. Then $B^2 \leq r_0 \leq 2A + B^2$.*

**C.2.2. Proofs for $p = 1$** To begin with, an application of Lemma 5 to the loss function $F = \sup_{f \in \mathcal{F}}\left\{\mathbb{E}_{\mathbb{P}_{\text{true}}}[f] - \mathbb{E}_{\mathbb{P}_n}[f]\right\} - \mathbb{E}_{\otimes}\left[\sup_{f \in \mathcal{F}}\left\{\mathbb{E}_{\mathbb{P}_{\text{true}}}[f] - \mathbb{E}_{\mathbb{P}_n}[f]\right\}\right]$ yields the following result.

LEMMA 11. *Assume $\mathbb{P}_{\text{true}}$ satisfies $T_1(\tau)$ and Assumption 1(I) holds. Let $t > 0$. Then with probability at least $1 - e^{-t}$,*

$$\mathbb{E}_{\mathbb{P}_{\text{true}}}[f] \leq \mathbb{E}_{\mathbb{P}_n}[f] + \sqrt{\frac{\tau t}{n}}\sup_{f \in \mathcal{F}}\|f\|_{\text{Lip}} + 2\mathbb{E}_{\otimes}[\mathfrak{R}_n(\mathcal{F})], \quad \forall f \in \mathcal{F}.$$

*Proof of Lemma 11.* In view of Lemma 9, it suffices to derive an upper bound on $\mathcal{R}_{\otimes,1}(\rho; -\mathcal{F})$. Assumption 1(I) implies that for any $\lambda > \sup_{f \in \mathcal{F}}\|f\|_{\text{Lip}}$,

$$\sup_{f \in (-\mathcal{F}), \tilde{z}^n \in \mathcal{Z}^n}\left\{\frac{1}{n}\sum_{i=1}^n f(\tilde{z}_i^n) - f(z_i^n) - \lambda\|\tilde{z}_i^n - z_i^n\|\right\} = 0.$$

Consequently by definition $\mathcal{R}_{\otimes,1}(\rho; -\mathcal{F}) \leq \rho \sup_{f \in \mathcal{F}}\|f\|_{\text{Lip}}$. □

Next, using the *peeling technique* [84, 47], we can remove the dependence on $\sup_{f \in \mathcal{F}}$ of the right side of the inequality in Lemma 11.

LEMMA 12. *Assume* $\mathbb{P}_{\text{true}}$ *satisfies* $T_1(\tau)$ *and Assumption* 1(I) *holds. Let* $t > 0$. *Then with probability at least* $1 - \lceil \log_2(\sqrt{\gamma_1 \tau t n}) \rceil e^{-t}$,

$$\mathbb{E}_{\mathbb{P}_{\text{true}}}[f] \leq \mathbb{E}_{\mathbb{P}_n}[f] + 2\sqrt{\frac{\tau t}{n}} \|f\|_{\text{Lip}} + 2\mathbb{E}_\otimes[\mathfrak{R}_n(\mathcal{F})] + \frac{1}{n}, \quad \forall f \in \mathcal{F}.$$

*Proof of Lemma 12.* Set $r = \sup_{f \in \mathcal{F}} \|f\|_{\text{Lip}} \leq \gamma_1$. Let $K$ be a positive integer whose value will be specified shortly. We define

$$\mathcal{F}_k := \{f \in \mathcal{F}: \ 2^{-k}r < \|f\|_{\text{Lip}} \leq 2^{-k+1}r\}, \quad 1 \leq k \leq K-1,$$
$$\mathcal{F}_K := \{f \in \mathcal{F}: \ \|f\|_{\text{Lip}} \leq 2^{-K}r\}.$$

Using Lemma 11, for $k = 1, \ldots, K-1$, with probability at least $1 - e^{-t}$, for every $f \in \mathcal{F}_k$,

$$\mathbb{E}_{\mathbb{P}_{\text{true}}}[f] - \mathbb{E}_{\mathbb{P}_n}[f] \leq \sqrt{\frac{\tau t}{n}} 2^{-k+1}r + 2\mathbb{E}_\otimes[\mathfrak{R}_n(\mathcal{F}_k)] \leq 2\sqrt{\frac{\tau t}{n}} \|f\|_{\text{Lip}} + 2\mathbb{E}_\otimes[\mathfrak{R}_n(\mathcal{F}_k)],$$

and with probability at least $1 - e^{-t}$, for every $f \in \mathcal{F}_K$,

$$\mathbb{E}_{\mathbb{P}_{\text{true}}}[f] - \mathbb{E}_{\mathbb{P}_n}[f] \leq \sqrt{\frac{\tau t}{n}} 2^{-K}r + 2\mathbb{E}_\otimes[\mathfrak{R}_n(\mathcal{F}_K)].$$

Taking the union bound, with probability at least $1 - Ke^{-t}$, for every $f \in \mathcal{F}$,

$$\mathbb{E}_{\mathbb{P}_{\text{true}}}[f] \leq \mathbb{E}_{\mathbb{P}_n}[f] + 2\sqrt{\frac{\tau t}{n}} \|f\|_{\text{Lip}} + 2\mathbb{E}_\otimes[\mathfrak{R}_n(\mathcal{F})] + \sqrt{\frac{\tau t}{n}} 2^{-K}r.$$

Note that $r \leq \gamma_1$ by Assumption 1(I). Setting $K = \lceil \log_2(\gamma_1 \sqrt{\tau t n}) \rceil$ yields the result. $\square$

Often, $\mathbb{E}_\otimes[\mathfrak{R}_n(\mathcal{F})]$ is of the order of $1/\sqrt{n}$. By applying Lemma 12 and another peeling argument to a weighted function class $\{\frac{\sqrt{r}}{\sqrt{r} \vee \|f\|_{\text{Lip}}} f : \ f \in \mathcal{F}\}$ and using the sub-root property of $\psi_n$, we can replace $\mathbb{E}_\otimes[\mathfrak{R}_n(\mathcal{F})]$ with the fixed point $r_{n\star}$ of $\psi_n$, often in the order of $1/n$.

*Proof of Theorem 2.* Let $r \geq r_{n\star}$ whose value will be specified shortly. The sub-root assumption on $\psi_n$ implies that

$$\psi_n(r) = \frac{\sqrt{r}\psi_n(r)}{\sqrt{r}} \leq \frac{\sqrt{r}\psi_n(r_{n\star})}{\sqrt{r_{n\star}}} = \sqrt{r r_{n\star}}.$$

Define

$$\mathcal{F}_r := \left\{ \frac{\sqrt{r}}{\sqrt{r} \vee \|f\|_{\text{Lip}}} f : \ f \in \mathcal{F} \right\}.$$

Then $\|g\|_{\text{Lip}} \leq \sqrt{r}$ for all $g \in \mathcal{F}_r$ and $\mathcal{F}_{\gamma_1^2} = \mathcal{F}$, thus

$$\mathbb{E}_\otimes[\mathfrak{R}_n(\mathcal{F}_r)] \leq \mathbb{E}_\otimes\big[\mathfrak{R}(\{cf : \ f \in \mathcal{F}, \ 0 \leq c \leq 1, \ c^2\|f\|_{\text{Lip}}^2 \leq r\})\big] \leq \psi_n(r).$$

By Lemma 12, with probability at least $1 - \lceil \log_2(\sqrt{r \tau t n}) \rceil e^{-t}$, for every $g \in \mathcal{F}_r$,

$$\mathbb{E}_{\mathbb{P}_{\text{true}}}[g] \leq \mathbb{E}_{\mathbb{P}_n}[g] + 2\sqrt{\frac{\tau t}{n}} \|g\|_{\text{Lip}} + 2\psi_n(r) + \frac{1}{n},$$

Choose $r = r_0$, where $r_0$ is the largest solution to $\frac{1}{n} + 2\sqrt{rr_{n\star}} = r$. By Lemma 10, $r_{n\star} \leq r_0 \leq 4r_{n\star} + \frac{2}{n}$. Note that $2\psi_n(r) + \frac{1}{n} \leq 2\sqrt{rr_{n\star}} + \frac{1}{n} \leq r_0$. Let $\mathcal{F}_{r_0} \ni g = \frac{\sqrt{r_0}}{\sqrt{r_0} \vee \|f\|_{\text{Lip}}} f$. If $\|f\|_{\text{Lip}}^2 \leq r_0$, then $g = f$, therefore

$$\mathbb{E}_{\mathbb{P}_{\text{true}}}[f] \leq \mathbb{E}_{\mathbb{P}_n}[f] + 2\sqrt{\frac{\tau t}{n}} \|f\|_{\text{Lip}} + 2\psi_n(r) + \frac{1}{n} \leq \mathbb{E}_{\mathbb{P}_n}[f] + 2\sqrt{\frac{\tau t}{n}} \|f\|_{\text{Lip}} + 4r_{n\star} + \frac{2}{n}.$$

If $\|f\|_{\text{Lip}}^2 > r_0$, then $g = \frac{\sqrt{r_0}}{\|f\|_{\text{Lip}}} f$, and

$$\mathbb{E}_{\mathbb{P}_{\text{true}}}\left[\frac{\sqrt{r_0}}{\|f\|_{\text{Lip}}} f\right] \leq \frac{1}{n}\sum_{i=1}^{n} \frac{\sqrt{r_0}}{\|f\|_{\text{Lip}}} f(z_i^n) + 2\sqrt{\frac{\tau t}{n}} \frac{\sqrt{r_0}}{\|f\|_{\text{Lip}}} \|f\|_{\text{Lip}} + 2\psi_n(r) + \frac{1}{n},$$

therefore,

$$\mathbb{E}_{\mathbb{P}_{\text{true}}}[f] \leq \mathbb{E}_{\mathbb{P}_n}[f] + 2\sqrt{\frac{\tau t}{n}} \|f\|_{\text{Lip}} + \frac{r_0}{\sqrt{r_0}} \|f\|_{\text{Lip}}$$

$$\leq \mathbb{E}_{\mathbb{P}_n}[f] + \left(2\sqrt{\frac{\tau t}{n}} + \sqrt{4r_{n\star} + \frac{2}{n}}\right) \|f\|_{\text{Lip}}.$$

Combining the two cases gives the result. □

*Proof of Corollary 5.* Define $\ell \circ \mathcal{F} := \{\ell \circ f : f \in \mathcal{F}\}$. Using Lemma 9, with probability at least $1 - e^{-t}$, for every $f \in \mathcal{F}$,

$$\mathbb{E}_{\mathbb{P}_{\text{true}}}[\ell \circ f] \leq \mathbb{E}_{\mathbb{P}_n}[\ell \circ f] + \mathcal{R}_{\otimes,1}\left(\sqrt{\frac{\tau t}{n}}; -\ell \circ \mathcal{F}\right) + 2\mathbb{E}_{\otimes}[\mathfrak{R}_n(\ell \circ \mathcal{F})]$$

$$\leq \mathbb{E}_{\mathbb{P}_n}[\ell \circ f] + \mathcal{R}_{\otimes,1}\left(\sqrt{\frac{\tau t}{n}}; -\ell \circ \mathcal{F}\right) + 2L_\ell \mathbb{E}_{\otimes}[\mathfrak{R}_n(\mathcal{F})],$$

where we have used Lemma 8 to obtain the second inequality. Using arguments similar to the proofs of Lemma 11 and Lemma 12, we obtain that with probability at least $1 - \lceil \log_2(\sqrt{L_\ell \gamma_1 \tau t n}) \rceil e^{-t}$,

$$\mathbb{E}_{\mathbb{P}_{\text{true}}}[\ell \circ f] \leq \mathbb{E}_{\mathbb{P}_n}[\ell \circ f] + \sqrt{\frac{\tau t}{n}} \sup_{f \in \mathcal{F}} \|\ell \circ f\|_{\text{Lip}} + 2\mathbb{E}_{\otimes}[\mathfrak{R}_n(\ell \circ \mathcal{F})]$$

$$\leq \mathbb{E}_{\mathbb{P}_n}[\ell \circ f] + 2\sqrt{\frac{\tau t}{n}} \|\ell \circ f\|_{\text{Lip}} + 2\mathbb{E}_{\otimes}[\mathfrak{R}_n(\ell \circ \mathcal{F})] + \frac{L_\ell}{n} \quad (6)$$

$$\leq \mathbb{E}_{\mathbb{P}_n}[\ell \circ f] + 2\sqrt{\frac{\tau t}{n}} L_\ell \|f\|_{\text{Lip}} + 2L_\ell \psi_n(r) + \frac{L_\ell}{n}.$$

Define for any $r > 0$ that

$$\mathcal{F}_r := \left\{\frac{\sqrt{r}}{\sqrt{r} \vee \|f\|_{\text{Lip}}} \ell \circ f : f \in \mathcal{F}\right\} \subset \{c \ell \circ f : f \in \mathcal{F}, 0 < c \leq 1, c^2 \|f\|_{\text{Lip}}^2 \leq r\}.$$

Substituting $\frac{\sqrt{r}}{\sqrt{r} \vee \|f\|_{\text{Lip}}} \ell$ for $\ell$ in (6), we obtain that with probability at least $1 - \lceil \log_2(\sqrt{L_\ell \gamma_1 \tau t n}) \rceil e^{-t}$, for every $\mathcal{F}_r \ni g = \frac{\sqrt{r}}{\sqrt{r} \vee \|f\|_{\text{Lip}}} \ell \circ f$,

$$\mathbb{E}_{\mathbb{P}_{\text{true}}}[g] \leq \mathbb{E}_{\mathbb{P}_n}[g] + 2\sqrt{\frac{\tau t}{n}} \frac{\sqrt{r}}{\sqrt{r} \vee \|f\|_{\text{Lip}}} L_\ell \|f\|_{\text{Lip}} + 2L_\ell \psi_n(r) + \frac{L_\ell}{n}.$$

Choose $r$ to be the largest solution $r_0$ to the equation $2L_\ell\sqrt{rr_{n\star}} + \frac{L_\ell}{n} = r$. Then $2L_\ell\psi_n(r_0) + \frac{L_\ell}{n} \leq r_0$. It follows from Lemma 10 that $4L_\ell^2 r_{n\star} \leq r_0 \leq 4L_\ell^2 r_{n\star} + \frac{2L_\ell}{n}$. When $\|f\|_{\text{Lip}} \leq \sqrt{r_0}$, we have $g = \ell \circ f$ and

$$\mathbb{E}_{\mathbb{P}_{\text{true}}}[\ell \circ f] \leq \mathbb{E}_{\mathbb{P}_n}[\ell \circ f] + 2\sqrt{\frac{\tau t}{n}}L_\ell\|f\|_{\text{Lip}} + r_0 \leq \mathbb{E}_{\mathbb{P}_n}[\ell \circ f] + 2\sqrt{\frac{\tau t}{n}}L_\ell\|f\|_{\text{Lip}} + 4L_\ell^2 r_{n\star} + \frac{2L_\ell}{n}.$$

When $\|f\|_{\text{Lip}} > \sqrt{r_0}$, we have $g = \frac{\sqrt{r_0}}{\|f\|_{\text{Lip}}}\ell \circ f$ and

$$\mathbb{E}_{\mathbb{P}_{\text{true}}}\left[\frac{\sqrt{r_0}}{\|f\|_{\text{Lip}}}\ell \circ f\right] \leq \mathbb{E}_{\mathbb{P}_n}\left[\frac{\sqrt{r_0}}{\|f\|_{\text{Lip}}}\ell \circ f\right] + 2\sqrt{\frac{\tau t}{n}}L_\ell\sqrt{r_0} + 2L_\ell\psi_n(r_0) + \frac{L_\ell}{n},$$

which implies that

$$\mathbb{E}_{\mathbb{P}_{\text{true}}}[\ell \circ f] \leq \mathbb{E}_{\mathbb{P}_n}[\ell \circ f] + 2\sqrt{\frac{\tau t}{n}}L_\ell\|f\|_{\text{Lip}} + \sqrt{r_0}\|f\|_{\text{Lip}}$$

$$\leq \mathbb{E}_{\mathbb{P}_n}[\ell \circ f] + \left(2\sqrt{\frac{\tau t}{n}}L_\ell + \sqrt{4L_\ell^2 r_{n\star} + \frac{2L_\ell}{n}}\right)\|f\|_{\text{Lip}}.$$

Combining the two cases yields the result. □

**C.2.3. Proofs for** $p = 2$  Lemmas 13 and 14 below are counterparts of Lemma 11 and Lemma 12.

LEMMA 13. *Assume* $\mathbb{P}_{\text{true}}$ *satisfies* $T_2(\tau)$ *and Assumptions 2 holds. Let* $t > 0$. *Then with probability at least* $1 - e^{-t}$,

$$\mathbb{E}_{\mathbb{P}_{\text{true}}}[f] \leq \mathbb{E}_{\mathbb{P}_n}[f] + \sqrt{\frac{\tau t}{n}}(1 + \mathbb{E}_\otimes[\mathfrak{R}_n(\mathcal{G})])\sup_{f \in \mathcal{F}}\|\,\|\nabla f\|_*\,\|_{\mathbb{P}_{\text{true}},2} + 2\mathbb{E}_\otimes[\mathfrak{R}_n(\mathcal{F})] + \frac{\hbar\tau t}{n}.$$

*Proof of Lemma 13.*  In view of Lemma 9, we derive an upper bound on $\mathcal{R}_{\otimes,2}(\rho; -\mathcal{F})$. By Assumption 2,

$$\mathcal{R}_{\otimes,2}(\rho; -\mathcal{F}) = \inf_{\lambda \geq 0}\left\{\lambda\rho^2 + \mathbb{E}_\otimes\left[\sup_{f \in (-\mathcal{F})}\sup_{\tilde{z}^n \in \mathcal{Z}^n}\left\{\frac{1}{n}\sum_{i=1}^n f(\tilde{z}_i^n) - f(z_i^n) - \lambda\|\tilde{z}_i^n - z_i^n\|^2\right\}\right]\right\}$$

$$\leq \inf_{\lambda \geq 0}\left\{\lambda\rho^2 + \mathbb{E}_\otimes\left[\sup_{f \in (-\mathcal{F})}\sup_{\tilde{z}^n \in \mathcal{Z}^n}\left\{\frac{1}{n}\sum_{i=1}^n \|\nabla f(z_i^n)\|_*\|\tilde{z}_i^n - z_i^n\| - (\lambda - \hbar)\|\tilde{z}_i^n - z_i^n\|^2\right\}\right]\right\},$$

where the inner supremum is infinite if $\lambda > \hbar$. It follows by a change of variable that

$$\mathcal{R}_{\otimes,2}(\rho; -\mathcal{F}) \leq \hbar\rho^2 + \inf_{\lambda \geq 0}\left\{\lambda\rho^2 + \frac{1}{4\lambda}\mathbb{E}_\otimes\left[\sup_{f \in (-\mathcal{F})}\frac{1}{n}\sum_{i=1}^n \|\nabla f(z_i^n)\|_*^2\right]\right\}.$$

Observe that by Lemma 7,

$$\mathbb{E}_\otimes\left[\sup_{f \in (-\mathcal{F})}\frac{1}{n}\sum_{i=1}^n \|\nabla f(z_i^n)\|_*^2\right] \leq \mathbb{E}_\otimes\left[\sup_{f \in (-\mathcal{F})}\left\{\frac{1}{n}\sum_{i=1}^n \|\nabla f(z_i^n)\|_*^2 - \mathbb{E}_{\mathbb{P}_{\text{true}}}\left[\|\nabla f\|_*^2\right]\right\}\right] + \sup_{f \in (-\mathcal{F})}\mathbb{E}_{\mathbb{P}_{\text{true}}}\left[\|\nabla f\|_*^2\right]$$

$$\leq 2\mathbb{E}_\otimes[\mathfrak{R}_n(\dot{\mathcal{F}})] + \sup_{f \in \mathcal{F}}\mathbb{E}_{\mathbb{P}_{\text{true}}}\left[\|\nabla f\|_*^2\right],$$

where $\dot{\mathcal{F}} := \{\|\nabla f\|_*^2 : f \in \mathcal{F}\}$. Hence we have

$$\mathcal{R}_{\otimes,2}(\rho; -\mathcal{F}) \leq \hbar\rho^2 + \inf_{\lambda \geq 0}\left\{\lambda\rho^2 + \frac{1}{4\lambda}\left(2\mathbb{E}_\otimes[\mathfrak{R}_n(\dot{\mathcal{F}})] + \sup_{f \in \mathcal{F}}\mathbb{E}_{\mathbb{P}_{\text{true}}}\left[\|\nabla f\|_*^2\right]\right)\right\}.$$

Without loss of generality we assume $\sup_{f \in \mathcal{F}} \|\|\nabla f\|_*\|_{\mathbb{P}_{\text{true}},2} > 0$. Picking $\lambda = \frac{\sup_{f \in \mathcal{F}} \|\|\nabla f\|_*\|_{\mathbb{P}_{\text{true}},2}}{2\rho}$ yields that

$$\mathcal{R}_{\otimes,2}(\rho; -\mathcal{F}) \leq \hbar \rho^2 + \rho \sup_{f \in \mathcal{F}} \|\|\nabla f\|_*\|_{\mathbb{P}_{\text{true}},2} + \rho \frac{\mathbb{E}_\otimes[\mathfrak{R}_n(\dot{\mathcal{F}})]}{\sup_{f \in \mathcal{F}} \|\|\nabla f\|_*\|_{\mathbb{P}_{\text{true}},2}}.$$

Note that

$$\mathfrak{R}_n(\dot{\mathcal{F}}) = \mathbb{E}_\sigma \left[ \sup_{f \in \mathcal{F}} \left\{ \sum_{i=1}^n \sigma_i \frac{\|\nabla f(z_i^n)\|_*^2}{\|\|\nabla f\|_*\|_{\mathbb{P}_{\text{true}},2}^2} \cdot \|\|\nabla f\|_*\|_{\mathbb{P}_{\text{true}},2}^2 \right\} \right]$$

$$\leq \mathbb{E}_\sigma \left[ \sup_{f \in \mathcal{F}} \left\{ \sum_{i=1}^n \sigma_i \frac{\|\nabla f(z_i^n)\|_*^2}{\|\|\nabla f\|_*\|_{\mathbb{P}_{\text{true}},2}^2} \right\} \sup_{f \in \mathcal{F}} \left\{ \|\|\nabla f\|_*\|_{\mathbb{P}_{\text{true}},2}^2 \right\} \right]$$

$$= \sup_{f \in \mathcal{F}} \|\|\nabla f\|_*\|_{\mathbb{P}_{\text{true}},2}^2 \mathbb{E}_\sigma \left[ \sup_{f \in \mathcal{F}} \left\{ \sum_{i=1}^n \sigma_i \frac{\|\nabla f(z_i^n)\|_*^2}{\|\|\nabla f\|_*\|_{\mathbb{P}_{\text{true}},2}^2} \right\} \right]$$

$$= \sup_{f \in \mathcal{F}} \|\|\nabla f\|_*\|_{\mathbb{P}_{\text{true}},2}^2 \cdot \mathbb{E}_\otimes[\mathfrak{R}_n(\mathcal{G})].$$

Therefore, the result follows from Lemma 9 with $\rho = \sqrt{\frac{\tau t}{n}}$. □

LEMMA 14. *Assume $\mathbb{P}_{\text{true}}$ satisfies $T_2(\tau)$ and Assumption 2 holds. Let $t > 0$. Then with probability at least $1 - \lceil \log_2(\gamma_2 \sqrt{\tau t n}) \rceil e^{-t}$,*

$$\mathbb{E}_{\mathbb{P}_{\text{true}}}[f] \leq \mathbb{E}_{\mathbb{P}_n}[f] + 2\sqrt{\frac{\tau t}{n}}(1 + \mathbb{E}_\otimes[\mathfrak{R}_n(\mathcal{G})])\|\|\nabla f\|_*\|_{\mathbb{P}_{\text{true}},2} + 2\mathbb{E}_\otimes[\mathfrak{R}_n(\mathcal{F})] + \frac{\hbar \tau t + 1 + \mathbb{E}_\otimes[\mathfrak{R}_n(\mathcal{G})]}{n}.$$

*Proof of Lemma 14.* Set $r = \sup_{f \in \mathcal{F}} \|\|\nabla f\|_*\|_{\mathbb{P}_{\text{true}},2} \leq \gamma_2$. We define

$$\mathcal{F}_k := \left\{ f \in \mathcal{F} : 2^{-k} r < \|\|\nabla f\|_*\|_{\mathbb{P}_{\text{true}},2} \leq 2^{-k+1} r \right\}, \quad k = 1, \ldots, K-1,$$

$$\mathcal{F}_K := \left\{ f \in \mathcal{F} : \|\|\nabla f\|_*\|_{\mathbb{P}_{\text{true}},2} \leq 2^{-K} r \right\},$$

$$\mathcal{G}_k := \left\{ \frac{\|\nabla f\|_*^2}{\|\|\nabla f\|_*\|_{\mathbb{P}_{\text{true}},2}^2} : f \in \mathcal{F}_k \right\}, \quad k = 1, \ldots, K.$$

By Lemma 13, for $k = 1, \ldots, K-1$, with probability at least $1 - e^{-t}$, for every $f \in \mathcal{F}_k$,

$$\mathbb{E}_{\mathbb{P}_{\text{true}}}[f] - \mathbb{E}_{\mathbb{P}_n}[f] \leq \sqrt{\frac{\tau t}{n}}(1 + \mathbb{E}_\otimes[\mathfrak{R}_n(\mathcal{G}_k)]) 2^{-k+1} r + 2\mathbb{E}_\otimes[\mathfrak{R}_n(\mathcal{F}_k)] + \frac{\hbar \tau t}{n}$$

$$\leq 2\sqrt{\frac{\tau t}{n}}(1 + \mathbb{E}_\otimes[\mathfrak{R}_n(\mathcal{G}_k)])\|\|\nabla f\|_*\|_{\mathbb{P}_{\text{true}},2} + 2\mathbb{E}_\otimes[\mathfrak{R}_n(\mathcal{F}_k)] + \frac{\hbar \tau t}{n},$$

and with probability at least $1 - e^{-t}$, for every $f \in \mathcal{F}_K$,

$$\mathbb{E}_{\mathbb{P}_{\text{true}}}[f] \leq \mathbb{E}_{\mathbb{P}_n}[f] + \sqrt{\frac{\tau t}{n}}(1 + \mathbb{E}_\otimes[\mathfrak{R}_n(\mathcal{G}_K)]) 2^{-K} r + 2\mathbb{E}_\otimes[\mathfrak{R}_n(\mathcal{F}_K)] + \frac{\hbar \tau t}{n}.$$

Taking the union bound, with probability at least $1 - K e^{-t}$, for every $f \in \mathcal{F}$,

$$\mathbb{E}_{\mathbb{P}_{\text{true}}}[f] \leq \mathbb{E}_{\mathbb{P}_n}[f] + 2\sqrt{\frac{\tau t}{n}}(1 + \mathbb{E}_\otimes[\mathfrak{R}_n(\mathcal{G})])\|\|\nabla f\|_*\|_{\mathbb{P}_{\text{true}},2} + 2\mathbb{E}_\otimes[\mathfrak{R}_n(\mathcal{F})] + \frac{\hbar \tau t}{n}$$

$$+ \sqrt{\frac{\tau t}{n}}(1 + \mathbb{E}_\otimes[\mathfrak{R}_n(\mathcal{G}_K)]) 2^{-K} r.$$

Setting $K = \lceil \log_2(\gamma_2 \sqrt{\tau t n}) \rceil$ yields the result. □

With the two lemmas above, we are ready to prove Theorem 3.

*Proof of Theorem 3.* Let $r \geq r_{n\star}$ whose value will be specified shortly. The sub-root assumption on $\psi_n$ implies that

$$\psi_n(r) = \frac{\sqrt{r}\psi_n(r)}{\sqrt{r}} \leq \frac{\sqrt{r}\psi_n(r_{n\star})}{\sqrt{r_{n\star}}} = \sqrt{r}\sqrt{r_{n\star}}.$$

Define

$$\mathcal{F}_r := \left\{ \frac{\sqrt{r}}{\sqrt{r} \vee \|\|\nabla f\|_*\|_{\mathbb{P}_{\text{true}},2}} f : f \in \mathcal{F} \right\}.$$

Then for all $g \in \mathcal{F}_r$ it holds that $\|\|\nabla g\|_*\|^2_{\mathbb{P}_{\text{true}},2} \leq r$, $\mathcal{F}_{\gamma_2^2} = \mathcal{F}$, and

$$\mathbb{E}_\otimes[\mathfrak{R}_n(\mathcal{F}_r)] \leq \mathbb{E}_\otimes\big[\mathfrak{R}_n\big(\{cf : f \in \mathcal{F},\ 0 \leq c \leq 1,\ c^2\|\|\nabla f\|_*\|^2_{\mathbb{P}_{\text{true}},2} \leq r\}\big)\big] \leq \psi_n(r).$$

By Lemma 14, with probability at least $1 - \lceil \log_2(\sqrt{r\tau t n}) \rceil e^{-t}$, for every $g \in \mathcal{F}_r$,

$$\mathbb{E}_{\mathbb{P}_{\text{true}}}[g] \leq \mathbb{E}_{\mathbb{P}_n}[g] + 2\sqrt{\frac{\tau t}{n}}(1 + \mathbb{E}_\otimes[\mathfrak{R}_n(\mathcal{G}_r)])\|\|\nabla g\|_*\|_{\mathbb{P}_{\text{true}},2} + 2\psi_n(r) + \epsilon_n,$$

recalling $\mathcal{G}_r$ is defined in Lemma 14. Choose $r = r_0$, where $r_0$ is the largest solution to $\epsilon_n + 2\sqrt{r}\sqrt{r_{n\star}} = r$. Then $2\psi_n(r_0) + \epsilon_n \leq r_0$. By Lemma 10, $4r_{n\star} \leq r_0 \leq 4r_{n\star} + 2\epsilon_n$. Let $\mathcal{G}_{r_0} \ni g = \frac{\sqrt{r_0}}{\sqrt{r_0} \vee \|\|\nabla f\|_*\|_{\mathbb{P}_{\text{true}},2}} f$. If $\|\|\nabla f\|_*\|^2_{\mathbb{P}_{\text{true}},2} \leq r_0$, then $f = g$ and

$$\mathbb{E}_{\mathbb{P}_{\text{true}}}[f] \leq \mathbb{E}_{\mathbb{P}_n}[f] + 2\sqrt{\frac{\tau t}{n}}(1 + \mathbb{E}_\otimes[\mathfrak{R}_n(\mathcal{G})])\|\|\nabla f\|_*\|_{\mathbb{P}_{\text{true}},2} + 2\psi_n(r) + \epsilon_n$$

$$\leq \mathbb{E}_{\mathbb{P}_n}[f] + 2\sqrt{\frac{\tau t}{n}}(1 + \mathbb{E}_\otimes[\mathfrak{R}_n(\mathcal{G})])\|\|\nabla f\|_*\|_{\mathbb{P}_{\text{true}},2} + 4r_{n\star} + 2\epsilon_n.$$

If $\|\|\nabla f\|_*\|^2_{\mathbb{P}_{\text{true}},2} > r_0$, then

$$\mathbb{E}_{\mathbb{P}_{\text{true}}}\left[\frac{\sqrt{r_0}}{\|\|\nabla f\|_*\|_{\mathbb{P}_{\text{true}},2}} f\right] - \mathbb{E}_{\mathbb{P}_n}\left[\frac{\sqrt{r_0}}{\|\|\nabla f\|_*\|_{\mathbb{P}_{\text{true}},2}} f\right]$$

$$\leq 2\sqrt{\frac{\tau t}{n}}(1 + \mathbb{E}_\otimes[\mathfrak{R}_n(\mathcal{G})])\frac{\sqrt{r_0}}{\|\|\nabla f\|_*\|_{\mathbb{P}_{\text{true}},2}}\|\|\nabla f\|_*\|_{\mathbb{P}_{\text{true}},2} + 2\psi_n(r) + \epsilon_n,$$

which implies that

$$\mathbb{E}_{\mathbb{P}_{\text{true}}}[f] - \mathbb{E}_{\mathbb{P}_n}[f]$$

$$\leq 2\sqrt{\frac{\tau t}{n}}(1 + \mathbb{E}_\otimes[\mathfrak{R}_n(\mathcal{G})])\|\|\nabla f\|_*\|_{\mathbb{P}_{\text{true}},2} + (2\psi_n(r) + \epsilon_n)\frac{\|\|\nabla f\|_*\|_{\mathbb{P}_{\text{true}},2}}{\sqrt{r_0}}$$

$$\leq \left(2\sqrt{\frac{\tau t}{n}}(1 + \mathbb{E}_\otimes[\mathfrak{R}_n(\mathcal{G})]) + \sqrt{4r_{n\star} + 2\epsilon_n}\right)\|\|\nabla f\|_*\|_{\mathbb{P}_{\text{true}},2}.$$

Combining the two cases above gives the desired result. □

*Proof of Corollary 6.* Using McDiarmid's inequality, with probability at least $1 - e^{-t}$,

$$\sup_{f \in \mathcal{F}} \left| \mathbb{E}_{\mathbb{P}_n}\left[\frac{\|\nabla f(z)\|_*^2}{\|\|\nabla f\|_*\|^2_{\mathbb{P}_{\text{true}},2}}\right] - 1 \right| \leq \mathbb{E}_\otimes\left[\sup_{f \in \mathcal{F}} \left| \mathbb{E}_{\mathbb{P}_n}\left[\frac{\|\nabla f(z)\|_*^2}{\|\|\nabla f\|_*\|^2_{\mathbb{P}_{\text{true}},2}}\right] - 1 \right|\right] + \kappa_g^2\sqrt{\frac{t}{2n}},$$

which implies that for every $f \in \mathcal{F}$,

$$\frac{\|\|\nabla f(z)\|_*\|^2_{\mathbb{P}_n,2}}{\|\|\nabla f(z)\|_*\|^2_{\mathbb{P}_{\text{true}},2}} - 1 \geq -2\mathbb{E}_\otimes[\mathfrak{R}_n(\mathcal{G})] - \kappa_g^2\sqrt{\frac{t}{2n}}.$$

Thus, whenever $2\mathbb{E}_\otimes[\mathfrak{R}_n(\mathcal{G})] + \kappa_g^2\sqrt{\frac{t}{2n}} < 1/2$, it holds that

$$\|\|\nabla f\|_*\|_{\mathbb{P}_{\text{true}},2} \leq \|\|\nabla f\|_*\|_{\mathbb{P}_n,2}\Big(1 - 2\mathbb{E}_\otimes[\mathfrak{R}_n(\mathcal{G})] - \kappa_g^2\sqrt{\frac{t}{2n}}\Big)^{-\frac{1}{2}} \leq \|\|\nabla f\|_*\|_{\mathbb{P}_n,2}\Big(1 + 2\mathbb{E}_\otimes[\mathfrak{R}_n(\mathcal{G})] + \kappa_g^2\sqrt{\frac{t}{2n}}\Big).$$

Hence, setting $\tilde{\rho}_n = \rho_n(1 + 2\mathbb{E}_\otimes[\mathfrak{R}_n(\mathcal{G})] + \kappa_g^2\sqrt{\frac{t}{2n}})$ and invoking Theorem 3 and Lemma 2 yields the results. $\square$

## Appendix D: Proofs for Section 5

### D.1. Proofs for Section 5.1.1

LEMMA 15. *Under the setting in Example 1, it holds that*

$$\mathcal{R}_{\mathbb{P}_{\text{true}},1}(f_\theta;\rho_n) \leq \rho_n\|f_\theta\|_{\text{Lip}} \leq \mathcal{R}_{\mathbb{P}_n,1}(f_\theta;\tfrac{h\vee b}{h\wedge b}\rho_n).$$

*Proof.* By Lemma 1, we have $\mathcal{R}_{\mathbb{P}_{\text{true}},1}(f_\theta;\rho_n) \leq \rho_n\|f_\theta\|_{\text{Lip}} \leq \rho_n\max(h,b)\|(\theta,-1)\|_*$. On the other hand, using the duality result (1), we have

$$\mathcal{R}_{\mathbb{P}_n,1}(f_\theta;\rho_n) \geq \min_{\lambda \geq 0}\left\{\lambda\rho_n + \mathbb{E}_{\mathbb{P}_n}\left[\sup_{(x,y)\in\mathcal{Z}}\left\{\min(h,b)(|\theta^\top x - y| - |\theta^\top X - Y|) - \lambda\|(x,y) - (X,Y)\|\right\}\right]\right\}$$
$$= \rho_n\min(h,b)\|(\theta,-1)\|_*$$

Combining the two inequalities yields that

$$\mathcal{R}_{\mathbb{P}_n,1}(f_\theta;\tfrac{h\vee b}{h\wedge b}\rho_n) \geq \rho_n\|f_\theta\|_{\text{Lip}}.$$

$\square$

### D.2. Proofs for Section 5.1.2

The following lemma is used for Example 2 (see also [77, Lemma 26.10]).

LEMMA 16. *Assume $\Theta \subset \{\theta : \|\theta\|_2 \leq B\}$. Then*

$$\mathbb{E}_\otimes[\mathfrak{R}_n(\{\theta^\top x : \theta \in \Theta\})] \leq \frac{B}{\sqrt{n}}\mathbb{E}_{\mathbb{P}_{\text{true}}}[\|x\|_2^2]^{\frac{1}{2}}.$$

*Proof.* Let $\sigma_i$ be i.i.d. Rademacher random variables. Using Cauchy-Schwarz inequality and Jensen's inequality,

$$\mathbb{E}_{\sigma,\otimes}\left[\sup_{\theta\in\Theta}\frac{1}{n}\sum_{i=1}^n\sigma_i\langle\theta,x_i^n\rangle\right] = \frac{1}{n}\mathbb{E}_{\sigma,\otimes}\left[\sup_{\theta\in\Theta}\langle\theta,\sum_{i=1}^n\sigma_ix_i^n\rangle\right] \leq \frac{1}{n}\mathbb{E}_{\sigma,\otimes}\left[\sup_{\|\theta\|_2\leq B}\|\theta\|_2\cdot\|\sum_{i=1}^n\sigma_ix_i^n\|_2\right]$$
$$\leq \frac{B}{n}\mathbb{E}_{\sigma,\otimes}\left[\|\sum_{i=1}^n\sigma_ix_i^n\|_2^2\right]^{1/2} \leq \frac{B}{n}\Big(\mathbb{E}_\otimes\Big[\sum_{i=1}^n\|x_i^n\|_2^2\Big]\Big)^{\frac{1}{2}} = \frac{B}{\sqrt{n}}\mathbb{E}_{\mathbb{P}_{\text{true}}}[\|x\|_2^2]^{\frac{1}{2}}.$$

$\square$

The following lemma is used for Example 3.

LEMMA 17. *Under the setting in Example 3, it holds that*

$$\|\nabla f_\theta(z)\|_* - \|\nabla f_{\tilde{\theta}}(z)\|_* \leq L_\ell \|\tilde{\theta} - \theta\|_* + B\hbar \|x\| \|\tilde{\theta} - \theta\|_*.$$

*Proof.* We have

$$\begin{aligned}
&\|\nabla f_\theta(z)\|_* - \|\nabla f_{\tilde{\theta}}(z)\|_* \\
&= \Big| \|\theta\|_* |\ell'(\theta^\top x, y)| - \|\tilde{\theta}\|_* |\ell'(\tilde{\theta}^\top x, y)| \Big| \\
&\leq \Big| \|\theta\|_* |\ell'(\theta^\top x, y)| - \|\tilde{\theta}\|_* |\ell'(\theta^\top x, y)| \Big| + \Big| \|\tilde{\theta}\|_* |\ell'(\theta^\top x, y)| - \|\tilde{\theta}\|_* |\ell'(\tilde{\theta}^\top x, y)| \Big| \\
&\leq L_\ell \|\tilde{\theta} - \theta\|_* + \|\tilde{\theta}\|_* |\ell'(\theta^\top x, y) - \ell'(\tilde{\theta}^\top x, y)| \\
&\leq L_\ell \|\tilde{\theta} - \theta\|_* + B\hbar \|x\| \|\tilde{\theta} - \theta\|_*.
\end{aligned}$$

□

### D.3. Proofs for Section 5.1.3

The following two results on subgaussian distributions are well-known, but for the reader's convenience, we here provide proofs as well. Recall a $d$-dimensional random variable $X$ is $\sigma$-subgaussian, if for every $w \in \mathbb{R}^d$ with $\|w\|_2 = 1$ and $\epsilon > 0$, $\mathbb{P}\{|w^\top X - \mathbb{E}[w^\top X]| \geq \epsilon\} \leq 2e^{-\frac{\epsilon^2}{2\sigma^2}}$.

LEMMA 18. *Assume $\mathbb{P}_{\text{true}}^x$ satisfies $T_2(\tau)$. Let $U_n$ be defined in Example 4. Then with probability at least $1 - e^{-t}$,*

$$\min_{u \in \mathbb{R}} \sup_{\mathbb{P}: \mathcal{W}_2(\mathbb{P}, \mathbb{P}_n) \leq \rho_n} \mathbb{E}_\mathbb{P}\big[(w^\top x - u)^2 + \alpha w^\top x\big] = \min_{|u| \leq U_n} \sup_{\mathbb{P}: \mathcal{W}_2(\mathbb{P}, \mathbb{P}_n) \leq \rho_n} \mathbb{E}_\mathbb{P}\big[(w^\top x - u)^2 + \alpha w^\top x\big].$$

*Proof.* Since the Wasserstein ball is weakly compact [107], applying Sion's minimax theorem we have

$$\min_{u \in \mathbb{R}} \sup_{\mathbb{P}: \mathcal{W}_2(\mathbb{P}, \mathbb{P}_n) \leq \rho_n} \mathbb{E}_\mathbb{P}\big[(w^\top x - u)^2 + \alpha w^\top x\big] = \sup_{\mathbb{P}: \mathcal{W}_2(\mathbb{P}, \mathbb{P}_n) \leq \rho_n} \min_{u \in \mathbb{R}} \mathbb{E}_\mathbb{P}\big[(w^\top x - u)^2 + \alpha w^\top x\big]$$

Observe that the minimizer of the inner minimization problem equals $\mathbb{E}_\mathbb{P}[w^\top x]$. Moreover,

$$\mathbb{E}_\mathbb{P}[w^\top x] \leq \mathbb{E}_{\mathbb{P}_n}[w^\top x] + \|w\|_2 \mathcal{W}_1(\mathbb{P}, \mathbb{P}_n) \leq B\mathbb{E}_{\mathbb{P}_n}[\|x\|_2] + B\rho_n.$$

By Corollary 1, with probability at least $1 - e^{-t}$,

$$\mathbb{E}_{\mathbb{P}_n}[\|x\|_2] \leq \mathbb{E}_{\mathbb{P}_{\text{true}}}[\|x\|_2] + \sqrt{\frac{\tau t}{n}}.$$

Thereby we complete the proof. □

In the next two results, we compute the Rademacher complexities.

LEMMA 19. *Under the setting of Example 4, it holds that*

$$\mathbb{E}_\otimes\big[\mathfrak{R}_n(\{cf_\theta : \theta \in \Theta, 0 \leq c \leq 1, c^2 \|\|\nabla f_\theta\|_2\|^2_{\mathbb{P}_{\text{true}},2} \leq r\})\big] \leq \mu_4^2 \sqrt{\frac{r}{4\zeta n}}.$$

*Proof.* We first prove the following result. Suppose $\Theta_b \subset \{\theta = (w, \tilde{u}) \in \mathbb{R}^{d+1} : \|\theta\|_2 \leq b\}$, where $b \geq 0$. Define $\mathcal{F} = \{z \mapsto (\theta^\top z)^2 : \theta \in \Theta_b\}$. Then

$$\mathbb{E}_\otimes[\mathfrak{R}_n(\mathcal{F})] \leq \frac{b^2 \mathbb{E}_{\mathbb{P}_{\text{true}}}[\|z\|_2^4]^{\frac{1}{2}}}{\sqrt{n}}.$$

To see this, let $\{\sigma_i\}$ be i.i.d. Rademacher random variables and $z_i^n$ be i.i.d samples from $\mathbb{P}_{\text{true}}$. We have

$$\mathbb{E}_\sigma\left[\sup_{\theta \in \Theta_b} \frac{1}{n} \sum_{i=1}^n \sigma_i\left((\theta^\top z_i^n)^2\right)\right]$$

$$\leq \mathbb{E}_\sigma\left[\sup_{\theta \in \Theta_b} \frac{1}{n} \sum_{i=1}^n \sigma_i (\theta^\top z_i^n)^2\right]$$

$$= \mathbb{E}_\sigma\left[\frac{1}{n} \sup_{\theta \in \Theta_b} \langle \theta\theta^\top, \sum_{i=1}^n \sigma_i z_i^n z_i^{n\top}\rangle\right]$$

$$\leq \mathbb{E}_\sigma\left[\frac{1}{n} \sup_{\|\theta\|_2 \leq b} \|\theta\theta^\top\|_F \|\sum_{i=1}^n \sigma_i z_i^n z_i^{n\top}\|_F\right]$$

$$\leq \frac{b^2}{n} \mathbb{E}_\sigma\left[\|\sum_{i=1}^n \sigma_i z_i^n z_i^{n\top}\|_F\right].$$

Hence, the result is proved by noticing that

$$\mathbb{E}_{\otimes,\sigma}\left[\|\sum_{i=1}^n \sigma_i z_i^n z_i^{n\top}\|_F\right] \leq \mathbb{E}_\otimes\left[\left(\sum_{i=1}^n \|z_i^n z_i^{n\top}\|_F^2\right)^{\frac{1}{2}}\right] \leq \sqrt{n} \mathbb{E}_{\mathbb{P}_{\text{true}}}[\|z\|_2^4]^{\frac{1}{2}}.$$

It then follows that

$$\mathbb{E}_\otimes\left[\mathfrak{R}_n(\{cf_\theta : \theta \in \Theta, 0 \leq c \leq 1, c^2\|\|\nabla f_\theta\|_2\|_{\mathbb{P}_{\text{true}},2}^2 \leq r\})\right]$$
$$= \mathbb{E}_\otimes\left[\mathfrak{R}_n(\{z \mapsto c(\theta^\top z)^2 : \theta \in \Theta, 0 \leq c \leq 1, c^2\|\|\nabla f_\theta\|_2\|_{\mathbb{P}_{\text{true}},2}^2 \leq r\})\right]$$
$$\leq \mathbb{E}_\otimes\left[\mathfrak{R}_n(\{z \mapsto c(\theta^\top z)^2 : \theta \in \Theta, 0 \leq c \leq 1, 4c^2\zeta\|\theta\|_2^4 \leq r\})\right]$$
$$= \mathbb{E}_\otimes\left[\mathfrak{R}_n(\{z \mapsto (\theta^\top z)^2 : \theta \in \Theta, 0 \leq c \leq 1, 4\zeta\|\theta\|_2^4 \leq r\})\right]$$
$$\leq \mathbb{E}_\otimes\left[\mathfrak{R}_n\left(\{z \mapsto (\theta^\top z)^2 : \|\theta\|_2 \leq (\frac{r}{4\zeta})^{\frac{1}{4}}\}\right)\right]$$
$$\leq \mu_4^2 \sqrt{\frac{r}{4\zeta n}},$$

where in the first equality, we have used the translation invariance property of Rademacher complexity, and the second equality is due to a change of variable. □

LEMMA 20. *Let $\Theta \subset \mathbb{R}^{d+1}$. Define*

$$\mathcal{G} = \left\{z \mapsto \frac{(\theta^\top z)^2}{\mathbb{E}_{\mathbb{P}_{\text{true}}}[(\theta^\top z)^2]} : \theta \in \Theta\right\}.$$

*Assume there exists $\zeta_z > 0$ such that $\mathbb{C}\text{ov}_{\mathbb{P}_{\text{true}}}[z] \succeq \zeta_z I$. Then*

$$\mathbb{E}_\otimes[\mathfrak{R}_n(\mathcal{G})] \leq \frac{1}{\zeta_z}\sqrt{\frac{\mathbb{E}_{\mathbb{P}_{\text{true}}}[\|z\|^4]}{n}}.$$

*Proof.* Let $\{\sigma_i\}_i$ be i.i.d. Rademacher random variables and $z_i^n$ be i.i.d. samples from $\mathbb{P}_{\text{true}}$. We have

$$n\mathfrak{R}_n(\mathcal{G}) = \mathbb{E}_\sigma\left[\sup_{\theta \in \Theta} \sum_{i=1}^n \sigma_i \frac{(\theta^\top z_i^n)^2}{\mathbb{E}_{\mathbb{P}_{\text{true}}}[(\theta^\top z)^2]}\right]$$

$$= \mathbb{E}_\sigma\left[\sup_{\theta \in \Theta} \langle \frac{\theta\theta^\top}{\mathbb{E}_{\mathbb{P}_{\text{true}}}[(\theta^\top z)^2]}, \sum_{i=1}^n \sigma_i z_i^n z_i^{n\top} \rangle\right]$$

$$\leq \mathbb{E}_\sigma\left[\sup_{\theta \in \Theta} \|\frac{\theta\theta^\top}{\mathbb{E}_{\mathbb{P}_{\text{true}}}[(\theta^\top z)^2]}\|_F \cdot \|\sum_{i=1}^n \sigma_i z_i^n z_i^{n\top}\|_F\right]$$

$$\leq \frac{1}{\zeta_z} \mathbb{E}_\sigma\left[\|\sum_{i=1}^n \sigma_i z_i^n z_i^{n\top}\|_F\right].$$

The result follows by noticing that

$$\mathbb{E}_{\otimes,\sigma}\left[\|\sum_{i=1}^n \sigma_i z_i^n z_i^{n\top}\|_F\right] \leq \left(\mathbb{E}_\otimes\left[\sum_{i=1}^n \|z_i^n z_i^{n\top}\|_F^2\right]\right)^{\frac{1}{2}} = \left(n\mathbb{E}_{\mathbb{P}_{\text{true}}}[\|zz^\top\|_F^2]\right)^{\frac{1}{2}} = \sqrt{n}\left(\mathbb{E}_{\mathbb{P}_{\text{true}}}[\|z\|_2^4]\right)^{\frac{1}{2}}.$$

□

In the next result, we relate the gradient norm $\|\|\nabla f_\theta\|_2\|_{\mathbb{P}_{\text{true}},2}$ to its empirical estimate. To this end, we need the following lemma.

LEMMA 21. *Let $x_1^n, \ldots, x_n^n$ be i.i.d. samples from $\mathbb{P}_{\text{true}}$. Assume $\mathbb{P}_{\text{true}}$ satisfies $T_1(\tau)$. Let $t > 0$. Then with probability more than $1 - e^{-t}$,*

$$\max_{1 \leq i \leq n} \|x_i^n\|_2 < \mathbb{E}_{\mathbb{P}_{\text{true}}}[\|x\|_2] + \sqrt{\tau(t + \log n)}.$$

*Proof.* Since $\mathbb{P}_{\text{true}}$ satisfies $T_1(\tau)$, for every $t > 0$, $\mathbb{E}_{\mathbb{P}_{\text{true}}}\left[\exp\left(t(\|x\|_2 - \mathbb{E}_{\mathbb{P}_{\text{true}}}[\|x\|_2])\right)\right] \leq \exp(\frac{\tau t^2}{4})$, (cf. [89, Proof of Theorem 3.19]). By Jensen's inequality,

$$\exp\left(t\mathbb{E}_\otimes\left[\max_{1 \leq i \leq n} \|x_i^n\|_2 - \mathbb{E}_{\mathbb{P}_{\text{true}}}[\|x\|_2]\right]\right) \leq \mathbb{E}_\otimes\left[\exp\left(t(\max_{1 \leq i \leq n} \|x_i^n\|_2 - \mathbb{E}_{\mathbb{P}_{\text{true}}}[\|x\|_2])\right)\right]$$

$$\leq \sum_{i=1}^n \mathbb{E}_{\mathbb{P}_{\text{true}}}\left[\exp\left(t(\|x_i^n\|_2 - \mathbb{E}_{\mathbb{P}_{\text{true}}}[\|x\|_2])\right)\right]$$

$$< n\exp(\frac{\tau t^2}{4}).$$

Thus, using Markov's inequality,

$$\mathbb{P}_\otimes\left\{\max_{1 \leq i \leq n} \|x_i^n\|_2 \geq \mathbb{E}_{\mathbb{P}_{\text{true}}}[\|x\|_2] + \epsilon\right\} < \inf_{t \in \mathbb{R}} \frac{n\exp(\frac{\tau t^2}{4})}{\exp(t\epsilon)} = n\exp(-\frac{\epsilon^2}{\tau}),$$

which completes the proof. □

LEMMA 22. *Under the setting of Example 4, let $L_n = 1 + \mathbb{E}_{\mathbb{P}_{\text{true}}^x}[\|x\|_2] + \sqrt{\tau(t + \log n)}$ and . Then there exists $C > 0$ such that with probability at least $1 - 2e^{-t}$,*

$$\|\|\nabla f_\theta\|_2\|_{\mathbb{P}_{\text{true}},2} \leq C\|\|\nabla f_\theta\|_2\|_{\mathbb{P}_n,2}\left(1 + \frac{1}{1 \wedge \zeta}\sqrt{\frac{\mu_4^4 + 2\mu_2^2 + 1}{n}} + L_n^2 \sqrt{\frac{t}{2n}}\right), \quad \forall \theta \in \Theta.$$

*Proof.* We have

$$g_\theta(z) := \frac{\|f_\theta(z)\|_2}{\|\|\nabla f_\theta\|_*\|_{\mathbb{P}_{\text{true}},2}} = \frac{|\theta^\top z|}{\mathbb{E}_{\mathbb{P}_{\text{true}}}[(\theta^\top z)^2]^{\frac{1}{2}}} \leq \frac{\|z\|_2}{1 \wedge \sqrt{\zeta}}.$$

Set
$$\bar{g}_\theta(z) := g_\theta(z)\mathbf{1}\{\|z\|_2 \leq L_n\}, \quad \bar{\mathcal{G}} := \{\bar{g}_\theta : \theta \in \Theta\}.$$

Then by Lemma 21, with probability at least $1 - e^{-t}$, $\|g_\theta\|_{\mathbb{P}_n,2}^2 = \|\bar{g}_\theta\|_{\mathbb{P}_n,2}^2$. According to the proof of Corollary 6 and applying Lemma 8 on $\bar{g}_\theta(z)$, with probability at least $1 - e^{-t}$, whenever $2\mathbb{E}_\otimes[\mathfrak{R}_n(\bar{\mathcal{G}})] + L_n^2\sqrt{\frac{t}{2n}} < 1/2$, it holds that

$$\|\bar{g}_\theta\|_{\mathbb{P}_{\text{true}},2} \leq \|\bar{g}_\theta\|_{\mathbb{P}_n,2}\Big(1 + 2\mathbb{E}_\otimes[\mathfrak{R}_n(\bar{\mathcal{G}})] + L_n^2\sqrt{\frac{t}{2n}}\Big).$$

Hence, with probability at least $1 - 2e^{-t}$,

$$\|\bar{g}_\theta\|_{\mathbb{P}_{\text{true}},2} \leq \|g_\theta\|_{\mathbb{P}_n,2}\Big(1 + 2\mathbb{E}_\otimes[\mathfrak{R}_n(\bar{\mathcal{G}})] + L_n^2\sqrt{\frac{t}{2n}}\Big).$$

Moreover, since

$$\sup_{\theta \in \Theta} \frac{\mathbb{E}_{\mathbb{P}_{\text{true}}}[(\theta^\top z)^2\mathbf{1}\{\|z\|_2 > L_n\}]}{\mathbb{E}_{\mathbb{P}_{\text{true}}}[(\theta^\top z)^2]} = \sup_{\theta: \|\theta\|_2=1} \frac{\mathbb{E}_{\mathbb{P}_{\text{true}}}[(\theta^\top z)^2\mathbf{1}\{\|z\|_2 > L_n\}]}{\mathbb{E}_{\mathbb{P}_{\text{true}}}[(\theta^\top z)^2]}$$

$$\leq \frac{1}{1 \wedge \zeta}\mathbb{E}_{\mathbb{P}_{\text{true}}}\big[\|z\|_2^2\mathbf{1}\{\|z\|_2 > L_1\}\big],$$

there exists $C > 0$ such that

$$\inf_{\theta \in \Theta} \frac{\|\bar{g}_\theta\|_{\mathbb{P}_{\text{true}},2}^2}{\|g_\theta\|_{\mathbb{P}_{\text{true}},2}^2} = \inf_{\theta \in \Theta} \frac{\mathbb{E}_{\mathbb{P}_{\text{true}}}[(\theta^\top z)^2\mathbf{1}\{\|z\|_2 \leq L_n\}]}{\mathbb{E}_{\mathbb{P}_{\text{true}}}[(\theta^\top z)^2]} \geq 1/C.$$

Finally, we bound $\mathbb{E}_\otimes[\mathfrak{R}_n(\bar{\mathcal{G}})]$. Observe that

$$\mathbb{E}_\sigma\left[\sup_{\theta \in \Theta}\sum_{i=1}^n \sigma_i \frac{\|\nabla f_\theta(z_i^n)\|_2^2\mathbf{1}\{\|z_i^n\|_2 \leq L_n\}}{\|\|\nabla f_\theta\|_2\|_{\mathbb{P}_{\text{true}},2}^2}\right]$$
$$= \mathbb{E}_\sigma\left[\sup_{\theta \in \Theta}\sum_{i=1}^n \sigma_i \frac{\|\nabla f_\theta(z_i^n)\|_2^2\mathbf{1}\{\|z_i^n\|_2 \leq L_n\}}{\|\theta\|_2^4}\right] \cdot \sup_{\theta \in \Theta} \frac{\|\theta\|_2^4}{\|\|\nabla f_\theta\|_2\|_{\mathbb{P}_{\text{true}},2}^2}$$
$$\leq \mathbb{E}_\sigma\left[\sup_{\theta \in \Theta}\sum_{i=1}^n \sigma_i (\tfrac{\theta}{\|\theta\|_2}^\top z_i^n)^2\mathbf{1}\{\|z_i^n\|_2 \leq L_n\}\right] \cdot \frac{1}{1 \wedge \zeta}$$
$$\leq \frac{1}{1 \wedge \zeta}\mathbb{E}_\sigma\left[\sup_{\|\theta\|_2=1}\sum_{i=1}^n \sigma_i (\theta^\top z_i^n)^2\mathbf{1}\{\|z_i^n\|_2 \leq L_n\}\right].$$

$$\mathbb{E}_{\otimes,\sigma}\left[\sup_{\|\theta\|_2=1}\sum_{i=1}^n \sigma_i (\theta^\top z_i^n)^2\mathbf{1}\{\|z_i^n\|_2 \leq L_n\}\right] = \mathbb{E}_{\otimes,\sigma}\left[\sup_{\|\theta\|_2=1}\langle \theta\theta^\top, \sum_{i=1}^n \sigma_i z_i^n z_i^{n\top}\mathbf{1}\{\|z_i^n\|_2 \leq L_n\}\rangle\right]$$
$$\leq \mathbb{E}_{\otimes,\sigma}\left[\sup_{\|\theta\|_2=1}\|\theta\theta^\top\|_F \cdot \|\sum_{i=1}^n \sigma_i \mathbf{1}\{\|z_i^n\|_2 \leq L_n\}z_i^n z_i^{n\top}\|_F\right]$$
$$\leq \mathbb{E}_{\otimes,\sigma}\left[\|\sum_{i=1}^n \sigma_i \mathbf{1}\{\|z_i^n\|_2 \leq L_n\}z_i^n z_i^{n\top}\|_F^2\right]^{\frac{1}{2}}$$
$$\leq \mathbb{E}_\otimes\left[\sum_{i=1}^n \|z_i^n z_i^{n\top}\|_F^2\right]^{\frac{1}{2}}$$
$$= \sqrt{n}\mathbb{E}_{\mathbb{P}_{\text{true}}}[\|z\|_2^4]^{\frac{1}{2}}.$$

$\square$

### D.4. Proofs for Section 5.2.2

The next two results are used for Example 6, which relies on the following lemma.

LEMMA 23 (**Contraction for vector-valued functions**). *Let $\mathcal{H}$ be a family of m-dimensional vector-valued functions on $\mathcal{X} \subset \mathbb{R}^d$. Let $\ell : \mathbb{R}^m \to \mathbb{R}$ be an $L_\ell$-Lipschitz continuous function. Denote by $\ell \circ \mathcal{H} := \{x \mapsto \ell(h(x)) : h = (h_1, \ldots, h_m) \in \mathcal{H}\}$. Then*

$$\mathbb{E}_\otimes[\mathfrak{R}_n(\ell \circ \mathcal{H})] \leq \frac{\sqrt{2} L_\ell}{n} \mathbb{E}_{\otimes,\sigma}\left[\sup_{h \in \mathcal{H}} \sum_{i=1}^n \sum_{j=1}^m \sigma_{ij} h_j(x_i^n)\right],$$

*where $\sigma_{ij}$'s are i.i.d. Rademacher random variables. In particular, when $\mathcal{H} = \{x \mapsto Wx : W \in \mathbb{R}^{m \times d}, WW^\top = I\}$, we have*

$$\mathbb{E}_\otimes[\mathfrak{R}_n(\ell \circ \mathcal{H})] \leq \frac{\sqrt{2} L_\ell \sqrt{m \mathbb{E}_{\mathbb{P}_{\text{true}}}[\|x\|_2^2]}}{\sqrt{n}}.$$

*Proof.* The first part is due to Maurer [63, Corollary 4]. For the second part, denote $W^\top = (w_1, \ldots, w_m)$, where $w_j \in \mathbb{R}^d$ and $\|w_j\|_2 = 1$, $j = 1, \ldots, m$. We have that

$$\sup_{WW^\top = I} \sum_{j=1}^m w_j^\top \sum_{i=1}^n \sigma_{ij} x_i^n$$
$$\leq \sup_{WW^\top = I} \sum_{j=1}^m \|w_j\|_2 \cdot \|\sum_{i=1}^n \sigma_{ij} x_i^n\|_2$$
$$\leq \sqrt{m} \left(\|\sum_{i=1}^n \sigma_{ij} x_i^n\|_2^2\right)^{\frac{1}{2}}$$
$$\leq \sqrt{m \sum_{i=1}^n \|x_i^n\|_2^2}.$$

Thereby the second part of the result follows from the first part by noticing that

$$\mathbb{E}_\otimes\left[\sqrt{\sum_{i=1}^n \|x_i^n\|_2^2}\right] \leq \sqrt{\mathbb{E}_\otimes\left[\sum_{i=1}^n \|x_i^n\|_2^2\right]} = \sqrt{n \mathbb{E}_{\mathbb{P}_{\text{true}}}[\|x\|_2^2]}.$$

□

LEMMA 24. *Under the setting in Example 6, it holds that*

$$\mathbb{E}_\otimes\left[\mathfrak{R}_n\left(\{x \mapsto W_2 \phi(W_1 x) : W_1 W_1^\top = I, \|W_2\|_2 \leq B\}\right)\right] \leq B \sqrt{\frac{2 d_2 \mathbb{E}_{\mathbb{P}_{\text{true}}}[\|x\|_2^2]}{n}}.$$

*Proof.* Applying Lemma 23 with $\mathcal{H} = \{x \mapsto W_1 x : W_1 W_1^\top = I\}$ and $\ell(\cdot) = W_2 \phi(\cdot)$ yields the result. □

LEMMA 25. *Under the setting in Example 6, it holds that*

$$\mathbb{E}_{\sigma,\otimes}\left[\sup_{\theta \in \Theta} \frac{1}{n} \sum_{i=1}^n \sigma_i \frac{\|\nabla_z f_\theta(x_i^n)\|_*^2}{\|\|\nabla f_\theta\|_*\|_{\mathbb{P}_{\text{true}},2}^2}\right] \leq \frac{2L(L\hbar_\phi + 1)\sqrt{2 d_1 \mathbb{E}_{\mathbb{P}_{\text{true}}}[\|x\|_2^2]}}{\eta^2 \zeta \sqrt{n}}.$$

*Proof.* Observe that

$$\mathbb{E}_{\sigma,\otimes}\left[\sup_{\theta \in \Theta} \frac{1}{n} \sum_{i=1}^n \sigma_i \frac{\|\nabla_z f_\theta(x_i^n)\|_*^2}{\|\|\nabla f_\theta\|_*\|_{\mathbb{P}_{\text{true}},2}^2}\right] \leq \mathbb{E}_{\sigma,\otimes}\left[\sup_{\theta \in \Theta} \frac{1}{n} \sum_{i=1}^n \sigma_i \frac{\|\nabla_z f_\theta(x_i^n)\|_*^2}{\|W_2\|_2^2}\right] \cdot \sup_{\theta \in \Theta} \frac{\|W_2\|_2^2}{\|\|\nabla f_\theta\|_*\|_{\mathbb{P}_{\text{true}},2}^2}$$
$$\leq \mathbb{E}_{\sigma,\otimes}\left[\sup_{\theta \in \Theta} \frac{1}{n} \sum_{i=1}^n \sigma_i \frac{l'(W_2 \phi(W_1 x), y)^2 \|W_2 \phi'(W_1 x)\|_2^2}{\|W_2\|_2^2}\right] \cdot \frac{1}{\eta^2 \zeta}.$$

38Moreover, we have that

$$\left| l'(W_2\phi(\tilde{t}), y)^2 \frac{\|W_2\phi'(\tilde{t})\|_2^2}{\|W_2\|_2^2} - l'(W_2\phi(t), y)^2 \frac{\|W_2\phi'(t)\|_2^2}{\|W_2\|_2^2} \right|$$
$$\leq \left| l'(W_2\phi(\tilde{t}), y)^2 \frac{\|W_2\phi'(\tilde{t})\|_2^2}{\|W_2\|_2^2} - l'(W_2\phi(\tilde{t}), y)^2 \frac{\|W_2\phi'(t)\|_2^2}{\|W_2\|_2^2} \right|$$
$$+ \left| l'(W_2\phi(\tilde{t}), y)^2 \frac{\|W_2\phi'(t)\|_2^2}{\|W_2\|_2^2} - l'(W_2\phi(t), y)^2 \frac{\|W_2\phi'(t)\|_2^2}{\|W_2\|_2^2} \right|$$
$$\leq L^2 \left| \frac{\|W_2\phi'(\tilde{t})\|_2^2 - \|W_2\phi'(t)\|_2^2}{\|W_2\|_2^2} \right| + \left| l'(W_2\phi(\tilde{t}), y)^2 - l'(W_2\phi(t), y)^2 \right|$$
$$\leq 2L^2 \hbar_\phi \|\tilde{t} - t\|_2 + 2LB\hbar_l \|\tilde{t} - t\|_2.$$

Since $\|W_1 x\|_2^2 = \sum_{j=1}^{d_2} \|w_j x\|_2^2 = d_2 \|x\|_2^2$, it follows that $\nabla f_\theta$ has $2L(L\hbar_\phi + B\hbar_l)\sqrt{d_2}$-Lipschitz gradient. Applying Lemma 23 to $\mathcal{H} = \{x \mapsto W_1 x : W_1 W_1^\top = I\}$ and $\ell(\cdot) = l'(W_2\phi(\cdot))^2 \frac{\|W_2\phi'(\cdot)\|_2^2}{\|W_2\|_2^2}$, we obtain

$$\mathbb{E}_{\sigma,\otimes}\left[ \sup_{\theta \in \Theta} \frac{1}{n} \sum_{i=1}^n \sigma_i \frac{\|\nabla_z f_\theta(x_i^n)\|_*^2}{\|W_2\|_2^2} \right] \leq 2L(L\hbar_\phi + B\hbar_l)\sqrt{\frac{2d_2 \mathbb{E}_{\mathbb{P}_{\text{true}}}[\|x\|_2^2]}{n}}.$$

Hence the proof is completed. □